# Kernel Meets Sieve: Post-Regularization Confidence Bands for Sparse Additive Model

Junwei Lu* Mladen Kolar† Han Liu‡

**Abstract**

We develop a novel procedure for constructing confidence bands for components of a sparse additive model. Our procedure is based on a new kernel-sieve hybrid estimator that combines two most popular nonparametric estimation methods in the literature, the kernel regression and the spline method, and is of interest in its own right. Existing methods for fitting sparse additive model are primarily based on sieve estimators, while the literature on confidence bands for nonparametric models are primarily based upon kernel or local polynomial estimators. Our kernel-sieve hybrid estimator combines the best of both worlds and allows us to provide a simple procedure for constructing confidence bands in high-dimensional sparse additive models. We prove that the confidence bands are asymptotically honest by studying approximation with a Gaussian process. Thorough numerical results on both synthetic data and real-world neuroscience data are provided to demonstrate the efficacy of the theory.

*Department of Operations Research and Financial Engineering, Princeton University, Princeton, NJ 08544, USA; Email: junweil@princeton.edu

†Booth School of Business, The University of Chicago, Chicago, IL 60637, USA; Email: mkolar@chicagobooth.edu

‡Department of Operations Research and Financial Engineering, Princeton University, Princeton, NJ 08544, USA; Email: hanliu@princeton.edu

## 1 Introduction

Nonparametric regression investigates the relationship between a target variable $Y$ and many input variables $\boldsymbol{X} = (X_1, \dots, X_d)^T$ without imposing strong assumptions. Consider a model

$$Y = f(\boldsymbol{X}) + \varepsilon, \tag{1.1}$$

where $\boldsymbol{X} \in \mathbb{R}^d$ is a $d$-dimensional random vector in $\mathcal{X}^d$, $\varepsilon$ is random error satisfying $\mathbb{E}[\varepsilon \mid \boldsymbol{X}] = 0$, and $Y$ is a target variable. The goal is to estimate the unknown function $f : \mathbb{R}^d \mapsto \mathbb{R}$. When $d$ is small, fitting a fully nonparametric model (1.1) is feasible (Wasserman, 2006). However, the interpretation of such a model is challenging. Furthermore, when $d$ is large, consistently fitting $f(\cdot)$ is only possible under additional structural assumptions due to the curse of dimensionality.

A commonly used structural assumption on $f(\cdot)$ is that it takes an additive form

$$Y = \mu + \sum_{j=1}^{d} f_j(X_j) + \varepsilon, \quad \text{and} \quad \mathbb{E}_{X_j}[f(X_j)] = 0, \tag{1.2}$$

where $\mu$ is a constant and $f_j(\cdot)$, $j = 1, \dots, d$, are smooth univariate functions (Friedman and Stuetzle, 1981; Stone, 1985; Hastie and Tibshirani, 1990). Under an additional assumption that only $s$ components are nonzero ($s \ll d$), significant progress has been made in understanding additive models in high dimensions (Sardy and Tseng, 2004; Lin and Zhang, 2006; Ravikumar et al., 2009; Meier et al., 2009; Huang et al., 2010; Koltchinskii and Yuan, 2010; Kato, 2012; Petersen et al., 2014; Lou et al., 2014). These papers establish theoretical results on the estimation rate of sparse additive models, however, it remains unclear how to perform statistical inference for the model. Confidence bands can provide uncertainty assessment for components of the model and have been widely studied in the literature with dimension fixed (Härdle, 1989; Sun and Loader, 1994; Fan and Zhang, 2000; Claeskens and Van Keilegom, 2003; Zhang and Peng, 2010). However, it remains an open question how to construct confidence bands in high-dimensional setting, primarily because the direct generalization of those ideas is challenging. Confidence bands proposed in the classical literature with fixed dimensionality $d$ are mostly built upon kernel or local polynomial methods (Opsomer and Ruppert, 1997; Fan and Jiang, 2005), while existing estimators for sparse additive model are sieve-type estimators based on basis expansion. To bridge the gap, we propose a novel sparse additive model estimator called kernel-sieve hybrid estimator, which combines advantages from both the sieve and kernel methods. On one side, we can uniformly control the supreme norm rate of our estimator as typical sieve estimators for sparse additive models, while on the other, we can utilize the extreme value theory of kernel-type estimator to construct the confidence band.

To establish the validity of the proposed confidence bands we develop three new technical ingredients: (1) the analysis of the suprema of a high dimensional empirical process that arises from kernel-sieve hybrid regression estimator, (2) a de-biasing method for the proposed estimator, and (3) the approximation analysis for the Gaussian multiplier bootstrap procedure. The supremum norm for our estimator is derived by applying results on the suprema of empirical processes (Koltchinskii, 2011; van der Vaart and Wellner, 1996; Bousquet, 2002). The de-biasing procedure for the kernel-sieve hybrid regression estimator extends the approach used in the $\ell_1$ penalized high dimensional linear regression (Zhang and Zhang, 2013; van de Geer et al., 2014; Javanmard and Montanari, 2014). Compared to the existing literature, this is the first work considering the de-biasing procedure for a high dimensional nonparametric model. To prove the validity of the confidence band constructed by the Gaussian multiplier bootstrap, we generalize the method proposed in Chernozhukov et al. (2014a) and Chernozhukov et al. (2014b) to the high dimensional nonparametric models.

### 1.1 Related Literature

Our work contributes to two different areas, and make new methodological and technical contributions in both of them.

First, we contribute to a growing literature on high dimensional inference. Initial work on high dimensional statistics has focused on estimation and prediction (see, for example, Bühlmann and van de Geer, 2011, for a recent overview) and much less work has been done on quantifying uncertainty, for example, hypothesis testing and confidence intervals. Recently, the focus has started to shift towards the latter problems. Initial work on construction of p-values in high dimensional models relied on correct inclusion of the relevant variables (Wasserman and Roeder, 2009; Meinshausen et al., 2009). Meinshausen and Bühlmann (2010) and Shah and Samworth (2013) study stability selection procedure, which provides the family-wise error rate for any selection procedure. Hypothesis testing and confidence intervals for low dimensional parameters in high dimensional linear and generalized linear models are studied in Belloni et al. (2013a), Belloni et al. (2013c), van de Geer et al. (2014), Javanmard and Montanari (2014), Javanmard and Montanari (2013), and Farrell (2013). These methods construct honest, uniformly valid confidence intervals and hypothesis test based on the $\ell_1$ penalized estimator in the first stage. Similar results are obtained in the context of $\ell_1$ penalized least absolute deviation and quantile regression (Belloni et al., 2015, 2013b). Kozbur (2013) extends the approach developed in Belloni et al. (2013a) to a nonparametric regression setting, where a pointwise confidence interval is obtained based on the penalized series estimator. Meinshausen (2013) studies construction of one-sided confidence intervals for groups of variables under weak assumptions on the design matrix. Lockhart et al. (2014) studies significance of the input variables that enter the model along the lasso path. Lee et al. (2013) and Taylor et al. (2014) perform post-selection inference conditional on the selected model. Chatterjee and Lahiri (2013), Liu and Yu (2013), Chernozhukov et al. (2013) and Lopes (2014) study properties of the bootstrap in high-dimensions. Our work is different to the existing literature as it enables statisticians to make global inference under a nonparametric high dimensional regression setting for the first time.

Second, we contribute to the literature on high dimensional nonparametric estimation, which has recently seen a lot of activity. Lafferty and Wasserman (2008), Bertin and Lecué (2008), Comminges and Dalalyan (2012), and Yang and Tokdar (2014) study variable selection in a high dimensional nonparametric regression setting without assuming structural assumptions on $f(\cdot)$ beyond that it depends only on a subset of variables. A large number of papers have studied the sparse additive model in (1.2) (Sardy and Tseng, 2004; Lin and Zhang, 2006; Avalos et al., 2007; Ravikumar et al., 2009; Meier et al., 2009; Huang et al., 2010; Koltchinskii and Yuan, 2010; Raskutti et al., 2012; Kato, 2012; Petersen et al., 2014; Rosasco et al., 2013; Lou et al., 2014; Wahl, 2014). In addition, Xu et al. (2014) study a high dimensional convex nonparametric regression. Dalalyan et al. (2014) study the compound model, which includes the additive model as a special case. Our approach differs from the existing literature in that we consider the ATLAS model, in which the additive model is only used as an approximation to the unknown function $f(\cdot)$ at a fixed point $z$ and allow such approximation to change with $z$. Our approach only imposes a local sparsity structure and thus allows for more flexible modeling. We also develop a novel method for estimation and inference. Meier et al. (2009), Huang et al. (2010), Koltchinskii and Yuan (2010), Raskutti et al. (2012), and Kato (2012) develop estimation schemes mainly based on the basis approximation and sparsity-smoothness regularization. Our estimator approximates the function locally using a loss function combining both basis expansion and kernel method with a hybrid $\ell_1/\ell_2$-penalty. Our theoretical analysis also provides novel technical tools that were not available before and are of independent interest.

### 1.2 Organization of the Paper

The rest of the paper is organized as follows. In Section 2, we introduce the penalized kernel-sieve hybrid regression estimator as a solution to an optimization program. We then construct a confidence band for a component of a sparse additive model based on the proposed estimator. Section 3 provides the theoretical results on the statistical rate of convergence for the estimator and show that the proposed confidence band is honest. In Section 4, we generalize our method to nonparametric functions beyond sparse additive model. The numerical experiments for synthetic and real data are collected in Section 5.

### 1.3 Notation

Let $[n]$ denote the set $\{1, \ldots, n\}$ and let $\mathbb{1}\{\cdot\}$ denote the indicator function. For a vector $\boldsymbol{a} \in \mathbb{R}^d$, we let $\text{supp}(\boldsymbol{a}) = \{j \mid a_j \neq 0\}$ be the support set (with an analogous definition for matrices $\mathbf{A} \in \mathbb{R}^{n_1 \times n_2}$), $\|\boldsymbol{a}\|_q$, for $q \in [1, \infty)$, the $\ell_q$-norm defined as $\|\boldsymbol{a}\|_q = (\sum_{i \in [n]} |a_i|^q)^{1/q}$ with the usual extensions for $q \in \{0, \infty\}$, that is, $\|\boldsymbol{a}\|_0 = |\text{supp}(\boldsymbol{a})|$ and $\|\boldsymbol{a}\|_\infty = \max_{i \in [n]} |a_i|$. If the vector $\boldsymbol{a} \in \mathbb{R}^d$ is decomposed into groups such that $\boldsymbol{a} = (\boldsymbol{a}_{\mathcal{G}_1}, \ldots, \boldsymbol{a}_{\mathcal{G}_g})^T$, where $\mathcal{G}_1, \ldots, \mathcal{G}_g \subset [d]$ are disjoint sets, we denote $\|\boldsymbol{a}\|_{p,q}^q = \sum_{k=1}^g \|\boldsymbol{a}_{\mathcal{G}_k}\|_p^q$ and $\|\boldsymbol{a}\|_{p,\infty} = \max_{k \in [g]} \|\boldsymbol{a}_{\mathcal{G}_k}\|_p$ for any $p, q \in [1, \infty)$. We also denote the set $\{1, \ldots, j-1, j+1, \ldots, d\}$ as $\backslash j$ and the vector $\boldsymbol{a}_{\backslash j} = (a_1, \ldots, a_{j-1}, a_{j+1} \ldots, a_d)^T$. For the function $f \in L^2(\mathbb{R})$, we define the $L^2$ norm $\|f\|_2 = [\int f^2(x)dx]^{1/2}$ and the supremum norm $\|f\|_\infty = \sup_{x\in\mathbb{R}} |f(x)|$. For a matrix $\mathbf{A} \in \mathbb{R}^{n_1\times n_2}$, we use the notation $\mathrm{vec}(\mathbf{A})$ to denote the vector in $\mathbb{R}^{n_1 n_2}$ formed by stacking the columns of $\mathbf{A}$. We denote the Frobenius norm of $\mathbf{A}$ by $||\mathbf{A}||_F^2 = \sum_{i\in[n_1],j\in[n_2]} \mathbf{A}_{ij}^2$ and denote the operator norm as $\|\mathbf{A}\|_2 = \sup_{\|\mathbf{v}\|_2=1} \|\mathbf{A}\mathbf{v}\|_2$. For two sequences of numbers $\{\alpha_n\}_{n=1}^\infty$ and $\{\beta_n\}_{n=1}^\infty$, we use $a_n = O(\beta_n)$ to denote that $\alpha_n \le C\beta_n$ for some finite positive constant $C$, and for all $n$ large enough. If $\alpha_n = O(\beta_n)$ and $\beta_n = O(\alpha_n)$, we use the notation $\alpha_n \asymp \beta_n$. The notation $\alpha_n = o(\beta_n)$ is used to denote that $a_n\beta_n^{-1} \xrightarrow{n\to\infty} 0$. Throughout the paper, we let $c, C$ be two generic absolute constants, whose values may change from line to line.

## 2 Penalized Kernel-Sieve Hybrid Regression

In this section, we describe our new nonparametric estimator that combines the local kernel regression with the B-spline based sieve method. The goal is too estimate component functions in the additive model (1.2) and construct a confidence band for one component of the model. The kernel-sieve hybrid regression applies the local kernel regression over the component of interest and uses basis expansion for the rest of components. The group lasso penalty is used to shrink the coefficients in the expansion and select relevant variables locally.

We first introduce the Hölder class $\mathcal{H}(\gamma, L)$ of functions.

**Definition 2.1.** The $\gamma$-th Hölder class $\mathcal{H}(\gamma, L)$ on $\mathcal{X}$ is the set of $\ell = \lfloor\gamma\rfloor$ times differentiable functions $f : \mathcal{X} \mapsto \mathbb{R}$, where $\lfloor\gamma\rfloor$ represents the largest integer smaller than $\gamma$. The derivative $f^{(\ell)}$ satisfies

$$|f^{(\ell)}(x) - f^{(\ell)}(y)| \le L|x - y|^{\gamma-\ell}, \quad \text{for any } x, y \in \mathcal{X}.$$

Let $\boldsymbol{X} = (X_1, \ldots, X_d)^T$ be a $d$-dimensional random vector in $\mathcal{X}^d$. Without the loss of generality, in this paper, we assume $\mathcal{X} = [0, 1]$. The sparse additive model (SpAM) is of the form given in (1.2), with only a small number of additive components nonzero. Let $\mathcal{S} \subseteq [d]$ be of size $s = |\mathcal{S}| \ll d$. Then the model in (1.2) can be written as

$$Y = \mu + \sum_{j\in\mathcal{S}} f_j(X_j) + \varepsilon \tag{2.1}$$

with $f_j \in \mathcal{H}(2, L)$ for any $j \in \mathcal{S}$. Moreover, we assume the identifiability condition that

$$\mathbb{E}[f_j(X_j)] = 0, \text{ for all } j = 1, \ldots, d. \tag{2.2}$$

Define the sparse additive functions class

$$\mathcal{K}_d(s) = \Big\{ f = \textstyle\sum_{j \in S} f_j(X_j) \,\Big|\, |\mathcal{S}| \le s, f_j \in \mathcal{H}(2, L) \text{ and } \mathbb{E}[f_j(X_j)] = 0, \text{ for } j \in \mathcal{S} \Big\}. \tag{2.3}$$

Let $\{(\boldsymbol{X}_i, Y_i)\}_{i=1}^n$ be $n$ independent random samples of $(\boldsymbol{X}, Y)$ distributed according to (2.1). Before describing our estimator, we first introduce the centered basis functions that will be used in the estimation. Let $\{\phi_1, \ldots, \phi_m\}$ be the normalized B-spline basis functions (Schumaker, 2007). Given $m$ basis functions, we denote $f_{jm}(x)$ as the projection of $f_j$ onto the space spanned by the basis, $\mathcal{B}_m = \text{Span}(\phi_1, \ldots, \phi_m)$. In particular, we define

$$f_{jm}(\cdot) := \mathop{\arg\min}_{f \in \mathcal{B}_m} \|f - f_j\|_2 = \sum_{k=1}^m \boldsymbol{\beta}_{jk}^* \psi_{jk}^*(\cdot), \tag{2.4}$$

where $\psi_{jk}^*$'s are the locally centered bases defined as

$$\psi_{jk}^*(x) = \phi_k(x) - \mathbb{E}[\phi_k(X_j)], \text{ for all } j \in [d], m \in [k]. \tag{2.5}$$

Notice that basis functions $\{\psi_{jk}^*\}_{j \in [d], k \in [m]}$ satisfy $\mathbb{E}[\psi_{jk}^*(X_j)] = 0$. This property ensures that $f_{jm}(\cdot)$ also satisfies the identifiability condition (2.2). To compute $\psi_{jk}^*$ we need to estimate the unknown $\mathbb{E}[\phi_k(X_j)]$ by $\bar{\phi}_{jk} = n^{-1} \sum_{i=1}^n \phi_k(X_{ij})$. The centered B-spline basis in (2.5) as then $\psi_{jk}(x) = \phi_k(x) - \bar{\phi}_{jk}$.

With this notation, we are ready to introduce the penalized kernel-sieve hybrid regression estimator. Let the kernel function $K : \mathcal{X} \mapsto \mathbb{R}$ be a symmetric density function with bounded support and denote $K_h(\cdot) = h^{-1} K(\cdot/h)$ where $h > 0$ is the bandwidth. The kernel-sieve hybrid loss function at a fixed point $z \in \mathcal{X}$ is given as

$$\mathcal{L}_z(\alpha, \boldsymbol{\beta}) = \frac{1}{n}\sum_{i=1}^{n} K_h(X_{i1} - z)\Big(Y_i - \bar{Y} - \alpha - \sum_{j=2}^{d}\sum_{k=1}^{m} \psi_{jk}(X_{ij})\boldsymbol{\beta}_{jk}\Big)^2, \tag{2.6}$$

where $\bar{Y} = n^{-1}\sum_{i=1}^{n} Y_i$. Let $\boldsymbol{\beta} = (\boldsymbol{\beta}_2^T, \ldots, \boldsymbol{\beta}_d^T)^T \in \mathbb{R}^{(d-1)m}$ with $\boldsymbol{\beta}_j = (\boldsymbol{\beta}_{j1}, \ldots, \boldsymbol{\beta}_{jm})^T \in \mathbb{R}^m$ be the coefficients of B-spline basis functions. The penalized kernel-sieve hybrid estimator at $z \in \mathcal{X}$ is defined as

$$\big(\widehat{\alpha}_z, \widehat{\boldsymbol{\beta}}_z\big) = \mathop{\arg\min}_{\alpha, \boldsymbol{\beta}} \mathcal{L}_z(\alpha, \boldsymbol{\beta}) + \lambda \mathcal{R}(\alpha, \boldsymbol{\beta}), \tag{2.7}$$

where the penalty function is

$$\mathcal{R}(\alpha, \boldsymbol{\beta}) = \sqrt{m} \cdot |\alpha| + \sum_{j \geq 2} ||\boldsymbol{\beta}_j||_2 \tag{2.8}$$

with $\lambda$ being a tuning parameter. We estimate the additive functions $\{f_j\}_{j \in [d]}$ by $\widehat{f}_1(z) = \widehat{\alpha}_z$ and $\widehat{f}_j(x) = \sum_{k=1}^{m} \psi_{jk}(x)\widehat{\boldsymbol{\beta}}_{jk;z}$ for $j \geq 2$. Based on $\widehat{\alpha}_z, \widehat{\boldsymbol{\beta}}_z$, we also estimate the $d$-dimensional function $f(z, x_2, \ldots, x_d) = f_1(z) + \sum_{j=2}^{d} f_j(x_j)$ by

$$\widehat{f}(z, x_2, \ldots, x_d) = \widehat{\alpha}_z + \sum_{j=2}^{d}\sum_{k=1}^{m} \psi_{jk}(x_j)\widehat{\boldsymbol{\beta}}_{jk;z}, \tag{2.9}$$

where $\widehat{\boldsymbol{\beta}}_{jk;z}$ is the coordinate of $\widehat{\boldsymbol{\beta}}_z$ corresponding to the $k$th B-spline basis of the $j$th covariate.

**Remark 2.2.** The estimators $\widehat{\alpha}_z$ and $\widehat{\boldsymbol{\beta}}_z$ are estimating different quantities. Notice that $\widehat{\alpha}_z$ estimates the scalar $f_1(z)$, while $\widehat{\boldsymbol{\beta}}_z$ estimates the coefficients of B-splines. Given a function $g(x) = \sum_{k=1}^{m} \beta_k \phi_k(x)$, we have $\|g\|_2^2 \asymp m^{-1}\sum_{k=1}^{m} \beta_k^2$ (see, e.g., Corollary 15 in Chapter XI of de Boor (2001)). From this we see that the scales of $\widehat{\alpha}_z$ and $\widehat{\boldsymbol{\beta}}_z$ are different, which explains the additional $\sqrt{m}$ term multiplying $|\alpha|$ in the penalty function (2.8).

## 2.1 Comparison to the Sieve Estimator

In this section, we explain why we consider the kernel-sieve estimator as the first step of a confidence band construction instead of the sieve estimator. In the literature of sparse additive model estimation, most papers consider the sieve-type estimator. For example, Huang et al. (2010) consider minimizing

$$\widehat{\boldsymbol{\beta}}^{\text{sieve}} = \arg\min_{\boldsymbol{\beta}} \frac{1}{n}\sum_{i=1}^{n}\Big(Y_i - \bar{Y} - \sum_{j=1}^{d}\sum_{k=1}^{m}\psi_{jk}(X_{ij})\boldsymbol{\beta}_{jk}\Big)^2 + \lambda\sum_{j=1}^{d}||\boldsymbol{\beta}_j||_2, \tag{2.10}$$

while similar variations were considered in Ravikumar et al. (2009), Meier et al. (2009), Koltchinskii and Yuan (2010), and Kato (2012). These papers show that estimators like (2.10) are good enough to achieve the estimation consistency under the sparse additive model. However, it is hard to derive a valid confidence band from the sieve-type estimators.

If we compare the loss functions of two estimators in (2.10) and (2.6), the sieve estimator approximates the function of interest $f_1$ through its global basis expansion, while the kernel-sieve hybrid estimator only approximates $f_1$ at the local point $z$ by a scalar $\alpha$. Therefore, in order to study the asymptotic properties of the sieve estimator $\widehat{f}_1^{\text{sieve}}(x_1) = \sum_{m=1}^{d}\psi_{1k}(x_1)\widehat{\boldsymbol{\beta}}_{1k}^{\text{sieve}}$, we need to analyze the $m$-dimensional estimator $\widehat{\boldsymbol{\beta}}_1^{\text{sieve}}$ whose dimension $m$ is increasing with sample size $n$ at the rate $m \asymp n^{1/6}$. This makes it challenging to estimate the asymptotic distribution of any debiased estimator based upon $\widehat{\boldsymbol{\beta}}_1^{\text{sieve}}$ when the dimension of variables is much larger than sample size. This is why most existing papers on confidence band are based on kernel or local polynomial methods (Härdle, 1989; Sun and Loader, 1994; Fan and Zhang, 2000; Claeskens and Van Keilegom, 2003; Zhang and Peng, 2010). In comparison, the advantage of the kernel-sieve hybrid estimator is that it directly outputs a scalar estimator $\widehat{\alpha}_z$ of $f_1(z)$. This one dimensional estimator $\widehat{\alpha}_z$ allows us to construct a confidence band as we explain below. Furthermore, as we discuss in Section 4, the idea of behind the kernel-sieve hybrid estimator can be extended to a number of different classes of nonparametric models for which the estimator in (2.10) does not generalize.

## 2.2 Computational Algorithm

In this section, we describe an algorithm to minimize (2.7). We start by introducing some extra notation. Denote $\boldsymbol{\Psi} = (\boldsymbol{\Psi}_{1\bullet}, \dots, \boldsymbol{\Psi}_{n\bullet})^T \in \mathbb{R}^{n\times(1+(d-1)m)}$, where $\boldsymbol{\Psi}_{ij} = (\psi_{j1}(X_{ij}), \dots, \psi_{jm}(X_{ij}))^T$ and $\boldsymbol{\Psi}_{i\bullet} = (1, \boldsymbol{\Psi}_{i2}^T, \dots, \boldsymbol{\Psi}_{id}^T)^T \in \mathbb{R}^{1+(d-1)m}$ for $i \in [n]$ and $j \geq 2$. We also write $\boldsymbol{\Psi} = (\boldsymbol{\Psi}_{\bullet 1}, \dots, \boldsymbol{\Psi}_{\bullet d})$,

**Algorithm 1** Randomized coordinate descent for group Lasso

**for** $t = 1, 2, \ldots$ **do**

  Let $\boldsymbol{\beta}_+^{(t)} = (\boldsymbol{\beta}_1^{(t)}, \boldsymbol{\beta}_2^{(t)T}, \ldots, \boldsymbol{\beta}_j^{(t)T})^T$.

  Choose $j_t = j \in [d]$ with probability $1/d$.

  Compute $T(\boldsymbol{\beta}_j^{(t)})$ for the $j$-th block as

$$T(\boldsymbol{\beta}_j^{(t)}) = \operatorname*{arg\,min}_{\boldsymbol{\theta} \in \mathbb{R}^{\dim(\boldsymbol{\beta}_j)}} \left\{ \frac{\mu}{2} \|\boldsymbol{\theta}\|_2^2 + \langle \nabla_j \mathcal{L}_z(\boldsymbol{\beta}_+^{(t)}), \boldsymbol{\theta} \rangle + \lambda_j \|\boldsymbol{\theta} + \boldsymbol{\beta}_j^{(t)}\|_2 \right\}. \tag{2.13}$$

  Update $\boldsymbol{\beta}_j^{(t+1)} = \boldsymbol{\beta}_j^{(t)} + T(\boldsymbol{\beta}_j^{(t)})$.

**end for**

where $\boldsymbol{\Psi}_{\bullet 1} = (1, \ldots, 1)^T \in \mathbb{R}^n$ and $\boldsymbol{\Psi}_{\bullet j} = (\boldsymbol{\Psi}_{1j}, \ldots, \boldsymbol{\Psi}_{nj})^T \in \mathbb{R}^{n \times m}$ for $j \geq 2$. We further denote

$$\boldsymbol{Y} = (Y_1 - \bar{Y}, \ldots, Y_n - \bar{Y})^T \in \mathbb{R}^n, \quad \boldsymbol{\beta}_+ = (\alpha, \boldsymbol{\beta}^T)^T \in \mathbb{R}^{1+(d-1)m}, \boldsymbol{\beta}_+^* = \big(f_1^*(z), \boldsymbol{\beta}^{*T}\big)^T \in \mathbb{R}^{1+(d-1)m}$$
$$\text{and } \mathbf{W}_z = \operatorname{diag}\big(K_h(X_{11} - z), \ldots, K_h(X_{n1} - z)\big) \in \mathbb{R}^{n \times n}. \tag{2.11}$$

To unify the notation in our algorithm, we also write $\boldsymbol{\beta}_+ = (\boldsymbol{\beta}_1, \boldsymbol{\beta}_2^T, \ldots, \boldsymbol{\beta}_d^T)^T$, where $\boldsymbol{\beta}_1 = \alpha$ and $\boldsymbol{\beta} = (\boldsymbol{\beta}_2^T, \ldots, \boldsymbol{\beta}_d^T)^T$. The tuning parameters are set as $\lambda_j = \lambda\sqrt{m}$ for $j = 1$ and $\lambda_j = \lambda$ for $j \geq 2$. Using the above notation, the objective function in (2.7) can be written as

$$\mathcal{L}_z(\boldsymbol{\beta}_+) + \lambda \mathcal{R}(\boldsymbol{\beta}_+) = \frac{1}{n} (\boldsymbol{Y} - \boldsymbol{\Psi}\boldsymbol{\beta}_+)^T \mathbf{W}_z (\boldsymbol{Y} - \boldsymbol{\Psi}\boldsymbol{\beta}_+) + \lambda \mathcal{R}(\boldsymbol{\beta}_+). \tag{2.12}$$

We minimize the objective function in (2.12) using the randomized coordinate descent for composite functions (RCDC) proposed in Richtárik and Takáč (2014). Details of the procedure are given in Algorithm 1, where $\nabla_j \mathcal{L}_z(\boldsymbol{\beta}_+) := \partial \mathcal{L}_z(\boldsymbol{\beta}_+)/\partial \boldsymbol{\beta}_j$ denotes the gradient. Suppose the result of the $t$-th iteration is $\boldsymbol{\beta}_+^{(t)}$. In the next iteration, we randomly choose one coordinate $j_{t+1}$ from $\{1, \ldots, d\}$ and update the $\boldsymbol{\beta}_{j_t}^{(t)}$. Each update in (2.13) can be obtained in a closed form as

$$T(\boldsymbol{\beta}_j^{(t)}) = \mathcal{T}_{\lambda_j/\mu}\Big(\boldsymbol{\beta}_j^{(t)} - \frac{1}{L} \nabla_j \mathcal{L}_z\big(\boldsymbol{\beta}_+^{(t)}\big)\Big) - \boldsymbol{\beta}_j^{(t)}, \tag{2.14}$$

where $\mu$ is certain regularized constant and $\mathcal{T}_\lambda$ is the soft-thresholding operator, which is defined as

$\mathcal{T}_\lambda(\mathbf{v}) = (\mathbf{v}/\|\mathbf{v}\|_2) \cdot \max\{0, \|\mathbf{v}\|_2 - \lambda\}$. If we evaluate the estimator $\widehat{\alpha}_z$ for $M$ different $z$'s, a naïve approach is to run Algorithm 1 for $M$ times. The computational complexity is $O(dm^2nM)$. However, we propose a method to accelerate Algorithm 1 by exploiting the special structure of kernel functions. The accelerated method improves the computational complexity to $O(dm^2(n + M))$. Therefore, the computational complexity of our method is comparable to applying RCDC to minimize the objective function in (2.10) for SpAM estimation. More details can be found in Section A in the supplementary material.

### 2.3 Confidence Band

In this section, we present a procedure for constructing confidence band for the additive component $f_1$ based on a de-biased estimator. A confidence band $\mathcal{C}_n$ is a set of confidence intervals $\mathcal{C}_n = \{\mathcal{C}_n(z) = [c_L(z), c_U(z)] \,|\, z \in \mathcal{X}\}$. For simplicity, we define the interval $c_0(z) \pm r_0(z) := [c_0(z) - r_0(z), c_0(z) + r_0(z)]$. We use $f \in \mathcal{C}_n$ to denote that $f$ lies in the confidence band, that is, $f(z) \in \mathcal{C}_n(z)$ for all $z \in \mathcal{X}$. Our idea for constructing the confidence band extends the results developed for de-biased estimators for high-dimensional linear regression in Zhang and Zhang (2013), van de Geer et al. (2014), and Javanmard and Montanari (2014). Our setting is much more challenging as it involves constructing a band for an infinite dimensional object and we need a novel correction for $\widehat{\alpha}_z$ that reduces the bias introduced by (2.7).

We define for any $\boldsymbol{v} = (v_1, \boldsymbol{v}_2^T, \ldots, \boldsymbol{v}_m^T)^T \in \mathbb{R}^{(d-1)m+1}$ with $v_1 \in \mathbb{R}$ and $\boldsymbol{v}_j \in \mathbb{R}^m$ for $j \geq 2$, the norm $\|\boldsymbol{v}\|_{2,\infty} = \max(|v_1|, \|\boldsymbol{v}_2\|_2, \ldots, \|\boldsymbol{v}_d\|_2)$. Consider the following convex program

$$\widehat{\boldsymbol{\theta}}_z = \operatorname*{arg\,min}_{\boldsymbol{\theta} \in \mathbb{R}^{(d-1)m+1}} \boldsymbol{\theta}^T \widehat{\boldsymbol{\Sigma}}_z \boldsymbol{\theta}, \qquad \text{subject to} \qquad \big\|\widehat{\boldsymbol{\Sigma}}_z \boldsymbol{\theta} - \mathbf{e}_1\big\|_{2,\infty} \leq \gamma, \tag{2.15}$$

where $\widehat{\boldsymbol{\Sigma}}_z = n^{-1} \boldsymbol{\Psi} \mathbf{W}_z \boldsymbol{\Psi}^T$ and $\mathbf{e}_1$ is the first canonical basis in $\mathbb{R}^{(d-1)m+1}$. The de-biased estimator is given as

$$\widehat{f}_1^u(z) = \widehat{\alpha}_z + \frac{1}{n} \widehat{\boldsymbol{\theta}}_z^T \boldsymbol{\Psi}^T \mathbf{W}_z (\boldsymbol{Y} - \boldsymbol{\Psi} \widehat{\boldsymbol{\beta}}_+). \tag{2.16}$$

We proceed to construct a confidence band based on this de-biased estimator by considering the distribution of the process $\sup_{z \in \mathcal{X}} \sqrt{nh}(\widehat{f}_1^u(z) - f_1(z))$. We can approximate the distribution of the empirical process by the Gaussian multiplier process

$$\widehat{\mathbb{H}}_n(z) = \frac{1}{\sqrt{nh^{-1}}} \sum_{i=1}^{n} \xi_i \cdot \frac{\widehat{\sigma} K_h(X_{i1} - z) \mathbf{\Psi}_i^T \widehat{\boldsymbol{\theta}}_z}{\widehat{\sigma}_n(z)}, \tag{2.17}$$

where $\xi_1, \dots, \xi_n$ are independent $N(0,1)$ random variables, and the variance estimators are given as $\widehat{\sigma}^2 = n^{-1} \sum_{i=1}^{n} (Y_i - \widehat{\alpha}_{X_i} - \sum_{j=2}^{d} \sum_{k=1}^{m} \mathbf{\Psi}_{ij}^T \widehat{\boldsymbol{\beta}}_{jk;X_i})^2$ and $\widehat{\sigma}_n^2(z) = n^{-1} \widehat{\boldsymbol{\theta}}_z^T \mathbf{\Psi} \mathbf{W}_z^2 \mathbf{\Psi}^T \widehat{\boldsymbol{\theta}}_z$. Let $\widehat{c}_n(\alpha)$ be the $(1-\alpha)$th quantile of $\sup_{z \in \mathcal{X}} \widehat{\mathbb{H}}_n(z)$. We construct the confidence band at level $100 \times (1-\alpha)\%$: $\mathcal{C}_{n,\alpha}^b = \{\mathcal{C}_{n,\alpha}^b(z) \mid z \in \mathcal{X}\}$, where

$$\mathcal{C}_{n,\alpha}^b(z) := \widehat{f}_1^u(z) \pm \widehat{c}_n(\alpha)(nh)^{-1/2} \widehat{\sigma}_n(z). \tag{2.18}$$

We will show that the confidence band is asymptotically honest in Section 3.2 by building on the framework developed in Chernozhukov et al. (2014a) and Chernozhukov et al. (2014b), who study Gaussian multiplier bootstrap for approximating the distribution of the suprema of an empirical process.

# 3 Theoretical Properties

We establish the rate of convergence for the proposed estimator in Section 3.1, while the confidence band for $f_1$ is analyzed in Section 3.2.

## 3.1 Estimation Consistency

We start with stating the required assumptions. Let $p(x_1, \dots, x_d)$ denote the joint density of $\boldsymbol{X} = (X_1, \dots, X_d)$ and let $p_j(x_j)$ denote the marginal density of $X_j$, for $j \in [d]$. Furthermore, let $p_{abc}(x_a, x_b, x_c)$ denote the joint density of $(X_a, X_b, X_c)$ for $a, b, c \in [d]$.

**(A1)** (Density function) The density function $p(x_1, \dots, x_d)$ is continuous on $\mathcal{X}^d$ and its support $\mathcal{X}$ is compact. For each $j \in [d]$, the marginal density $p_j \in \mathcal{H}(L, 2)$. There exist fixed constants $0 < b \le B < \infty$ such that $b \le p_{1ac}(x_1, x_a, x_b) \le B$ for all $a, b \in \{2, \dots, d\}$.

**(A2)** (Kernel function) The kernel $K(u)$ is a continuous function with a bounded support satisfying

$$\int_{\mathcal{X}} K(u)du = 1 \quad \text{and} \quad \int_{\mathcal{X}} uK(u)du = 0.$$

**(A3)** (Design Matrix) Let $\boldsymbol{\Sigma}_z = \mathbb{E}[K_h(X_1 - z)\boldsymbol{\Psi}_{1\bullet}\boldsymbol{\Psi}_{1\bullet}^T]$, recalling that $\boldsymbol{\Psi}_{1\bullet} = (1, \boldsymbol{\Psi}_{12}, \ldots, \boldsymbol{\Psi}_{1d})^T$. For any $J \subset [d]$, we define a cone

$$\mathbb{C}_\beta^{(\kappa)}(J) = \Big\{ \boldsymbol{\beta}_+ = (\alpha, \boldsymbol{\beta}^T)^T \,\big|\, \textstyle\sum_{j \notin J, j \neq 1} \|\boldsymbol{\beta}_j\|_2 \leq \kappa \sum_{j \in J, j \neq 1} \|\boldsymbol{\beta}_j\|_2 + \kappa\sqrt{m}|\alpha| \Big\}. \tag{3.1}$$

There exists a universal constant $\rho_{\min}$ independent to $n, d, z$ such that the restricted minimum eigenvalue on $\mathbb{C}_\beta^{(\kappa)}(J)$ satisfies

$$\inf_{z \in \mathcal{X}} \inf_{|J| \leq s} \inf_{\boldsymbol{\beta}_+ \in \mathbb{C}_\beta^{(\kappa)}(J)} \frac{\boldsymbol{\beta}_+^T \boldsymbol{\Sigma}_z \boldsymbol{\beta}_+}{\|\boldsymbol{\beta}\|_2^2 + m\alpha^2} \geq \frac{\rho_{\min}}{m}. \tag{3.2}$$

**(A4)** (Noise Term) The error term $\varepsilon$ satisfies $\mathbb{E}[\varepsilon \,|\, \boldsymbol{X}] = 0$ almost surely and is a subgaussian random variable such that $\mathbb{E}[\exp(\lambda\varepsilon)] \leq \exp(\lambda^2\sigma_\epsilon^2/2)$ for any $\lambda$.

**(A5)** The nonparametric function $f(x_1, \ldots, x_d) \in \mathcal{K}_d(s)$ defined in Definition 2.3.

Assumption (**A1**) on the density function $p(\cdot)$ of covariates is stronger than the one used in Huang et al. (2010), where the univariate densities $\{p_j(x_j)\}_{j\in[d]}$ are bounded away from infinity and zero. However, Assumption (**A1**) is commonly used for the kernel-type methods. For example, Opsomer and Ruppert (1997) and Fan and Jiang (2005) study the additive model with two covariates: $Y = \mu + f_1(X_1) + f_2(X_2) + \epsilon$ and impose

$$\sup_{x_1, x_2 \in \mathcal{X}} \left| \frac{p(x_1, x_2)}{p_1(x_1)p_2(x_2)} - 1 \right| < 1, \tag{3.3}$$

which implies that $p(x_1, x_2)$ is bounded from infinity and zero. As we describe in Section 4, estimating a local additive model boils down to estimating additive components that are functions of two variables. Therefore, we need a boundedness assumption on the joint density of three covariates due to more complicated interactions. Assumption (**A2**) is standard in the literature on local linear regression (Fan, 1993), while Assumption (**A4**) is standard in the literature on sparse additive modeling (Meier et al., 2009; Huang et al., 2010; Koltchinskii and Yuan, 2010; Raskutti et al., 2012; Kato, 2012).

Assumption (**A3**) is similar to the restricted strong convexity condition in Negahban et al. (2012). Note that $\boldsymbol{\Sigma}_z$ is the expectation of the Hessian matrix of the loss function $\mathcal{L}(\boldsymbol{\beta}_+)$. We require $\boldsymbol{\Sigma}_z$ to be positive definite when restricted to vectors in the cone $\mathbb{C}_{\beta}^{(\kappa)}(J)$. Again, the additional factor $\sqrt{m}$ in front of $|\alpha|$ makes sure that $\alpha$ and $\boldsymbol{\beta}_z$ are calibrated on the same scale (see Remark 2.2). Assumption (**A3**) can be derived from the assumption on the design in Koltchinskii and Yuan (2010). They consider the quantity

$$\beta_{2,\kappa}(J) = \inf\left\{\beta > 0 \,\middle|\, \textstyle\sum_{j\in J}\|h_j\|_2^2 \le \beta^2\big\|\sum_{j=1}^d h_j\big\|_2^2,\ (h_1,\dots,h_d)\in\mathbb{C}_h^{(\kappa)}(J)\right\}, \tag{3.4}$$

where $\mathbb{C}_h^{(\kappa)}(J) = \big\{(h_1,\dots,h_d)\,\big|\,\sum_{j\notin J}\|h_j\|_2 \le \kappa\sum_{j\in J}\|h_j\|_2\big\}$ for $J \subset [d]$.

The following proposition describes the connection between $\beta_{2,\kappa}(J)$ and Assumption (**A3**).

**Proposition 3.1.** We define a uniform quantity based on the constant (3.4) as

$$\bar{\beta}_{2,\kappa} = \sup_{|J|\le s}\inf\left\{\beta > 0 \,\middle|\, \textstyle\sum_{j\in J}\|h_j\|_2^2 \le \beta^2\big\|\sum_{j=1}^d h_j\big\|_2^2,\ (h_1,\dots,h_d)\in\mathbb{C}_h^{(\kappa)}(J)\right\}. \tag{3.5}$$

Under Assumption (**A1**), there exist constants $c, C > 0$ such that

$$\inf_{z\in\mathcal{X}}\inf_{|J|\le s}\inf_{\boldsymbol{\beta}_+\in\mathbb{C}_{\beta}^{(\kappa)}(J)}\frac{\boldsymbol{\beta}_+^T\boldsymbol{\Sigma}_z\boldsymbol{\beta}_+}{\|\boldsymbol{\beta}\|_2^2 + m\alpha^2} \ge \frac{Cb}{sB^2(c\kappa+1)^2\bar{\beta}_{2,c\kappa}^2}\frac{1}{m}.$$

Proposition 3.1 implies that if the number of active components $s$ is finite, $\bar{\beta}_{2,\kappa} < \infty$ is sufficient to guarantee Assumption (**A3**). The assumption that $s$ is finite is required in the previous works (Meier et al., 2009; Huang et al., 2010; Koltchinskii and Yuan, 2010; Kato, 2012). The proof of Proposition 3.1 is stated in Appendix C.1 in the supplementary material.

In the following, we present the rate of convergence of the kernel-sieve hybrid regression estimator.

**Theorem 3.2.** Suppose that Assumptions (**A1**)-(**A5**) are satisfied. There exists a constant $C$ such that if $h = o(1), m \to \infty$ as $n \to \infty$, and

$$\lambda \geq C\left(\sqrt{\frac{\log(dmh^{-1})}{nh}} + \frac{\sqrt{s}}{m^{5/2}} + \frac{m^{3/2}\log(dh^{-1})}{n} + \frac{h^2}{\sqrt{m}}\right), \tag{3.6}$$

the estimator $(\widehat{\alpha}_z, \widehat{\boldsymbol{\beta}}_z^T)^T$ defined in (2.7) satisfies

$$\sup_{z\in\mathcal{X}} \sum_{j=2}^{d} ||\widehat{\boldsymbol{\beta}}_{j;z} - \boldsymbol{\beta}_j^*||_2 \leq \frac{sm}{\rho_{\min}}\lambda \qquad \text{and} \qquad \sup_{z\in\mathcal{X}} |\widehat{a}_z - f_1(z)| \leq \frac{s\sqrt{m}}{\rho_{\min}}\lambda$$

with probability $1 - c/n$ for some constant $c > 0$, where $\widehat{\boldsymbol{\beta}}_{j;z}$ is a sub-vector of $\widehat{\boldsymbol{\beta}}_z$ corresponding to the coefficients of B-spline basis of the $j$th covariate and same for $\boldsymbol{\beta}_j^*$ to $\boldsymbol{\beta}^*$ defined in (2.4). Furthermore, the estimator $\widehat{f}$ in (2.9) satisfies

$$\|\widehat{f} - f\|_2 \leq \rho_{\min}^{-1} s\sqrt{m}\lambda \tag{3.7}$$

with probability $1 - c/n$.

The estimation error comes from four sources. The noise $\varepsilon$ contributes $O\left(\sqrt{\log(dmh^{-1})/nh}\right)$ in (3.6). The second term in (3.6), $O\left(\sqrt{s}m^{-5/2}\right)$, comes from the approximation error introduced by using $m$ B-spline basis functions to estimate the true functions $\{f_j\}_{j=2}^d$. The third source of error comes from the kernel method, which uses a constant to estimate $f_{1z}$ locally. The fourth source of error comes from searching for correct local approximation by $s$ additive functions due to (4.1). Both the third and fourth sources contribute $O\left(n^{-1}m^{3/2}\log(dh^{-1}) + h^2/\sqrt{m}\right)$ to the estimation error. The detailed proof of Theorem 3.2 is shown in Section 7.

The statistical rate in (3.7) is minimized when we choose $h \asymp n^{-1/6}$, $m \asymp n^{1/6}$ and $\lambda \asymp n^{-5/12}\sqrt{\log(dn)}$. With these choices, we obtain $\|\widehat{f} - f\|_2^2 = O_P\left(n^{-2/3}\log(dn)\right)$. This convergence rate is slower than the optimal rate $O_P\left(n^{-4/5} + \log d/n\right)$ for estimating the sparse additive model (Raskutti et al., 2012). However, we will show that this rate is enough to construct an honest confidence band for $f_1$ in Section 3.2. Besides, our kernel-sieve hybrid estimator can be applied to functions beyond the sparse additive model. It can actually estimate the functions in the form $f_1(x_1) + \sum_{j=2}^{d} f_j(x_j, x_1)$, which has two dimensional additive functions. We refer Section 4 for the details of the generalization. In fact, the rate $\|\widehat{f} - f\|_2^2 = O_P\left(n^{-2/3}\log(dn)\right)$ we achieve is nearly optimal up to logarithmic factors for the two dimensional Hölder class (Stone, 1980). Technically, the slower rate comes from the error term $T_n = \sup_{z\in\mathcal{X}} \max_{j\in[d]} \frac{1}{n}\|\boldsymbol{\Psi}_{\bullet j}^T \mathbf{W}_z \boldsymbol{\varepsilon}\|_2 = O_P\left(\sqrt{\log(dn)/(nh)}\right)$, where $\mathbf{W}_z$ is defined in (2.2). In comparison, Huang et al. (2010) only need to bound $T_n' = \sup_{z\in\mathcal{X}} \max_{j\in[d]} \frac{1}{n}\|\boldsymbol{\Psi}_{\bullet j}^T \boldsymbol{\varepsilon}\|_2 = O_P\left(\sqrt{\log(dn)/n}\right)$ (see their Lemma 2). Note that $T_n = O_P\left(h^{-1/2}T_n'\right)$ because the kernel matrix $\mathbf{W}_z$ increases its variance by $O_P(h^{-1/2})$. Detailed technical analysis of $T_n$ is given in Lemma 7.4.

### 3.2 Theoretical Results for Confidence Band

In order to establish valid theoretical results on the confidence band $\mathcal{C}_{n,\alpha}^b$, we need to strengthen the weak dependency assumption in Assumption (**A3**) as follows.

**Assumption (A6).** (Nonparametric Weak Dependency) Recall that the constant $B$ is defined in Assumption (**A1**) and $\rho_{\min}$ is defined in (3.2). We assume that the 3-tuple density functions of $\boldsymbol{X}$ satisfies

$$\sup_{k\geq 2}\sum_{j\geq 2}\iiint_{\mathcal{X}^3}\left|\frac{p_{1,j,k}(x_1,x_j,x_k)}{p_1(x_1)p_j(x_j)p_k(x_k)}-1\right|dx_1dx_jdx_k \leq \frac{\rho_{\min}}{2B}. \tag{3.8}$$

The nonparametric weak dependency assumption above quantifies how strong the dependency can be among the covariates to guarantee an honest confidence band. This assumption is a high dimensional analogue of the assumption in (3.3), which is considered by Opsomer and Ruppert (1997) and Fan and Jiang (2005) for fixed dimensional additive model.

The next lemma provides guidance to the selection of the tuning parameter $\gamma$ in (2.15).

**Lemma 3.3.** Suppose that Assumptions (**A1**), (**A3**) and (**A6**) hold. Let

$$\gamma = C\log d\sqrt{m/nh}, \tag{3.9}$$

for sufficiently large constant $C$. Then the vector $\boldsymbol{\theta}_z = \boldsymbol{\Sigma}_z^{-1}\mathbf{e}_1$ is a feasible solution to the optimization program in (2.15) with high probability. In particular, we have

$$\mathbb{P}\left(\|\widehat{\boldsymbol{\Sigma}}_z\boldsymbol{\theta}_z - \mathbf{e}_1\|_{2,\infty} \leq \gamma\right) \geq 1 - c/d$$

for some constant $c$.

We defer the proof of this lemma to Appendix E.2 in the supplementary material. We are now ready to present the main theorem of this section which establishes a valid confidence band for a component in the sparse additive model under the identifiability condition (2.2).

**Theorem 3.4.** We consider the SpAM model in (2.1) with identifiability condition (2.2). Suppose $\varepsilon \sim N(0, \sigma^2)$ and that Assumptions (**A1**) - (**A6**) hold. If $h \asymp n^{-\delta}$ for $\delta > 1/5$ and $m \asymp n^p$ for $p \in (0, (10\delta - 2)/3)$, there exist constants $c, C > 0$ such that for any $\alpha \in (0, 1)$, the covering probability of $\mathcal{C}^b_{n,\alpha}$ in (2.18) is

$$\mathbb{P}\big(f_1 \in \mathcal{C}^b_{n,\alpha}\big) \geq 1 - \alpha - Cn^{-c}. \tag{3.10}$$

In particular, the confidence band $\mathcal{C}^b_{n,\alpha}$ is asymptotically honest, that is,

$$\liminf_{n\to\infty} \mathbb{P}\big(f_1 \in \mathcal{C}^b_{n,\alpha}\big) \geq 1 - \alpha.$$

For the detailed proof of this theorem, see Appendix B.1 in the supplementary material.

# 4 Generalization to Larger Nonparametric Family

In this section, we will show that our kernel-sieve estimator defined in (2.7) can be applied to a family of functions larger than the sparse additive model. We call this new function family as the additive local approximation model with sparsity (ATLAS). Notice that under the SpAM model, there are no interaction terms between different covariates. In addition, the set of covariates in $\mathcal{S}$ affect the response $Y$ globally. The ATLAS model relaxes these two structural constraints.

**Definition 4.1.** A $d$-dimensional function $f(x_1, \ldots, x_d)$ has a local sparse additive approximation for $x_1$ if for any $z \in \mathcal{X}$, there exist functions $f_{1z}(\cdot), \ldots, f_{dz}(\cdot) \in \mathcal{H}(2, L)$, two bounded functions $L(\cdot) : \mathcal{X}^d \mapsto \mathbb{R}$, $Q(\cdot) : \mathcal{X} \mapsto \mathbb{R}$ and a constant $\delta_0 > 0$ such that for any $\boldsymbol{x}_{-1} = (x_2, \ldots, x_d)^T \in \mathcal{X}^{d-1}$, if $x_1 \in (z-\delta_0, z+\delta_0)$, we have the approximation

$$\left| f(x_1,\ldots,x_d) - f_1(z) - \sum_{j=1}^{d} f_{jz}(x_j) - L(z,\boldsymbol{x}_{-1})(x_1 - z) \right| \le Q(z)(x_1-z)^2. \tag{4.1}$$

Furthermore, we assume that the locally additive approximation functions are sparse in that at most $s$ of the functions $\{f_{jz}(\cdot)\}_{j=1}^d$ are not identical to zero. The sparsity pattern at each $z \in \mathcal{X}$ is denoted as $\mathcal{S}_z = \{j \in [d] : f_{jz}(\cdot) \not\equiv 0\}$. We call the function class containing functions satisfying Definition 4.1 the ATLAS model and denote it as $\mathcal{A}_d(s)$.

By letting $z \to x_1$ in (4.1), we observe that a function in the ATLAS model can be written as

$$f(x_1,\ldots,x_d) = f_1(x_1) + \sum_{j=2}^{d} f_j(x_j, x_1), \tag{4.2}$$

where $\{f_j(x_j,x_1)\}_{j=2}^d$ are $d$ bivariate functions belonging to $\mathcal{H}(2,L)$. Similar to (2.2), we impose the identifiability condition

$$\mathbb{E}\big[f_1(X_1)\big] = 0 \text{ and } \mathbb{E}\big[f_j(X_j,x_1)\big] = 0 \text{ for any } x_1 \in \mathcal{X} \text{ and } j = 2,\ldots,d. \tag{4.3}$$

We call $X_1$ the longitude variable and the functions $f_2(\cdot,z),\ldots,f_d(\cdot,z)$ for each $z \in \mathcal{X}$ as charts at longitude $z$. Notice that the sparsity patterns of charts may change with $z \in \mathcal{X}$, allowing for more flexible modeling compared to SpAM which assume a fixed sparsity pattern. Therefore, ATLAS allows complex nonlinear interaction between $X_1$ and other covariates. A visualization of a $d$-dimension function in ATLAS is illustrated in Figure 1.

It is obvious that the sparse additive model is a subset of ATLAS with the fixed charts $\{f_j\}_{j=1}^d$ which are invariant to any longitude variable. In fact, ATLAS model generalizes many existing nonparametric models in the literature. Functions like (4.2) are studied under the framework of time-varying additive models for longitudinal data (Zhang and Wang, 2014) when the dimension is fixed. It has also been considered as compound functional model proposed in Dalalyan et al. (2014) under the high dimensional setting. However, ATLAS allows the sparsity pattern to vary with the longitude covariate $x_1$ while the compound functional model in Dalalyan et al. (2014) must have fixed support. The following example gives another subset of ATLAS model.

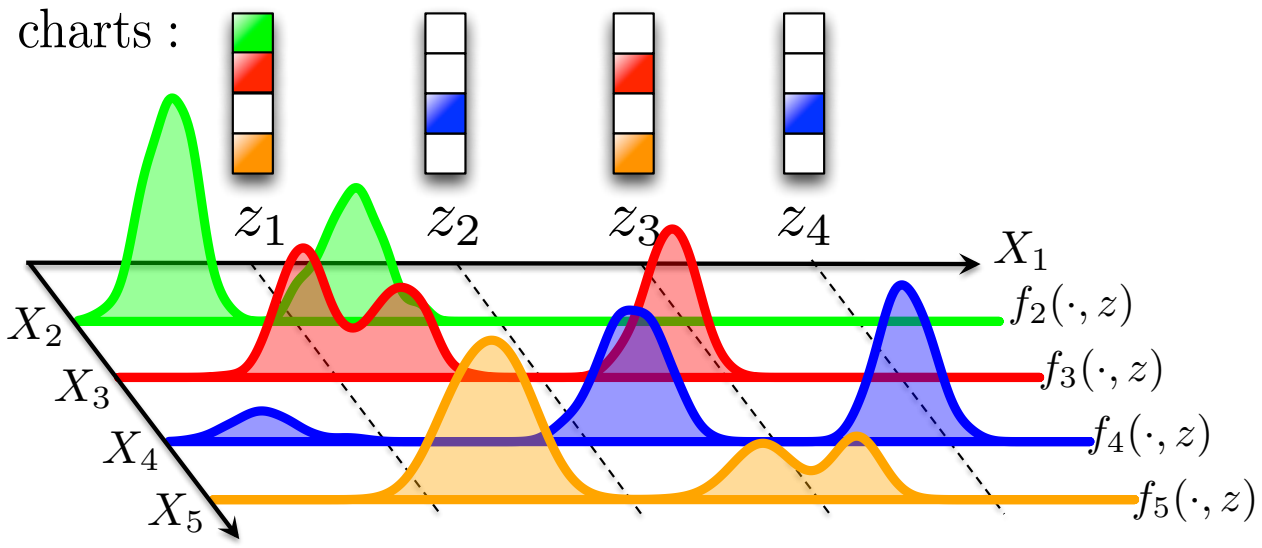

Figure 1: The illustration of ATLAS. As the longitude variable $X_1$ changes as $X_1 \in \{z_1, z_2, z_3, z_4\}$, the sparsity patterns of the charts are different. By fixing the lattitude variable $X_j$ for $j = 2, \ldots, 5$, the values of charts $f_j(\cdot, z)$ change with $z$. Under the sparsity assumption, $f_j(\cdot, z)$ is zero for most of the range of $z$.

**Example 4.2.** Consider a $d$-dimensional function with the structure

$$f(x_1, \ldots, x_d) = f_1(x_1) + \sum_{j=2}^{d} a_j(x_1) f_j(x_j), \tag{4.4}$$

where $a_j(\cdot), f_k(\cdot) \in \mathcal{H}(2, L)$ for all $k \in [d]$ and $j \geq 2$. Moreover, for any fixed $z \in \mathcal{X}$, at most $s$ of $\{a_j(z)\}_{j \geq 2}$ are nonzero. The function in (4.4) satisfies Definition 4.1. We define $f_j(x_j, x_1) = a_j(x_1) f_j(x_j)$ for $j = 2, \ldots, d$ and let $L(z, \boldsymbol{x}_{-1}) = \sum_{j \geq 2} a_j'(z) f_j(x_j)$. Then for any $x_1 \in (z - \delta_0, z + \delta_0)$ and $\boldsymbol{x}_{-1} \in \mathcal{X}^{d-1}$, we have

$$\left| f(x_1, \ldots, x_d) - \sum_{j=1}^{d} f_j(x_j, z) - L(z, \boldsymbol{x}_{-1})(x_1 - z) \right| \leq s \max_{j \in [d]} \|f_j\|_\infty \|a_j''\|_\infty (x_1 - z)^2 := Q(z)(x_1 - z)^2,$$

which satisfies Definition 4.1 if $s$ is finite. The nonparametric function in (4.4) allows nontrivial interactions between $X_1$ and $X_j$ for $j \geq 2$, which cannot be modeled with SpAM. The sparsity of the function in (4.4) originates from $a_j(x_1)$ and there is no sparsity assumption on $f_j(x_j)$.

Example 4.2 shows that the ATLAS model is a generalization of varying coefficient additive model for functional data (Zhang and Wang, 2014). If $f_j(x_j)$'s are linear functions for all $j \geq 2$, we can write (4.4) as

$$f(x_1, \ldots, x_d) = f_1(x_1) + \sum_{j=2}^{d} a_j(x_1) x_j, \tag{4.5}$$

which is a high dimensional varying coefficient linear model, where the support of the linear coefficients may vary with $x_1$. Varying coefficient linear models in fixed dimension have been extensively studied Hastie and Tibshirani (1993); Fan and Zhang (1999); Berhane and Tibshirani (1998); Zhu et al. (2012), while Wei et al. (2011) study high dimensional varying coefficient linear models with fixed sparsity.

The locally additive assumption in (4.1) for the ATLAS model makes it possible for us to use the kernel-sieve hybrid estimator to estimate functions in $\mathcal{A}_d(s)$. The loss function for the kernel-sieve hybrid estimator in (2.6) has two parts: the kernel function makes the loss function only involve data points within the area $(z-h, z+h) \times \mathcal{X}^{d-1}$ and the sieve approximation part is therefore good enough to approximate the true function according to (4.1). In particular, let $(\widehat{\alpha}_z, \widehat{\boldsymbol{\beta}}_z)$ be the output of (2.7), we estimate the true functions $f_1(z)$ and $f_j(x_j, z)$ by

$$\widehat{f}_1(z) = \widehat{\alpha}_z \text{ and } \widehat{f}_j(x_j, z) = \sum_{k=1}^{m} \psi_{jk}(x_j)\widehat{\boldsymbol{\beta}}_{jk;z}, \text{ for } j = 2, \ldots, d.$$

We can thus estimate the bivariate charts $\{f_j(x_j, x_1)\}_{j=2}^d$ by "gluing" the local charts $\{f_j(x_j, z)\}_{j=1}^d$ over different longitudes $z \in \mathcal{X}$ through a fast algorithm proposed in Appendix A in the supplementary material. Moreover, we can also construct a confidence band for $f_1$ following procedure in Section 3.2.

If we weaken Assumption (**A5**) and generalize it to the assumption that $f(x_1, \ldots, x_d) \in \mathcal{A}_d(s)$, the estimation rates in Theorem 3.2 and the property of confidence band in Theorem 3.4 remain true. In fact, we will prove these theorems under the ATLAS model and apply them to SpAM.

# 5 Numerical Experiments

In this section, we study the finite sample properties of confidence bands for the ATLAS model and sparse additive model. We apply the SpAM to a genomic dataset and the ATLAS model to a fMRI dataset.

### 5.1 Synthetic Data

We consider two kinds of synthetic models. In the first example we evaluate the empirical properties of the bootstrap confidence band for sparse additive model. In the second example, we apply it to the ATLAS model.

In both examples, we use the quadratic kernel $K_q(u) = (15/16) \cdot (1-u^2)^2 \mathbb{1}(|u| < 1)$ as the kernel function in (2.7).

**Example 5.1.** We consider the sparse additive model $Y_i = \sum_{j=1}^{4} f_j(X_{ij}) + \varepsilon_i$, where

$$f_1(t) = 6\big(0.1\sin(2\pi t) + 0.2\cos(2\pi t) + 0.3(\sin(2\pi t))^2 + 0.4(\cos(2\pi t))^3 + 0.5(\sin(2\pi t))^3\big),$$
$$f_2(t) = 3(2t-1)^2, \quad f_3(t) = 5t, \quad f_4(t) = 4\sin(2\pi t)/(2-\sin(2\pi t)).$$

The model is considered by Zhang and Lin (2006), Meier et al. (2009), and Huang et al. (2010). Let $W_1, \ldots, W_d$ and $U$ follow i.i.d. Uniform$[0,1]$ and

$$X_j = \frac{W_j + tU}{1+t} \text{ for } j = 1, \ldots, d.$$

The data sample $X_{1j}, \ldots, X_{nj}$ are i.i.d. copies of $X_j$. The correlation between $X_j, X_{j'}$ is therefore $t^2/(1+t^2)$ for $j \neq j'$. We set $t = 0.3$. The noise $\{\varepsilon_i\}_{i=1}^n$ are i.i.d. $N(0, 1.5^2)$. Let the dimension $d = 600$ and the sample sizes $n \in \{400, 500, 600\}$. In the kernel-sieve hybrid estimator (2.7), we use the cubic B-splines with nine evenly distributed knots and $m = 5$. The parameter $\gamma$ in (2.15) is set to be $\gamma = 0.05 \log d \sqrt{m/nh}$. The tuning parameter $\lambda$ and bandwidth $h$ are chosen by cross validation according to the BIC criterion defined as

$$\text{BIC} = \log\left(\frac{\text{RSS}}{nh}\right) + \text{df} \cdot \frac{\log nh}{nh},$$

where RSS is the residual sums of squares and the degrees of freedom is defined as $\text{df} = \widehat{s} \cdot m$ with $\widehat{s}$ being the number of variables selected by the estimator. We aim to construct the confidence band for $f_1^*(t) = f_1(t) - \mathbb{E}[f_1(X_1)]$. In the simulation, we use the sample mean $\mathbb{E}_n[f_1(X_1)] := n^{-1}\sum_{i=1}^n f(X_{i1})$ to center $f_1(t)$. To test the coverage probability of confidence bands for inactive covariates, we also construct the confidence band for $f_5(t) = 0$. The empirical probability that the confidence bands cover the true function on the first 100 data points is computed based on 500 repetitions. The results are summarized in Figure 2 and Table 1.

**Example 5.2.** We generate data from the following ATLAS model

$$Y_i = a_1 f_1(X_{i1}) + \sum_{j=2}^{4} a_j(X_{i1}) f_j(X_{ij}) + \varepsilon_i,$$

where the additive functions are designed as follows

$$f_1(t) = -2\sin(2\pi t), \quad f_2(t) = t^2 - 1/3, \quad f_3(t) = t - 1/2, \quad f_4(t) = e^t + e^{-1} - 1;$$

$$a_1 \in \{0, 1\}, \quad a_2(t) = 2K_q(4t-1), \quad a_2(t) = 3\cos(2\pi t), \quad a_3(t) = 4.$$

Here two values of $a_1 \in \{0,1\}$ correspond to two scenarios that the true function is zero and nonzero. The noise $\epsilon_i \sim N(0, \sigma^2)$ for $i = 1, \ldots, n$ with $\sigma = 1.5$. This ATLAS model is constructed based on the synthetic example in Ravikumar et al. (2009) by adding $a_j(t)$'s according to Example 4.2. The covariates $X_{ij}$ are independently and identically generated from Uniform$[0,1]$ distributions for $i = 1, \ldots, n$ and $j = 1, \ldots, d$. It can be checked that this model follows the identifiability condition in (2.2). According to the argument in Example 4.2, the true function $f_1^*(t) = a_1 f_1(t)$ . We set the dimension of covariates to be $d = 600$ and consider three sample sizes $n \in \{400, 500, 600\}$. We again use the cubic B-spline basis with nine evenly distributed knots and $m = 5$. We again tune $\lambda$ and $h$ through cross validation by minimizing the BIC criterion. The confidence bands are constructed at the significance level 95% and the quantile estimator $\widehat{c}_n(\alpha)$ is computed by bootstrap with 500 repetitions. To measure the coverage probability of the confidence bands, we compute empirical probability that the confidence bands cover the true function on the first 100 data points based on 500 repetitions. The numerical results are reported in Figure 3 and Table 1.

## 5.2 Real Data

We apply the kernel-sieve estimator to two types real datasets: genomic dataset and neural imaging dataset. We aim to test our model's performance in variable selection and inferential analysis under real applications.

### 5.2.1 Genomic Data

We first consider the genomic dataset on the relation between gene and riboflavin (vitamin $B_2$) production with bacillus subtilis. Instead of evaluating the performance of variable selection in the previous neural imaging application, we aim to demonstrate the inference analysis of our method. The dataset is provided by DSM (Kaiseraugst, Switzerland) and it is publicly available in Supplementary Section A.1 of Bühlmann et al. (2014). The response variable $Y$ represents the logarithm of the riboflavin production rate. The covariates are the logarithm of gene expression levels with dimension $d = 4,088$ and sample size $n = 71$. van de Geer et al. (2014), Bühlmann et al. (2014) and Javanmard and Montanari (2014) use the linear model to find potentially significant genes. van de Geer et al. (2014) finds no significant genes, Bühlmann et al. (2014) finds the gene YXLD-at and Javanmard and Montanari (2014) finds two genes YXLD-at and YXLE-at to be significant. In this paper, we use the sparse additive model to find whether the two genes YXLD-at and YXLE-at are significant. We first normalize the covariates onto $[0, 1]$ and use (2.18) to construct confidence bands for the two genes YXLD-at and YXLE-at at significance level 95%. The results are illustrated in Figure 5. We can see that both genes have significantly nonzero effects. However, the gene YXLE-at has larger part of the domain where zero locates within the confidence band comparing to YXLD-at. Moreover, the magnitude of regressed function on YXLE-at is smaller than YXLD-at. These explain the reason why YXLE-at is less significant than YXLD-at in the previous analysis.

### 5.2.2 Neural Imaging Data

The second application we consider is the ADHD-200 dataset (Biswal et al., 2010) on the resting-state fMRI of 195 children and adolescents diagnosed with attention deficit hyperactive disorder (ADHD) as long as 491 typical developing controls. Among them, 246 individuals are measured by the ADHD index (Conners, 2008) which assesses the level of disorder. In order to explore the connection between ADHD and the brain activities, we aim to regress the ADHD index by the fMRI data of 264 voxels selected by Power et al. (2011) as the representative functional cerebral areas. Phenotypic information including age, gender and intelligence quotient (IQ) is also provided.

Several studies have revealed that the maturation of the brains for the youth with ADHD is delayed in some cortical regions, compared to the ones without disorder (Mann et al., 1992; El-Sayed et al., 2003; Shaw et al., 2007). For example, Shaw et al. (2007) find that the cortical development for the individuals with ADHD is significantly slower in the frontal lobe and temporal lobe. Therefore, the functioning voxels related to ADHD are varying with the age and the ATLAS model can characterize such variation unlike the sparse additive model. We set the age as the longitude variable and the fMRI of 264 voxels as the other covariates. All the covariates are normalized to $[0, 1]$. Each of 246 subjects with ADHD indices has 76 to 276 scans and all the scans are treated as independent observation.

The results of regression are illustrated in Figure 6 and Figure 7. We show the first eight estimated surfaces with largest maximum norms among $\{\widehat{f}_j(x_j, x_1)\}_{j=1}^d$ in Figure 6. In the middle of eight surfaces in Figure 6, we demonstrate all voxels being activated (nonzero) at certain time by small balls. The radius of a ball represents the length of time the corresponding voxel is activated and the maximum norm is represented by the ball's color where red means the largest values and yellow means the smallest (see the colorbar on the right bottom of Figure 6). We can see that most of voxels with strongest signal strength are in the frontal and temporal lobes which matches the results in Shaw et al. (2007). Moreover, the different flat zero areas of different surfaces in Figure 6 imply that the voxels are not activated simultaneously, which supports the necessity of ATLAS model. In Figure 7, we show the activated voxels at different ages. The radii and colors of the balls are same as Figure 6. We observe that, with the increasing age, the number of activated voxels first ascends and then reduces. This is similar to the results in Shaw et al. (2007) showing that 50% cortical points of ADHD groups attain peak thickness around the age of 10.5 years. The decreasing number of activated voxels after age 15 is also congruent with the discovery in Shaw et al. (2007).

# 6 Discussion

In this paper, we consider a novel nonparametric model, ATLAS, which is a generalization of the sparse additive model. ATLAS naturally models high-dimensional nonparametric functions having different sparsity in different local regions of the domain. We consider the kernel-sieve hybrid regression to estimate the unknown function. Since we consider functions in the 2nd order Hölder class, only Nadaraya-Watson-type kernel estimator is considered. However, it is not hard to generalize the loss function in (2.6) to local polynomial regression

$$\mathcal{L}_z(\alpha, \boldsymbol{\beta}) = \frac{1}{n} \sum_{i=1}^{n} K_h(X_{i1} - z)\Big(Y_i - \bar{Y} - \alpha - \sum_{\ell=1}^{p} \frac{(X_{i1} - z)^{\ell}}{\ell!} - \sum_{j=2}^{d} \sum_{k=1}^{m} \psi_{jk}(X_{ij})\boldsymbol{\beta}_{jk}\Big)^2.$$

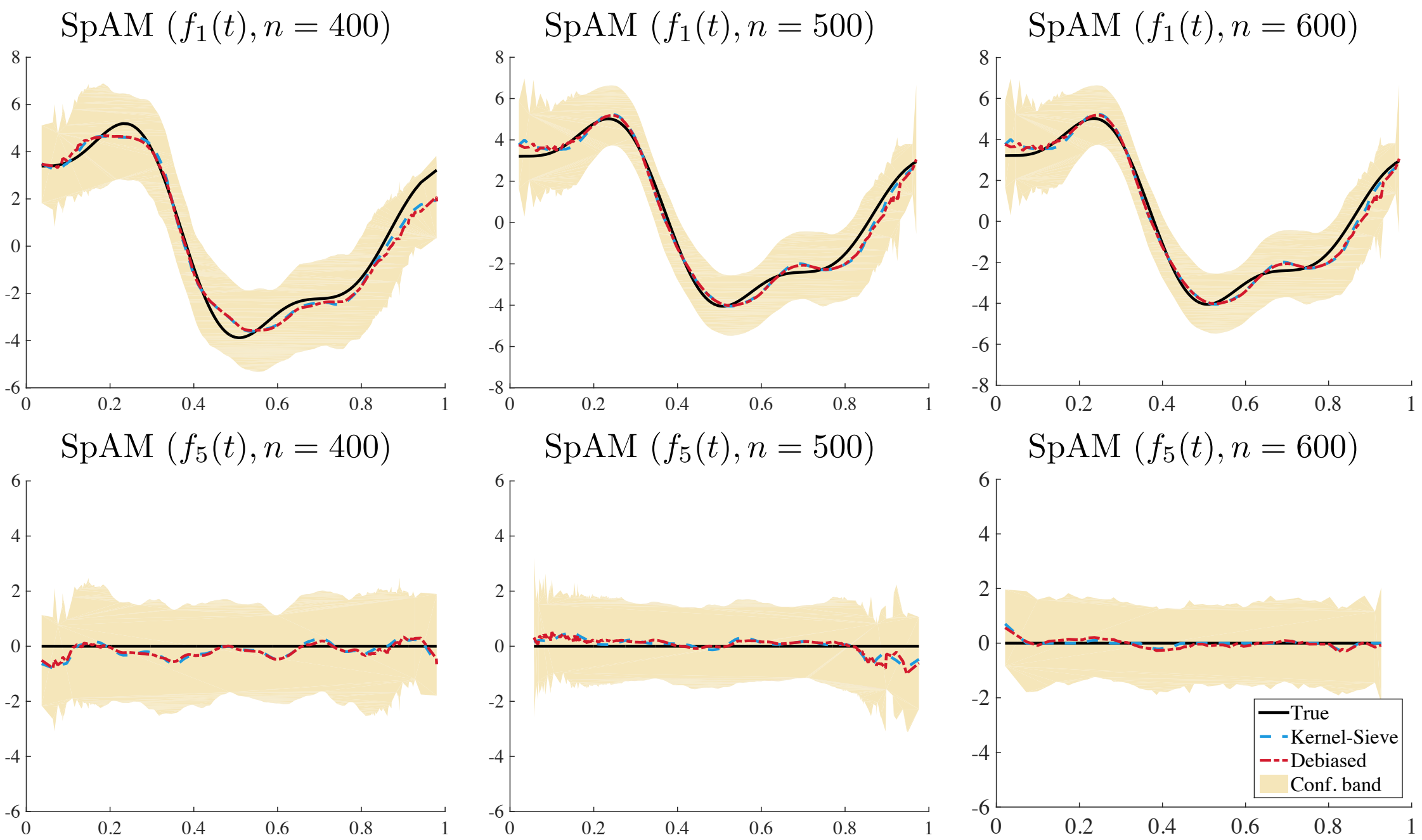

Figure 2: Kernel-sieve hybrid estimators for the $d = 600$ dimensional SpAM model $Y = \sum_{j=1}^{4} f_j(X_j) + \varepsilon$, for $n = 400, 500, 600$ and the noise $\varepsilon \sim N(0, 1.5^2)$. The confidence bands at significant level 95% cover $f_1(t)$ on the first row and $f_5(t) = 0$ on the second row.

We can apply a similar proof technique to show the statistical rate of the estimator based on the generalized loss in higher order Hölder classes. Corresponding methods to construct confidence bands can also be applied.

# 7 Proof of the Statistical Rate of Kernel-Sieve Hybrid Estimator

For all the proofs in the following of the paper (including the supplementary material), we consider the most general case that true nonparametric function $f(x_1, \ldots, x_d)$ belongs to the ATLAS model $\mathcal{A}_d(s)$. Since SpAM is a strictly smaller family of $\mathcal{A}_d(s)$, all the proofs apply to $\mathcal{K}_d(s)$ as well.

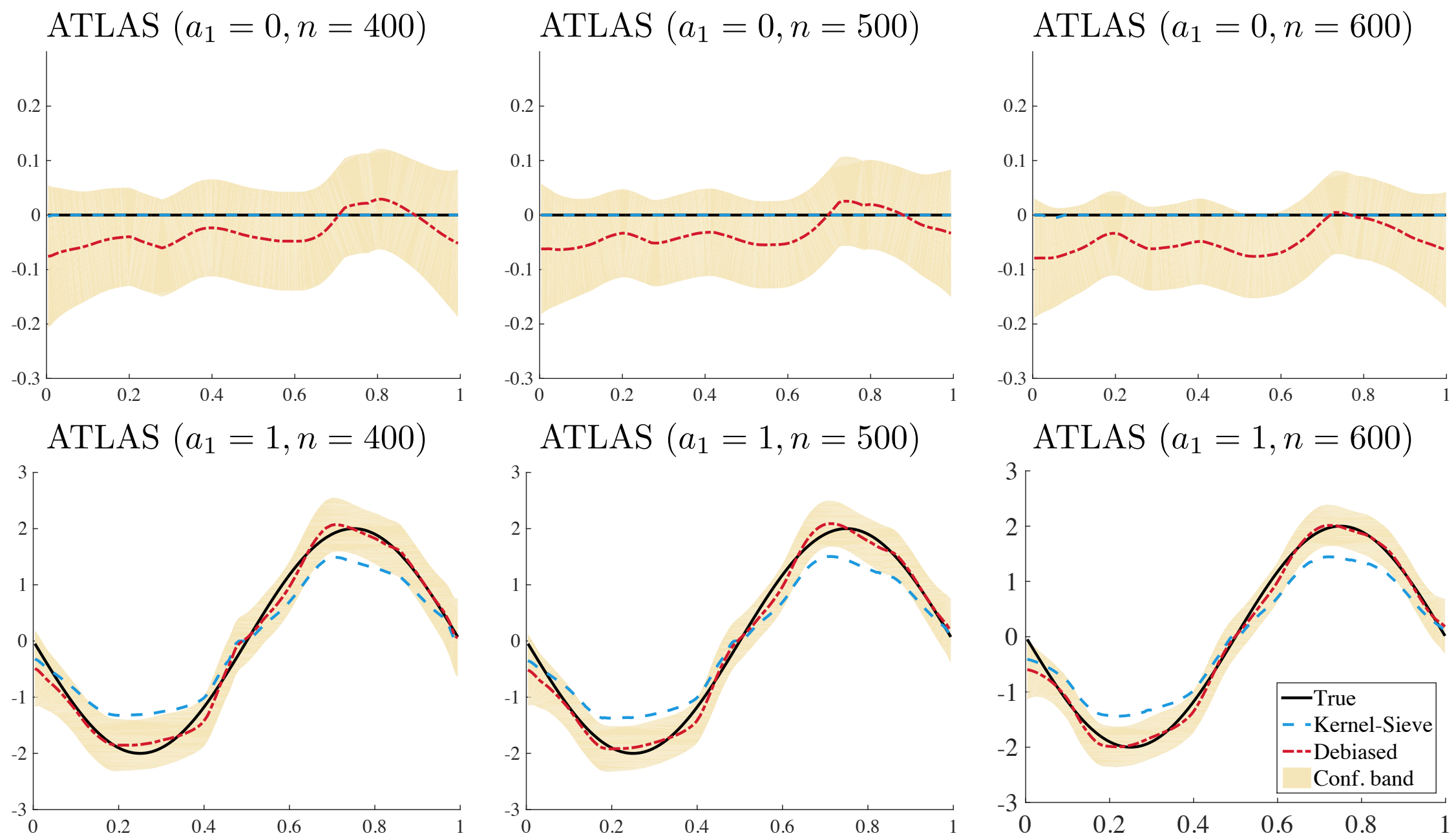

Figure 3: Kernel-sieve hybrid estimators for the $d = 600$ dimensional ATLAS model $Y = a_1 f_1(X_1) + \sum_{j=2}^{4} a_j(X_1) f_j(X_j) + \varepsilon$, for $n = 400, 500, 600$ and the noise $\varepsilon \sim N(0, 1.5^2)$. The confidence bands at significant level 95% cover $f_1^* = a_1 f_1$ for $a_1 \in \{0, 1\}$ respectively.

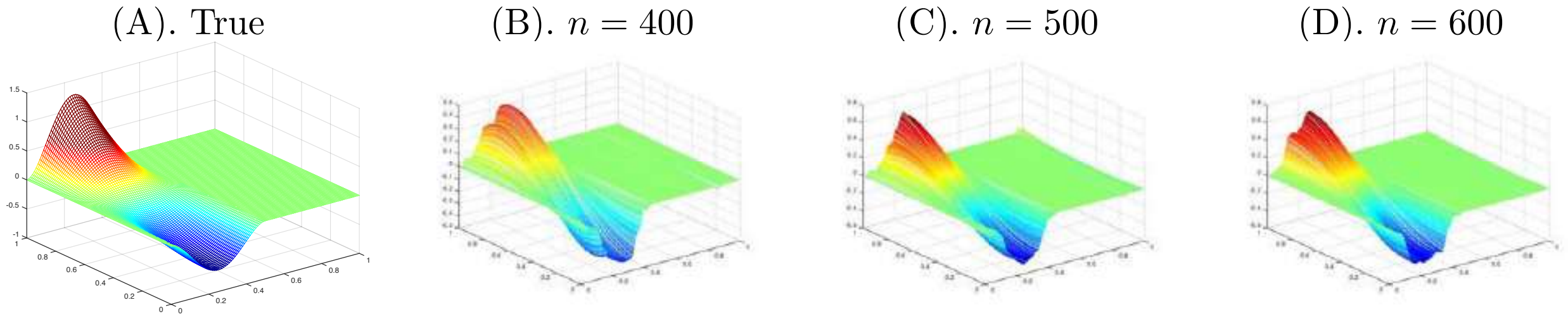

Figure 4: Kernel-sieve hybrid estimators for the two dimensional surface $a_2(x_1) f_2(x_2)$.

This section outlines the proof of Theorem 3.2 on the statistical estimation rate of the kernel-sieve hybrid estimator in (2.7). Before presenting the main proof, we list several technical lemmas whose proofs are deferred to Appendix D in the supplementary material.

The following lemma provides the restricted eigenvalue condition on the empirical Hessian matrix of the kernel-sieve hybrid loss in (2.6), which is $\widehat{\mathbf{\Sigma}}_z = n^{-1} \mathbf{\Psi} \mathbf{W}_z \mathbf{\Psi}^T$.

**Lemma 7.1.** Under Assumptions (**A1**)-(**A5**), suppose $\boldsymbol{\beta} \in \mathbb{R}^{(d-1)m}$ and $\alpha \in \mathbb{R}$ satisfy the cone

| $n$ | Method | Zero function | | Non-zero function | |
|---|---|---|---|---|---|
| | | Coverage probability | Area | Coverage probability | Area |
| 400 | SpAM | 0.882 | 0.336 | 0.916 | 0.119 |
| | ATLAS | 0.954 | 0.375 | 0.936 | 0.129 |
| 500 | SpAM | 0.888 | 0.310 | 0.932 | 0.110 |
| | ATLAS | 0.958 | 0.346 | 0.950 | 0.118 |
| 600 | SpAM | 0.896 | 0.290 | 0.936 | 0.102 |
| | ATLAS | 0.960 | 0.324 | 0.952 | 0.110 |

Table 1: Comparison of coverage probability for confidence bands at significant level 95% for the zero function $f_5$ and non-zero function $f_1$ in SpAM model $Y = \sum_{j=1}^4 f_j(X_j) + \varepsilon$ as long as the zero function $a_1 f_1$ for $a_1 = 1$ and non-zero function $a_1 f_1$ for $a_1 = 0$ in the ATLAS model $Y = a_1 f_1(X_1) + \sum_{j=2}^4 a_j(X_1) f_j(X_j) + \varepsilon$. Here we set dimension $d = 600$, sample size $n = 400, 500, 600$ and $\varepsilon \sim N(0, 1.5^2)$.

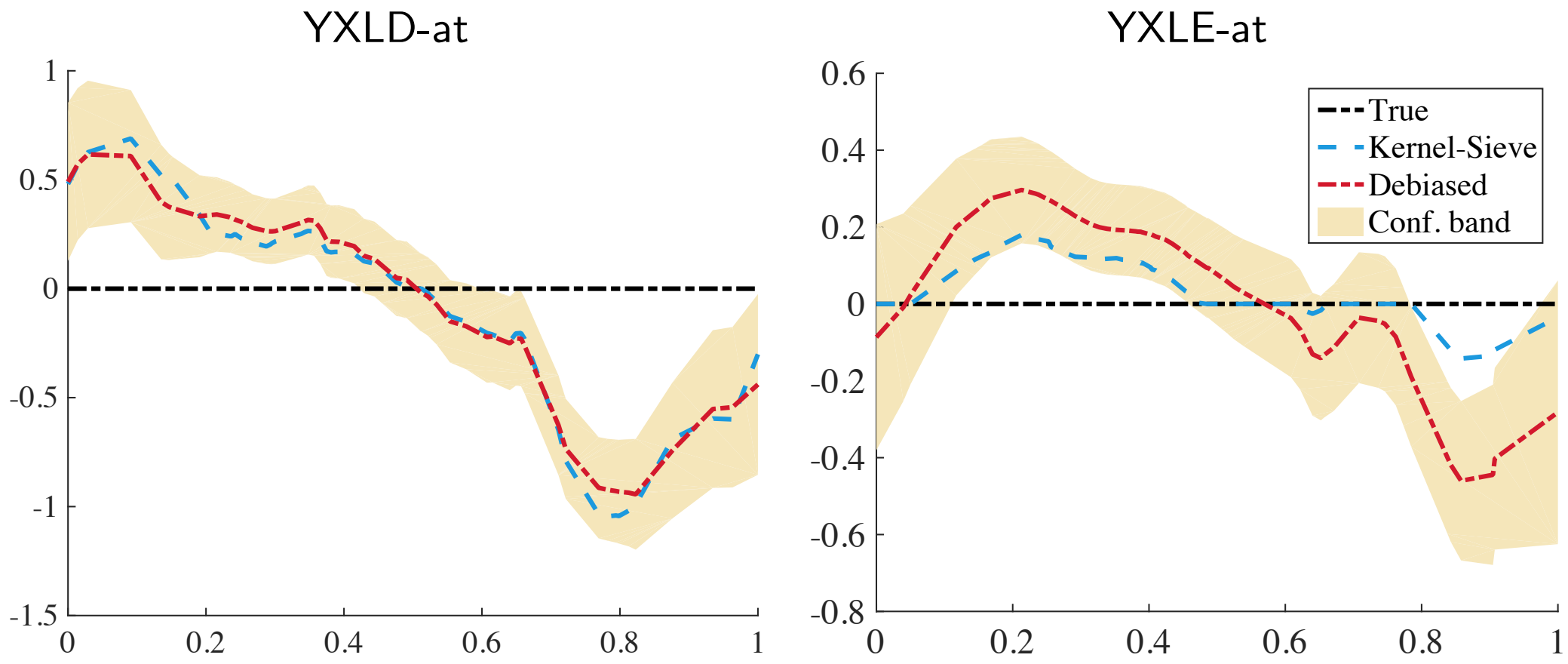

Figure 5: Kernel-sieve hybrid estimators for the riboflavin dataset using ATLAS model.

restriction

$$\sum_{j \in S^c} \|\boldsymbol{\beta}_j\|_2 \le 3 \sum_{j \in S} \|\boldsymbol{\beta}_j\|_2 + 3\sqrt{m}|\alpha|$$

for some index set $S \subset [d]$ with cardinality $s$. Denote $\boldsymbol{\theta} = (\alpha, \boldsymbol{\beta}^T)^T$. If $s\sqrt{m^3 \log(dm)/(nh)} + sm^2/(nh) = o(1)$, there exists a constant $\rho_{\min}$ such that with high probability,

$$\inf_{z \in \mathcal{X}} \boldsymbol{\theta}^T \widehat{\boldsymbol{\Sigma}}_z \boldsymbol{\theta} \ge \frac{\rho_{\min}}{2m} \|\boldsymbol{\beta}\|_2^2 + \frac{\rho_{\min}}{2} |\alpha|_2^2.$$

The estimation error for the kernel-sieve hybrid estimator comes from three sources: (1) noise $\varepsilon$,

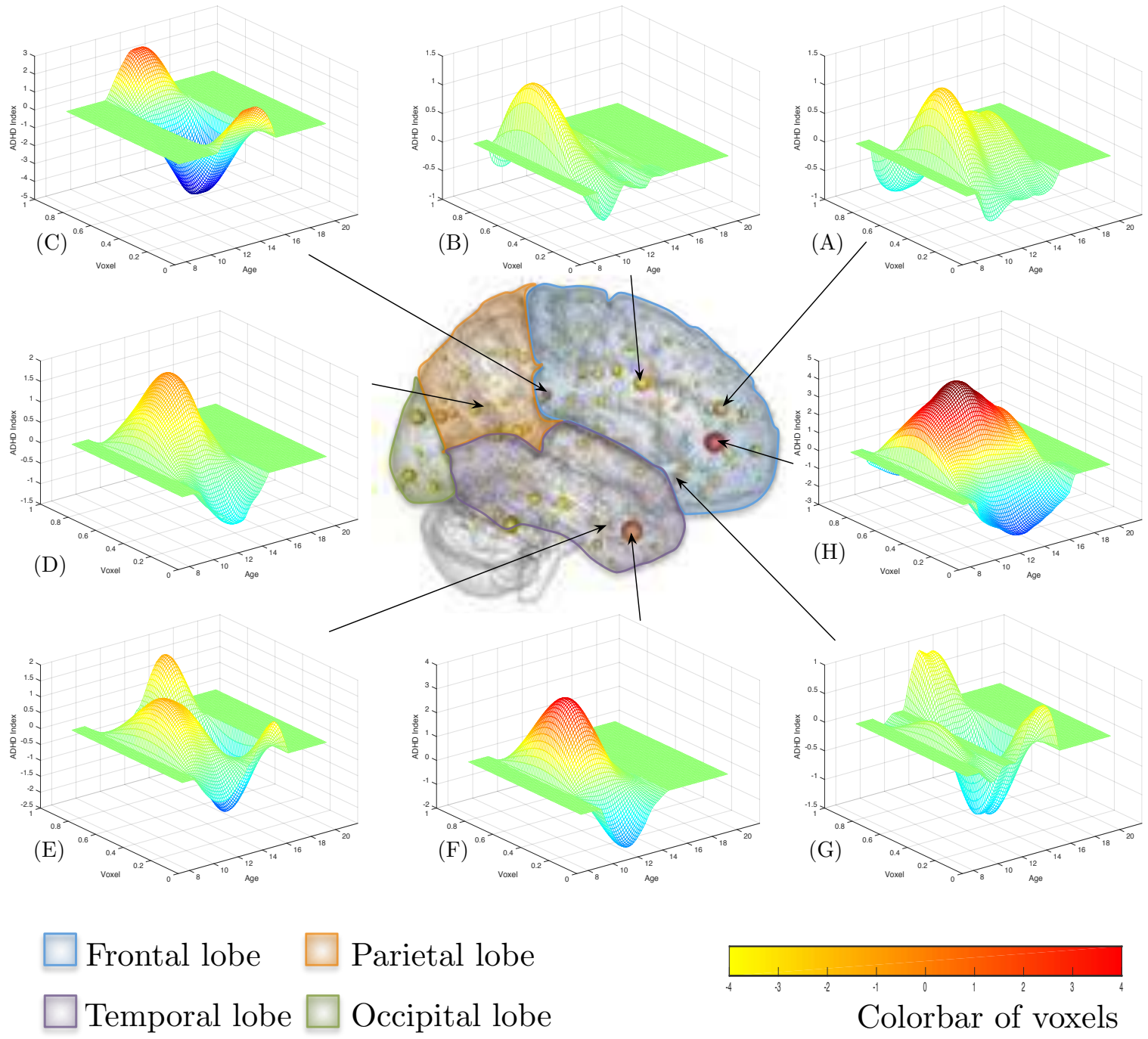

Figure 6: The estimated surfaces of first eight voxels with largest maximum norms. The radii of the balls in the brain represent the duration the voxels being active and the colors represent the maximum norms of the surfaces, whose corresponding values are indicated by the colorbar on the right bottom of the figure.

(2) approximation error by finite B-spline bases, and (3) approximation error by $s$ local additive functions to the true function. The following lemma provides the rate for the B-spline approximation error, which further illustrates how the number of B-spline basis functions $m$ influence the rate.

**Lemma 7.2.** Recall that $\{f_{jz}\}_{j=1}^d$ are defined in Definition 4.1. Let $\boldsymbol{\delta}_z = (\delta_1(z), \ldots, \delta_n(z))^T$ where $\delta_i(z) = \sum_{j=2}^d f_{jz}(X_{ji}) - f_{mj}(X_{ji})$ for $i = 1, \ldots n$, where $f_{mj}(\cdot)$ is defined in (2.4). Under Assumptions (**A1**)-(**A5**) there exists a constant $C > 0$ such that the following three inequalities hold with probability at least $1 - 1/n$,

$$\sup_{z \in \mathcal{X}} \max_{j \geq 2} \frac{1}{n} \|\boldsymbol{\Psi}_{\bullet j}^T \mathbf{W}_z \boldsymbol{\delta}_z\|_2 \leq C\sqrt{s} \cdot m^{-5/2}, \tag{7.1}$$

$$\sup_{z \in \mathcal{X}} \frac{1}{n} |\boldsymbol{\Psi}_{\bullet 1}^T \mathbf{W}_z \boldsymbol{\delta}_z| \leq C\sqrt{s} \cdot m^{-2}, \tag{7.2}$$

$$\sup_{z \in \mathcal{X}} \frac{1}{n} \|\mathbf{W}_z^{1/2} \boldsymbol{\delta}_z\|_2^2 \leq Csm^{-4}. \tag{7.3}$$

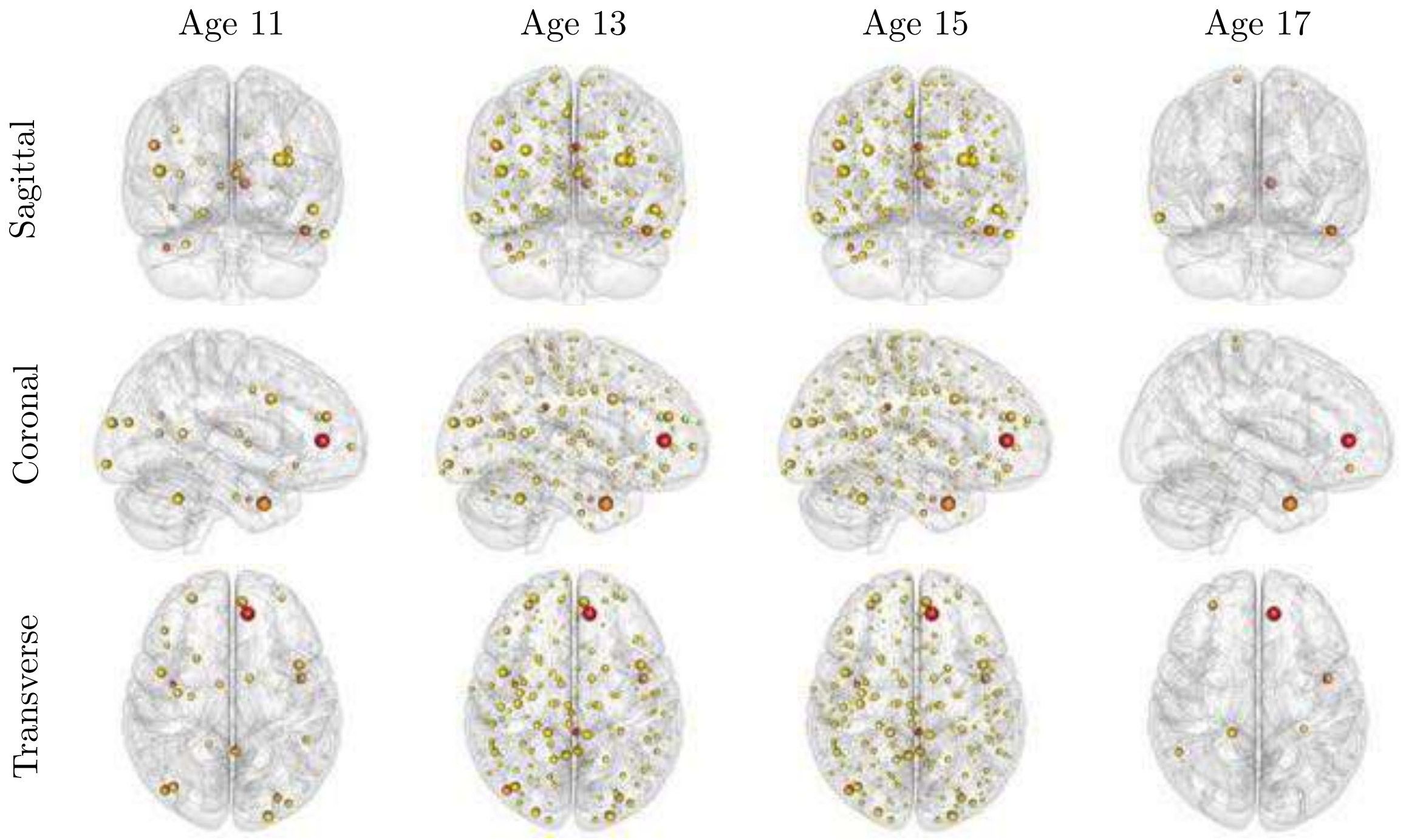

Figure 7: Active voxels varying with age. Each column shows the active voxels at each age. The radii and colors of the balls in a brain represent the duration and maximum norms of the active voxels as in Figure 6.

Our next lemma bounds the approximation error of charts under the ATLAS model (4.1). We can see that both the number of bases $m$ and the bandwidth $h$ play a role in the estimation.

**Lemma 7.3.** Let $\boldsymbol{\xi}_z = (\xi_1(z), \ldots, \xi_n(z))^T$ and $\boldsymbol{\zeta}_z = (\zeta_1(z), \ldots, \zeta_n(z))^T$, where $\xi_i(z) = f_1(X_{1i}) - f_1(z)$ and $\zeta_i(z) = f(X_{1i}, \ldots, X_{di}) - \sum_{j=1}^d f_{jz}(X_{ji})$ for $i \in [n]$. Under Assumptions (**A1**)-(**A5**) there exists a constant $C > 0$ such that the following three inequalities hold with probability at least $1 - 1/n$,

$$\sup_{z\in\mathcal{X}} \max_{j\in[d]} \frac{1}{n} \|\boldsymbol{\Psi}_{\bullet j}^T \mathbf{W}_z(\boldsymbol{\xi}_z + \boldsymbol{\zeta}_z)\|_2 \le C \left( \sqrt{\frac{h\log(dh^{-1})}{n}} + \frac{m^{3/2}\log(dh^{-1})}{n} + \frac{h^2}{\sqrt{m}} \right),$$

$$\sup_{z\in\mathcal{X}} \frac{1}{n} |\boldsymbol{\Psi}_{\bullet 1}^T \mathbf{W}_z(\boldsymbol{\xi}_z + \boldsymbol{\zeta}_z)| \le C \left( h^2 + \sqrt{h/n} \right),$$

$$\sup_{z\in\mathcal{X}} \frac{1}{n} \|\mathbf{W}_z^{1/2}(\boldsymbol{\xi}_z + \boldsymbol{\zeta}_z)\|_2^2 \le Ch^2.$$

The following lemma quantifies the statistical error arising from the noise $\varepsilon$.

**Lemma 7.4.** Let $T_n = \sup_{z\in\mathcal{X}} \max_{j\in[d]} n^{-1} \|\boldsymbol{\Psi}_{\bullet j}^T \mathbf{W}_z \boldsymbol{\varepsilon}\|_2$, where $\boldsymbol{\varepsilon} = (\varepsilon_1, \ldots, \varepsilon_n)^T$. Under Assump-

tions (**A1**)-(**A5**) and if $m(nh)^{-1} = o(1)$, there exists a constant $C > 0$ such that with probability at least $1 - 1/n$,

$$T_n \le C\sqrt{\log(dm^2h^{-2})/(nh)}.$$

We are now ready to present the main proof of Theorem 3.2.

*Proof of Theorem 3.2.* We denote $\boldsymbol{\eta}_z = \boldsymbol{\varepsilon} + \boldsymbol{\delta}_z + \boldsymbol{\xi}_z + \boldsymbol{\zeta}_z$ and define the event

$$\mathcal{E} = \left\{\sup_{z\in\mathcal{X}} \max_{j\ge 2} \frac{4}{n}\|\boldsymbol{\Psi}_{\bullet j}^T \mathbf{W}_z \boldsymbol{\eta}_z\|_2 \le \lambda\right\} \bigcup \left\{\sup_{z\in\mathcal{X}} \frac{4}{n}|\boldsymbol{\Psi}_{\bullet 1}^T \mathbf{W}_z \boldsymbol{\eta}_z|_2 \le \lambda\sqrt{m}\right\}.$$

Using Lemma 7.2, Lemma 7.3 and Lemma 7.4, there exist constants $c, C$ such that $\mathbb{P}(\mathcal{E}) \ge 1 - c/n$ if the tuning parameter satisfies the following inequality

$$\lambda \ge C\left(\sqrt{\frac{\log(dm^2h^{-2})}{nh}} + \sqrt{s}\cdot m^{-5/2} + \sqrt{\frac{h\log(dh^{-1})}{n}} + \frac{m^{3/2}\log(dh^{-1})}{n} + \frac{h^2}{\sqrt{m}}\right). \tag{7.4}$$

In the rest of this proof, we are always conditioning on the event $\mathcal{E}$.

Denote $S_z := \{j \in \{2, \ldots, d\} \,|\, f_{jz} \not\equiv 0\}$ and $\boldsymbol{\Delta} = \widehat{\boldsymbol{\beta}}_+ - \boldsymbol{\beta}_+$, where $\boldsymbol{\Delta}_1 = \widehat{\alpha}_z - f_1(z)$ and $\boldsymbol{\Delta}_j = \widehat{\boldsymbol{\beta}}_j - \boldsymbol{\beta}_j$ for $j \ge 2$. We start by showing that $\boldsymbol{\Delta}$ falls into the cone

$$\mathcal{A}_z := \left\{\boldsymbol{\Delta} : \textstyle\sum_{j\in S_z^c} \|\boldsymbol{\Delta}_j\|_2 \le 3\sum_{j\in S_z} \|\boldsymbol{\Delta}_j\|_2 + 3\sqrt{m}|\boldsymbol{\Delta}_1|\right\}.$$

Since $\widehat{\boldsymbol{\beta}}_+$ is a minimizer of the objective function,

$$\frac{1}{n}\|\mathbf{W}_z^{1/2}(\boldsymbol{Y} - \boldsymbol{\Psi}\widehat{\boldsymbol{\beta}}_+)\|_2^2 - \frac{1}{n}\|\mathbf{W}_z^{1/2}(\boldsymbol{Y} - \boldsymbol{\Psi}\boldsymbol{\beta}_+)\|_2^2 + \lambda\|\widehat{\boldsymbol{\beta}}\|_{1,2} - \lambda\|\boldsymbol{\beta}\|_{1,2} + \lambda\sqrt{m}(|\widehat{\alpha}_z| - |f_1(z)|) \le 0.$$

On the event $\mathcal{E}$, we have the following inequality

$$\begin{aligned}\sup_{z\in\mathcal{X}} \frac{4}{n}\boldsymbol{\eta}_z^T \mathbf{W}_z \boldsymbol{\Psi}\boldsymbol{\Delta} &\le \sup_{z\in\mathcal{X}} \max_{j\ge 2} \frac{4}{n}\|\boldsymbol{\Psi}_{\bullet j}^T \mathbf{W}_z \boldsymbol{\eta}_z\|_2 \|\boldsymbol{\Delta}_{2:d}\|_{1,2} + \sup_{z\in\mathcal{X}} \frac{4}{n}\|\boldsymbol{\Psi}_{\bullet 1}^T \mathbf{W}_z \boldsymbol{\eta}_z\|_2 |\boldsymbol{\Delta}_1| \\ &\le \lambda\|\boldsymbol{\Delta}_{2:d}\|_{1,2} + \lambda\sqrt{m}|\boldsymbol{\Delta}_1|.\end{aligned}$$

The first inequality is due to the Hölder's inequality and the second one is by the definition of $\mathcal{E}$.

Furthermore, we derive the following inequality

$$\begin{aligned}
\frac{1}{n}\|\mathbf{W}_z^{1/2}\boldsymbol{\Psi}\boldsymbol{\Delta}\|_2^2 &\leq \frac{2}{n}\boldsymbol{\eta}_z^T\mathbf{W}_z\boldsymbol{\Psi}\boldsymbol{\Delta} - \lambda\sum_{j=2}^{d}(\|\widehat{\boldsymbol{\beta}}_j\| - \|\boldsymbol{\beta}_j\|) - \lambda\sqrt{m}(|\widehat{\alpha}_z| - |f_1(z)|) \\
&\leq \frac{\lambda}{2}\|\widehat{\boldsymbol{\beta}} - \boldsymbol{\beta}\|_{1,2} + \lambda\sqrt{m}|\widehat{\alpha}_z - f_1(z)| - \lambda\sum_{j=2}^{d}(\|\widehat{\boldsymbol{\beta}}_j\| - \|\boldsymbol{\beta}_j\|) \\
&\leq \frac{3\lambda}{2}\sum_{j\in S_z}\|\boldsymbol{\Delta}_j\| + \frac{3\lambda}{2}\sqrt{m}|\boldsymbol{\Delta}_1| - \frac{\lambda}{2}\sum_{j\in S_z^c}\|\boldsymbol{\Delta}_j\|.
\end{aligned}$$

The last inequality shows that $\boldsymbol{\Delta} \in \mathcal{A}_z$.

Next, we prove the rate of convergence by contradiction. Suppose that for some fixed $t$, which will be specified later, we have

$$\exists z \in \mathcal{X}, \quad \frac{1}{\sqrt{n}}\|\mathbf{W}_z^{1/2}\boldsymbol{\Psi}\boldsymbol{\Delta}\| > t. \tag{7.5}$$

Equation (7.5) implies that there exists some $z \in \mathcal{X}$ such that

$$0 > \min_{\boldsymbol{\Delta}\in\mathcal{A}_z, \|\widehat{\boldsymbol{\Sigma}}_z^{1/2}\boldsymbol{\Delta}\|\geq t} \frac{1}{n}\|\mathbf{W}_z^{1/2}(\boldsymbol{Y} - \boldsymbol{\Psi}\widehat{\boldsymbol{\beta}}_+)\|_2^2 - \frac{1}{n}\|\mathbf{W}_z^{1/2}(\boldsymbol{Y} - \boldsymbol{\Psi}\boldsymbol{\beta}_+)\|_2^2 + \lambda\|\widehat{\boldsymbol{\beta}}_+\|_{1,2} - \lambda\|\boldsymbol{\beta}_+\|_{1,2}.$$

Using the fact that $\mathcal{A}_z$ is a cone, we can replace the constraint $\|\widehat{\boldsymbol{\Sigma}}_z^{1/2}\boldsymbol{\Delta}\| \geq t$ by $\|\widehat{\boldsymbol{\Sigma}}_z^{1/2}\boldsymbol{\Delta}\| = t$ and the above inequality still preserves. Combining the event $\mathcal{E}$, we have

$$\begin{aligned}
0 &> \min_{\boldsymbol{\Delta}\in\mathcal{A}_z, \|\widehat{\boldsymbol{\Sigma}}_z^{1/2}\boldsymbol{\Delta}\|=t} \frac{1}{n}\|\mathbf{W}_z^{1/2}(\boldsymbol{Y} - \boldsymbol{\Psi}\widehat{\boldsymbol{\beta}}_+)\|_2^2 - \frac{1}{n}\|\mathbf{W}_z^{1/2}(\boldsymbol{Y} - \boldsymbol{\Psi}\boldsymbol{\beta}_+)\|_2^2 + \lambda\mathcal{R}(\widehat{\boldsymbol{\beta}}_+) - \lambda\mathcal{R}(\boldsymbol{\beta}_+) \\
&\geq \min_{\boldsymbol{\Delta}\in\mathcal{A}_z, \|\widehat{\boldsymbol{\Sigma}}_z^{1/2}\boldsymbol{\Delta}\|=t} \frac{1}{n}\|\mathbf{W}_z^{1/2}\boldsymbol{\Psi}\boldsymbol{\Delta}\|_2^2 - 2\lambda\mathcal{R}(\boldsymbol{\Delta}) + \lambda\mathcal{R}(\widehat{\boldsymbol{\beta}}_+) - \lambda\mathcal{R}(\boldsymbol{\beta}_+).
\end{aligned} \tag{7.6}$$

From Lemma 7.1, we can bound the R.H.S. by

$$\begin{aligned}
2\mathcal{R}(\boldsymbol{\Delta}) - \mathcal{R}(\widehat{\boldsymbol{\beta}}_+) + \mathcal{R}(\boldsymbol{\beta}_+) &\leq 3\sum_{j\in S}\|\boldsymbol{\Delta}_j\| + 3\sqrt{m}|\boldsymbol{\Delta}_1| \\
&\leq 3\sqrt{s}\,\|\boldsymbol{\Delta}_{S\cup\widehat{S}}\|_2 + 3\sqrt{m}|\boldsymbol{\Delta}_1| \\
&\leq 6\sqrt{2sm/\rho_{\min}}\,\|\widehat{\boldsymbol{\Sigma}}_z^{1/2}\boldsymbol{\Delta}\|_2.
\end{aligned} \tag{7.7}$$

Combining (7.6) and (7.7), we get a quadratic inequality

$$0 > t^2 - \frac{2\lambda\sqrt{sm}}{\rho_{\min}} t. \tag{7.8}$$

Setting $t = 2\sqrt{sm/\rho_{\min}} \cdot \lambda$, we obtain from (7.8) that $0 > t^2 - \left[2\lambda\sqrt{sm/\rho_{\min}}\right] t = 0$, which is a contradiction. Therefore, $\sup_{z\in\mathcal{X}} n^{-1/2}\|\mathbf{W}_z^{1/2}\boldsymbol{\Psi}\boldsymbol{\Delta}\|_2 \leq 2\lambda\sqrt{sm}/\rho_{\min}$. Using the rate for $\lambda$ in (7.4) and $h = o(1)$, we have

$$\sup_{z\in\mathcal{X}} \frac{1}{\sqrt{n}}\|\mathbf{W}_z^{1/2}\boldsymbol{\Psi}\boldsymbol{\Delta}\|_2 \leq C\sqrt{sm}\left(\sqrt{\frac{\log(dmh^{-1})}{nh}} + \frac{\sqrt{s}}{m^{5/2}} + \frac{m^{3/2}\log(dh^{-1})}{n} + \frac{h^2}{\sqrt{m}}\right). \tag{7.9}$$

Now, using Lemma 7.1, since $\boldsymbol{\Delta} \in \mathcal{A}_z$, we have that

$$\|\boldsymbol{\Delta}_{2:d}\|_{1,2} + \sqrt{m}|\boldsymbol{\Delta}_1| \leq \sqrt{s}\|\boldsymbol{\Delta}_{2:d}\|_2 + \sqrt{m}|\boldsymbol{\Delta}_1| \leq \sqrt{sm/(\rho_{\min} n)}\|\mathbf{W}_z^{1/2}\boldsymbol{\Psi}\boldsymbol{\Delta}\|_2$$

for any $z \in \mathcal{X}$, which leads to the following inequality

$$\sup_{z\in\mathcal{X}} \sqrt{m}|\widehat{\alpha}_z - f_1(z)| + \|\widehat{\boldsymbol{\beta}} - \boldsymbol{\beta}\|_{1,2} \leq Csm\left(\sqrt{\frac{\log(dmh^{-1})}{nh}} + \frac{\sqrt{s}}{m^{5/2}} + \frac{m^{3/2}\log(dh^{-1})}{n} + \frac{h^2}{\sqrt{m}}\right),$$

with probability at least $1 - c/n$. To obtain the best rate on the right hand side of the equation, we choose $h \asymp n^{-1/6}$ and $m \asymp n^{1/6}$ to obtain

$$\sup_{z\in\mathcal{X}} \left\{\sqrt{m}|\widehat{a}_z - f_1(z)| + \sum_{i=2}^{d} \|\widehat{\boldsymbol{\beta}}_j - \boldsymbol{\beta}_j\|_2\right\} = O_P\left(\log(dn)n^{-1/4}\right).$$

According to Corollary 15 in Chapter XI of de Boor (2001), given a function $g(x) = \sum_{k=1}^m \beta_k\phi_k(x)$, we have

$$\|g\|_2^2 \asymp m^{-1}\sum_{k=1}^{m} \beta_k^2. \tag{7.10}$$

Therefore, we have $\|\widehat{f} - f\|_2 \leq \rho_{\min}^{-1} s\sqrt{m}\lambda$ and, when $h \asymp n^{-1/6}$ and $m \asymp n^{1/6}$, the rate becomes

$$\|\widehat{f} - f\|_2^2 = O_P\left(n^{-2/3}\log(dn)\right).$$

This completes the proof. □

**Acknowledgement**

The authors are grateful for the support of NSF CAREER Award DMS1454377, NSF IIS1408910, NSF IIS1332109, NIH R01MH102339, NIH R01GM083084, and NIH R01HG06841. This work is also supported by an IBM Corporation Faculty Research Fund at the University of Chicago Booth School of Business. This work was completed in part with resources provided by the University of Chicago Research Computing Center.

*Supplementary material to*

# Kernel Meets Sieve: Post-Regularization Confidence Bands for Sparse Additive Model

Junwei Lu* Mladen Kolar† Han Liu‡

**Abstract**

This document contains the supplementary material to the paper "Kernel Meets Sieve: Post-Regularization Confidence Bands for Sparse Additive Model". All the proofs in the supplementary material assume that true nonparametric function $f(x_1, \dots, x_d)$ belongs to the ATLAS model $\mathcal{A}_d(s)$. In Appendix A, we introduce an accelerated method to derive our estimator. Appendix B proves the validity of bootstrap confidence bands. In Appendix C, we justify the assumptions required in the paper. Appendix D collects the technical lemmas on the estimation rate. Appendix E states some auxiliary results on the bootstrap confidence bands. In Appendix F, we lists several useful results on empirical processes.

## A Accelerated Algorithm

This section presents details of our method to accelerate Algorithm 1. To estimate $f_1$, we need to compute the estimator $\widehat{\alpha}_z$ for a number of $z$ values $z \in \{z_1, \dots, z_M\}$. A naïve approach is to run Algorithm 1 $M$ times, once for each value of $z$'s. We provide a more efficient algorithm which significantly reduces the computational cost. From Algorithm 1 and (2.14), the most expensive operation is evaluation of the gradient

$$\nabla_j \mathcal{L}_z(\boldsymbol{\beta}_+^{(t)}) = -\frac{1}{n} \boldsymbol{\Psi}_{\bullet j}^T \mathbf{W}_z \left( \boldsymbol{Y} - \boldsymbol{\Psi} \boldsymbol{\beta}_+^{(t)} \right). \tag{A.1}$$

Computing $\nabla_j \mathcal{L}_z(\boldsymbol{\beta}_+^{(t)})$ for a single $z$ requires $O(dm^2 n)$ flops. If we trivially repeat the computation for $M$ different $z$'s, the computational complexity is $O(dm^2 nM)$ which is challenging when $M$ is large. However, we can exploit the structure of $\nabla_j \mathcal{L}_z(\boldsymbol{\beta}_+^{(t)})$ to reduce the computational complexity. According to (A.1) and the fact that $\psi_{jk}(X_{i1}) = \phi_k(X_{i1}) - \bar{\phi}_{jk}(z)$, the $k$-th coordinate of $\nabla_j \mathcal{L}_z(\boldsymbol{\beta}_+^{(t)})$ has a formulation

$$
\begin{aligned}
\Big(\nabla_j \mathcal{L}_z(\boldsymbol{\beta}_+^{(t)})\Big)_k = & -\frac{1}{n}\sum_{i\in[n]} K_h(X_{i1}-z)\phi_k(X_{i1})Y_i + \bar{\phi}_{jk}(z)\cdot\frac{1}{n}\sum_{i\in[n]} K_h(X_{i1}-z)Y_i \\
& + \sum_{\ell\in[d],s\in[m]} \boldsymbol{\beta}_{\ell s}^{(t)}\bigg\{\frac{1}{n}\sum_{i=1}^{n} K_h(X_{i1}-z)\big[\phi_k(X_{i1})-\bar{\phi}_{jk}(z)\big]\big[\phi_s(X_{i1})-\bar{\phi}_{\ell s}(z)\big]\bigg\}.
\end{aligned} \tag{A.2}
$$

*Department of Operations Research and Financial Engineering, Princeton University, Princeton, NJ 08544, USA; Email: junweil@princeton.edu †Booth School of Business, The University of Chicago, Chicago, IL 60637, USA; Email: mkolar@chicagobooth.edu ‡Department of Operations Research and Financial Engineering, Princeton University, Princeton, NJ 08544, USA; Email: hanliu@princeton.edu

The computation of $\nabla_j \mathcal{L}_z(\boldsymbol{\beta}_+^{(t)})$ is mostly spent on evaluating the formulation

$$
q(z) = \sum_{i=1}^{n} K_h(X_{i1}-z)u_i \tag{A.3}
$$

for $z \in \{z_1,\ldots,z_M\}$ where $u_1,\ldots,u_n$ are fixed quantities (e.g., $u_i$ could be $Y_i$, $\phi_k(X_{i1})$ or $Y_i\phi_k(X_{i1})$ when evaluating (A.2)) independent of $z$. We introduce a fast method to calculate the general form $q(z)$ and apply it to the computation of (A.2). Without loss of generality, we assume that $z_1 < \ldots < z_M$. The naïve method to evaluate $\{q(z_\ell)\}_{\ell\in[M]}$ separately for different $z$ has the computational complexity $O(nM)$. However, if the kernel function has some special structure, we can reduce the complexity to $O(n+M)$. For example, for the uniform kernel $K(u) = \frac{1}{2}\mathbb{1}\{|u|\le 1\}$, when we vary the value of $z$ from $z_\ell$ to $z_{\ell+1}$, we just need to subtract $u_i$ for those $i \in \{v : X_v \in [z_\ell - h, z_{\ell+1}-h)\}$ and add $u_i$ for those $i \in \{v : X_v \in (z_\ell + h, z_{\ell+1}+h]\}$. For $M \gg h^{-1}$, the cardinality of $\{i : X_{i1} \in (z_\ell - h, z_{\ell+1}-h] \cup (z_\ell + h, z_{\ell+1}+h]\}$ does not increase with $n$ or $d$. Therefore, the complexity to evaluate $\{q(z_\ell)\}_{\ell\in[M]}$ is reduced to $O(n+K)$. For the Epanechnikov kernel $K(u) = (3/4)\cdot(1-u^2)\mathbb{1}\{|u|\le 1\}$, suppose $q(z_\ell)$ is known and define $I_\ell = \{i : X_{i1} \in (z_\ell - h, z_{\ell+1}-h] \cup (z_\ell + h, z_{\ell+1}+h]\}$. We have $q(z_{\ell+1}) = q(z_\ell) + \Delta q(z_\ell)$, where

$$
\Delta q(z_\ell) = q(z_{\ell+1}) - q(z_\ell) = \frac{3}{4}\sum_{i\in I_\ell}\big(1-(X_{i1}/h)^2\big)u_i + \frac{3z}{2h^2}\sum_{i\in I_\ell} X_{i1} + \frac{3z^2}{4}\sum_{i\in I_\ell} u_i.
$$

Similar to the argument for the case of uniform kernel, we also have $|I_\ell| = O(1)$ if $K \gg h^{-1}$. The computational complexity of $\sum_{i\in I_z}(1-X_{i1}^2)u_i$ and the other two summations above for $z = 1,\ldots,z_K$ is $O(n+K)$ and hence the computational complexity of $\{q(z_\ell)\}_{k\in[K]}$ for Epanechnikov kernel is also $O(n+K)$. We can also apply a similar trick to many other kernels. Now we turn back to the calculation of the gradient $\nabla_j \mathcal{L}_z(\boldsymbol{\beta}_+^{(t)})$. Let $\widehat{p}_1(z) = n^{-1}\sum_{i=1}^n K_h(X_{i1}-z)$,

$$Y_k^{(1)}(z) = \frac{1}{n}\sum_{i=1}^n K_h(X_{i1}-z)Y_i, \quad Y_k^{(2)}(z) = \frac{1}{n}\sum_{i=1}^n K_h(X_{i1}-z)\phi_k(X_{ij})Y_i,$$
$$Y_k^{(3)}(z) = \frac{1}{n}\sum_{i=1}^n K_h(X_{i1}-z)\phi_k(X_{ij}) \text{ and } R_{ks}(z) = \frac{1}{n}\sum_{i=1}^n K_h(X_{i1}-z)\phi_k(X_{ij})\phi_s(X_{iu}).$$

For different values of $z$, we denote the components of $\boldsymbol{\beta}_+$ corresponding to the $k$-th B-spline basis for the $j$-th covariate as $\boldsymbol{\beta}_{jk;z}$. According to the expansion in (A.2), we can write the $k$-th coordinate of $\nabla_j \mathcal{L}_z(\boldsymbol{\beta}_+^{(t)})$ as

$$\begin{aligned}\left(\nabla_j \mathcal{L}_z(\boldsymbol{\beta}_+^{(t)})\right)_k &= -Y_k^{(1)}(z) + \bar{\phi}_{jk}(z)Y_k^{(2)}(z) + \frac{1}{n}\sum_{\ell\in[d],s\in[m]} \boldsymbol{\beta}_{\ell s;z}\left(R_{kv}(z) - \bar{\phi}_{jk}(z)Y_v^{(2)}(z)\right)\\ &\quad - \frac{1}{n}\sum_{\ell\in[d],s\in[m]} \boldsymbol{\beta}_{\ell s;z}\left(\bar{\phi}_{\ell s}(z)Y_k^{(3)}(z) - \bar{\phi}_{jk}(z)\bar{\phi}_{\ell s}(z)\widehat{p}_1(z)\right).\end{aligned}$$

Based on the previous discussion on the calculation of $q(z)$ in (A.3), we note that it takes $O(n+M)$ operations to evaluate $\widehat{p}_1(z), Y_k^{(1)}(z), Y_k^{(2)}(z), Y_k^{(3)}(z)$ and $R_{ks}(z)$ for $M$ different values of $z$. Therefore, the computational complexity of each iteration in Algorithm 1 can be reduced from $O(dm^2nM)$ to $O(dm^2(n+M))$. Therefore under the case $M = O(n)$, we can estimate $f_1$ using the introduced procedure with the same computational complexity as (2.10). Since most of existing algorithms for the group Lasso involve evaluating the gradient (Yuan and Lin, 2007; Friedman et al., 2010; Farrell, 2013; Qin et al., 2013), the above argument is applicable to other solvers as well.

# B Covering Properties of the Bootstrap Confidence Bands

In this section, we prove the theorems on the coverage probabilities for the Gaussian multiplier bootstrap confidence bands $\mathcal{C}_{n,\alpha}^b$ in (2.18).

## B.1 Proof of Theorem 3.4

We first prove that $\mathcal{C}^b_{n,\alpha}$ in (2.18) is honest. The strategy to prove the result is to establish a sequence of processes from $\widehat{\mathbb{H}}_n(z)$ that approximate $\widetilde{Z}_n(z)$. We consider the following four Gaussian processes

$$\widehat{\mathbb{H}}_n(z) = \frac{1}{\sqrt{nh^{-1}}}\sum_{i=1}^n \xi_i \cdot \frac{\widehat{\sigma} K_h(X_{i1}-z)\mathbf{\Psi}_{i\bullet}^T\widehat{\boldsymbol{\theta}}_z}{\widehat{\sigma}_n(z)}; \tag{B.1}$$

$$\widehat{\mathbb{H}}_n^{(1)}(z) = \frac{1}{\sqrt{nh^{-1}}}\sum_{i=1}^n \xi_i \cdot \frac{\sigma K_h(X_{i1}-z)\mathbf{\Psi}_{i\bullet}^T\widehat{\boldsymbol{\theta}}_z}{\widehat{\sigma}_n(z)}, \tag{B.2}$$

$$\widetilde{\mathbb{H}}_n(z) = \frac{1}{\sqrt{nh^{-1}}}\sum_{i=1}^n \varepsilon_i \frac{K_h(X_{i1}-z)\mathbf{\Psi}_{i\bullet}^T\widehat{\boldsymbol{\theta}}_z}{\widehat{\sigma}_n(z)}, \tag{B.3}$$

$$\widetilde{Z}_n(z) = \sqrt{nh}\cdot\widehat{\sigma}_n^{-1}(z)\left(\widetilde{f}_1^u(z) - f_1(z)\right). \tag{B.4}$$

Corollary 3.1 of Chernozhukov et al. (2014a) provides sufficient conditions for the confidence band to be asymptotically honest. Specifically, we need to verify the following high-level conditions:

**H1** There exists a Gaussian process $\mathbb{G}_n(z)$ and a sequence of random variables $W_n^0$ such that $W_n^0 \overset{d}{=} \sup_{z\in\mathcal{X}}\mathbb{G}_n(z)$. Furthermore, $\mathbb{E}[\sup_{z\in\mathcal{X}}\mathbb{G}_n] \le C\sqrt{\log n}$ and

$$\mathbb{P}(|W_n^Z - W_n^0| > \varepsilon_{1n}) < \delta_{1n}$$

for some $\varepsilon_{1n}$ and $\delta_{1n}$.

**H2** For any $\epsilon > 0$, the anti-concentration inequality

$$\sup_{x\in\mathbb{R}}\mathbb{P}\left(\left|\sup_{z\in\mathcal{X}}|\mathbb{G}_n(z)| - x\right| \le \epsilon\right) \le C\epsilon\sqrt{\log n}.$$

holds.

**H3** Let $c_n(\alpha)$ be the $(1-\alpha)$-quantile of $W_n^Z$ and $\widehat{c}_n(\alpha)$ be the $1-\alpha$ quantile of $\widehat{W}_n$. There exists $\tau_n$, $\varepsilon_{2n}$ and $\delta_{2n}$ such that

$$\mathbb{P}\left(\widehat{c}_n(\alpha) < c_n(\alpha+\tau_n) - \varepsilon_{2n}\right) \le \delta_{2n} \quad \text{and} \quad \mathbb{P}\left(\widehat{c}_n(\alpha) > c_n(\alpha-\tau_n) + \varepsilon_{2n}\right) \le \delta_{2n}.$$

**H4** There exists $\varepsilon_{3n}$ and $\delta_{3n}$ such that

$$\mathbb{P}\left(\sup_{z\in\mathcal{X}}\left|\frac{\widehat{\sigma}\sqrt{\widehat{p}_1(z)}}{\sigma\sqrt{p_1(z)}}-1\right|>\varepsilon_{3n}\right)\leq\delta_{3n}.$$

If the high-level conditions **H1** - **H4** are verified, Corollary 3.1 in Chernozhukov et al. (2014a) implies that

$$\mathbb{P}(f_1\in\mathcal{C}^b_{n,\alpha})\geq 1-\alpha-(\varepsilon_{1n}+\varepsilon_{2n}+\varepsilon_{3n}+\delta_{1n}+\delta_{2n}+\delta_{3n}).$$

In the remaining part of the proof, we show that the conditions are satisfied.

The roadmap is to establish that the process in (B.4) is close to the process in (B.1) following the chain $\widetilde{Z}_n\to\widetilde{\mathbb{H}}_n\to\widehat{\mathbb{H}}_n^{(1)}\to\widehat{\mathbb{H}}_n$. After that, we can check conditions **H1** – **H3**. Since we do not use the population $\sigma_n(z)=\mathbb{E}[\widehat{\sigma}_n(z)]$ in the intermediate processes, we do not need to check the condition **H4**.

In order to verify the condition **H1**, we first bound the difference between $\sup_{z\in\mathcal{X}}\widetilde{\mathbb{H}}_n(z)$ and $\sup_{z\in\mathcal{X}}Z_n(z)$. We begin by considering two auxiliary processes

$$\widetilde{\mathbb{H}}_n'(z)=\frac{1}{\sqrt{nh^{-1}}}\sum_{i=1}^{n}\varepsilon_iK_h(X_{i1}-z)\mathbf{\Psi}_{i\bullet}^T\widehat{\boldsymbol{\theta}}_z\quad\text{and}\quad\widetilde{Z}_n'(z)=\sqrt{nh}\left(\widetilde{f}_1^u(z)-f_1(z)\right).$$

Notice that the above processes are un-normalized version of (B.2) and (B.4), that is, $\widetilde{\mathbb{H}}_n'(z)=\widehat{\sigma}_n(z)\widetilde{\mathbb{H}}_n(z)$ and $\widetilde{Z}_n'(z)=\widehat{\sigma}_n(z)\widetilde{Z}_n(z)$. The following lemma provides a direct bound for the difference between $\widetilde{\mathbb{H}}_n'(z)$ and $\widetilde{Z}_n'(z)$.

**Lemma B.1.** Suppose that Assumptions (**A1**) - (**A6**) hold. If $h\asymp n^{-\delta}$ for $\delta>1/5$ and $m\asymp n^p$ for $p\in(0,(10\delta-2)/3)$, there exists a constant $c_0>\delta/2$ such that with probability $1-c/n$,

$$\sup_{z\in\mathcal{X}}\left|\widetilde{\mathbb{H}}_n'(z)-\widetilde{Z}_n'(z)\right|\leq Cn^{-c_0}.$$

We defer the proof of the lemma to Section E.5 and proceed to prove Theorem 3.4. We also need to study $\widehat{\sigma}$ and $\widehat{\sigma}_n(z)$ in the following lemmas.

**Lemma B.2.** Let the estimator for $\mathrm{Var}(\varepsilon) = \sigma^2$ be $\widehat{\sigma}^2 = \frac{1}{n}\sum_{i=1}^n \widehat{\varepsilon}_i^2$. Let

$$r_n := \sqrt{\frac{s^2 \log(dmh^{-1})}{nm^{-2}h}} + \frac{s^{3/2}}{m^{3/2}} + \frac{s\log(dh^{-1})}{nm^{-5/2}} + s\sqrt{m}h^2. \tag{B.5}$$

Under Assumption (**A1**) and (**A4**), there exists constants $C$ such that $\mathbb{P}\big(|\widehat{\sigma}^2 - \sigma^2| \geq Cr_n\sqrt{m}\big) \leq 6/n$.

**Lemma B.3.** Let $\boldsymbol{\Sigma}_z' = n^{-1}\boldsymbol{\Psi}\mathbf{W}_z^2\boldsymbol{\Psi}^T$. If $mh = o(1), h = n^{-\delta}$ for $\delta > 1/5$, there exist constants $c, C$ such that for sufficiently large $n$, with probability $1 - c/n$, for any $z \in \mathcal{X}$,

$$ch^{-1}\mathbf{e}_1^T\boldsymbol{\theta}_z \leq \widehat{\boldsymbol{\theta}}_z^T\boldsymbol{\Sigma}_z'\widehat{\boldsymbol{\theta}}_z \leq Ch^{-1}\mathbf{e}_1^T\boldsymbol{\theta}_z.$$

We defer the proof of this lemma to Section E.6. From Lemmas B.3 and E.3, we have an upper bound of the inverse of $\widehat{\sigma}_n^2(z) = \widehat{\boldsymbol{\theta}}_z^T\boldsymbol{\Sigma}_z'\widehat{\boldsymbol{\theta}}_z$ as

$$\sup_{z\in\mathcal{X}} \sqrt{h}\cdot\widehat{\sigma}_n^{-1}(z) \leq C. \tag{B.6}$$

With Lemma B.1 and Lemma B.3, we are ready to bound the difference between $\sup_{z\in\mathcal{X}} \widetilde{\mathbb{H}}_n(z)$ and $\sup_{z\in\mathcal{X}} Z_n(z)$. Let $c_0$ be the constant in Lemma B.1. Let $h \asymp n^{-\delta}$ for $\delta > 1/5$ and $m \asymp n^p$ for $p \in (0, (10\delta - 2)/3)$. We denote $c = c_0 - \delta/2$ and observe that $c > 0$ by Lemma B.1. From Lemma B.1 and (B.6), we have

$$\begin{aligned}\mathbb{P}\left(\sup_{z\in\mathcal{X}}\left|\widetilde{\mathbb{H}}_n(z) - \widetilde{Z}_n(z)\right| \geq Cn^{-c}\right) &\leq \mathbb{P}\left(\sup_{z\in\mathcal{X}}\left|\widetilde{\mathbb{H}}_n'(z) - \widetilde{Z}_n'(z)\right| \geq C\widehat{\sigma}_n(z)n^{-c_0}/\sqrt{h}\right)\\ &\leq \mathbb{P}\left(\sup_{z\in\mathcal{X}}\left|\widetilde{\mathbb{H}}_n'(z) - \widetilde{Z}_n'(z)\right| \geq C^2n^{-c_0}\right) \leq 1/n.\end{aligned}$$

Define $V_n^0 = \sup_{z\in\mathcal{X}} \widetilde{\mathbb{H}}_n(z)$ and $\widetilde{V}^Z = \sup_{z\in\mathcal{X}} \widetilde{\mathbb{H}}_n(z)$. Since $\sup_{z\in\mathcal{X}} \widetilde{\mathbb{H}}_n(z)$ is a Gaussian process conditional on $\{\boldsymbol{X}_{i1}\}_{i\in[n]}$, we verify **H1** by

$$\mathbb{P}\left(|V_n^0 - \widetilde{V}^Z| \geq Cn^{-c}\right) \leq \mathbb{P}\left(\sup_{z\in\mathcal{X}}|\widetilde{\mathbb{H}}_n(z) - \widetilde{Z}_n(z)| \geq Cn^{-c}\right) \leq \frac{1}{n}. \tag{B.7}$$

The condition **H2** follows from **H1** and the anti-concentration inequality in Corollary 2.1 of Chernozhukov et al. (2014a).

Next, we check **H3** by bounding the difference between (B.1) and (B.2). We first approximate $\widehat{\mathbb{H}}_n(z)$ by $\widehat{\mathbb{H}}_n^{(1)}(z)$. By Lemma B.2, if $mh = o(1)$ and $h \asymp n^{-\delta}$ for $\delta > 1/5$, with probability $1 - 6/n$,

$$|\widehat{\sigma} - \sigma| < C\sqrt{r_n} m^{1/4} = o\Big(n^{-1/10}\Big),$$

where $r_n$ is defined in (B.5). We denote $\widehat{V}_n = \sup_{z\in\mathcal{X}} \widehat{\mathbb{H}}_n(z)$, $\widehat{V}_n^{(1)} = \sup_{z\in\mathcal{X}} \widehat{\mathbb{H}}_n^{(1)}(z)$ and the difference between $\widehat{V}_n - \widehat{V}_n^{(1)}$. Let $\Delta\mathbb{H}^{(1)}(z) = \widehat{\mathbb{H}}_n^{(1)}(z) - \widehat{\mathbb{H}}_n(z)$. We have

$$\sup_{z\in\mathcal{X}} \big|\Delta\mathbb{H}^{(1)}(z)\big| \leq |\widehat{\sigma} - \sigma| \sup_{z\in\mathcal{X}} \sqrt{h} \cdot \widehat{\sigma}_n^{-1}(z) \left(\sup_{z\in\mathcal{X}} I_1(z) + \sup_{z\in\mathcal{X}} I_2(z)\right),$$

where $I_1(z) = n^{-1}\sum_{i=1}^n K_h(X_{i1} - z)\big|\boldsymbol{\Psi}_{i\bullet}^T(\widehat{\boldsymbol{\theta}}_z - \boldsymbol{\theta}_z)\big|$ and $I_2(z) = n^{-1}\sum_{i=1}^n K_h(X_{i1} - z)\,\big|\boldsymbol{\Psi}_{i\bullet}^T\boldsymbol{\theta}_z\big|$.

In order to bound $I_1(z)$, we first state a technical lemma that characterizes the estimation error between $\widehat{\boldsymbol{\theta}}_z$ and $\boldsymbol{\theta}_z$.

**Lemma B.4.** Let $\widehat{\boldsymbol{\theta}}_z$ be a minimizer of (2.15). Suppose that Assumptions (**A1**) - (**A6**) hold. If the parameter $\gamma$ in the optimization program (2.15) is chosen as in (3.9), then with probability $1 - c/d$,

$$\sup_{z\in\mathcal{X}} (\widehat{\boldsymbol{\theta}}_z - \boldsymbol{\theta}_z)^T \widehat{\boldsymbol{\Sigma}}_z (\widehat{\boldsymbol{\theta}}_z - \boldsymbol{\theta}_z) \leq Cm\Bigg(\sqrt{\frac{m\log(dm)}{nh}} + \frac{m}{nh} + \sqrt{\frac{\log(1/h)}{nh}}\Bigg). \tag{B.8}$$

We defer the proof of this lemma to Section E.4. Using Lemma B.4 we bound $I_1(z)$. Applying Cauchy-Schwarz inequality, we have

$$\begin{aligned}\sup_{z\in\mathcal{X}} |I_1(z)| &\leq \sup_{z\in\mathcal{X}} \left(\frac{1}{n}\sum_{i=1}^n K_h(X_{i1} - z)\left(\boldsymbol{\Psi}_{i\bullet}^T(\widehat{\boldsymbol{\theta}}_z - \boldsymbol{\theta}_z)\right)^2\right)^{1/2} \left(\frac{1}{n}\sum_{i=1}^n K_h(X_{i1} - z)\right)^{1/2} \\ &\leq C\sqrt{m}\Bigg(\sqrt{\frac{m\log(dm)}{nh}} + \frac{m}{nh} + \sqrt{\frac{\log(1/h)}{nh}}\Bigg),\end{aligned} \tag{B.9}$$

where the last inequality is due to Lemma B.4 and

$$\sup_{z\in\mathcal{X}} n^{-1}\sum_{i=1}^n K_h(X_{i1} - z) = \sup_{z\in\mathcal{X}} p_1(z) + o(1).$$

For $I_2(z)$, we have the following inequality

$$\begin{aligned}
\sup_{z\in\mathcal{X}} |I_2(z)| &\le \sup_{z\in\mathcal{X}} \frac{1}{n}\|\mathbf{\Psi}^T\mathbf{W}_z\mathbf{1}\|_{2,\infty}\|\boldsymbol{\theta}_z\|_1 \\
&\le \sup_{z\in\mathcal{X}} \frac{1}{n}\|\mathbf{W}_z^{1/2}\mathbf{\Psi}_{\bullet j}\mathbf{\Psi}_{\bullet j}^T\mathbf{W}_z^{1/2}\|_2\|\mathbf{W}_z^{1/2}\mathbf{1}\|_2\|\boldsymbol{\theta}_z\|_1 \qquad \text{(B.10)} \\
&\le \frac{C}{\sqrt{m}}\cdot\sup_{z\in\mathcal{X}}\sqrt{p(z)}\cdot\sqrt{m} = O(1).
\end{aligned}$$

Therefore, combining (B.9) and (B.10), we have

$$\mathbb{P}\left(\left|\widehat{V}_n - \widehat{V}_n^{(1)}\right| > C\sqrt{r_n}m^{1/4}\right) \le n^{-1}.$$

When $h \asymp n^{-\delta}$ for $\delta > 1/5$ and $m \asymp n^p$ for $p \in (0, (10\delta-2)/3)$, there exists a constant $c$ such that $\sqrt{r_n}m^{1/4} = O(n^{-c})$. Since $\sigma\xi_i \stackrel{d}{=} \varepsilon_i$, we also have $\sup_{z\in\mathcal{X}}\widehat{\mathbb{H}}_n^{(1)}(z) \stackrel{d}{=} \sup_{z\in\mathcal{X}}\widetilde{\mathbb{H}}_n(z)$. Combining with (B.7), we have

$$\mathbb{P}\left(\left|\widehat{V}_n - \widetilde{V}_n^Z\right| > 2Cn^{-c}\right) \le 2n^{-1}.$$

Therefore, we can bound the probability

$$\begin{aligned}
\mathbb{P}(\widetilde{V}_n^Z \le \widehat{c}_n(\alpha) + 2Cn^{-c}) &\ge \mathbb{P}(\widetilde{V}_n^Z \le \widehat{c}_n(\alpha)) - \mathbb{P}(|\widehat{V}_n - \widetilde{V}_n^Z| > 2Cn^{-c}) \\
&\ge 1 - \alpha - 2c/n^c, \qquad \text{(B.11)}
\end{aligned}$$

which implies that the estimated quantile has the following lower bound

$$\widehat{c}_n(\alpha) \ge c_n(\alpha + 2Cn^{-c}) - 2cn^{-c}. \qquad \text{(B.12)}$$

Similarly, we also have $\widehat{c}_n(\alpha) \le c_n(\alpha - 2Cn^{-c}) + 2cn^{-c}$. By setting $\tau_n = 2Cn^{-c}$, $\varepsilon_{2n} = 2cn^{-c}$ and $\delta_{2n} = 2cn^{-c}$, we have

$$\mathbb{P}\left(\widehat{c}_n(\alpha) \ge c_n(\alpha + 2Cn^{-c}) - 2cn^{-c} \text{ and } \widehat{c}_n(\alpha) \le c_n(\alpha - 2Cn^{-c}) + 2cn^{-c}\right) \le 2c/n^c,$$

which verifies the condition **H3**.

Now, since we have checked the high-level conditions **H1** – **H3**, since the high-level conditions **H1** – **H4** are verified, the result follows from Corollary 3.1 in Chernozhukov et al. (2014a) such that

$$\mathbb{P}\big(f_1 \in \mathcal{C}^b_{n,\alpha}\big) \geq 1 - \alpha - Cn^{-c},$$

which completes the proof of the theorem.

# C Properties of Model Assumptions

In this section, we give proof to the proposition justifying certain assumptions.

## C.1 Proof of Propositions 3.1

Let $J$ be arbitrary subset of $[d]$ and for any $(\alpha, \boldsymbol{\beta}) = (\alpha, \boldsymbol{\beta}_2^T, \ldots, \boldsymbol{\beta}_d^T)^T \in \mathbb{C}_\beta^{(\kappa)}(J)$, where $\mathbb{C}_\beta^{(\kappa)}(J)$ is defined (3.1), we consider the functions $h_1(x_1) \equiv \alpha$ and $h_j(x_j) = \sum_{k=1}^m \boldsymbol{\beta}_{jk}\psi_{jk}(x_j)$, for $j = 2, \ldots, d$. From the cone restriction in (3.1), we have

$$\sum_{j\notin J, j\neq 1} \|\boldsymbol{\beta}_j\|_2 \leq \kappa \sum_{j\in J, j\neq 1} \|\boldsymbol{\beta}_j\|_2 + \kappa\sqrt{m}|\alpha|$$

and the B-spline property $\|h_j\|_2^2 \asymp m^{-1}\|\boldsymbol{\beta}\|_2^2$ in (7.10), there exists a constant $c_1 > 0$ such that

$$\begin{aligned}\sqrt{\frac{m}{c_1}}\sum_{j\notin J}\|h_j\|_2 &\leq \sum_{j\notin J\setminus\{1\}} \|\boldsymbol{\beta}_j\|_2 + \sqrt{\frac{m}{c_1}}|\alpha| \\ &\leq \kappa \sum_{j\in J\setminus\{1\}} \|\boldsymbol{\beta}_j\|_2 + \left(\kappa + c_1^{-1/2}\right)\sqrt{m}|\alpha| \\ &\leq \left((\sqrt{c_1}+1)\kappa + c_1^{-1/2}\right)\sqrt{m}\cdot\kappa\sum_{j\in J}\|h_j\|_2.\end{aligned}$$

Let $c$ be the smallest constant satisfying $c\kappa \geq \left(\sqrt{c_1}+1\right)\kappa + c_1^{-1/2}$. Therefore, we have $(h_1, \ldots, h_d) \in \mathbb{C}_h^{(c\kappa)}(J)$. Consider the $L^2$ norm $\|\cdot\|_{L^2(\mu_z)}$ induced by the measure $\mu_z(f) = \mathbb{E}[K_h(X_1 - z)g(\cdot)]$ for any bounded measurable $g$. For any $j \geq 2$, let $p_{1,j}(x_1, x_j)$ be the joint density of $(X_1, X_j)$. Under

Assumption (**A1**), for any measurable $g$ and any $z \in \mathcal{X}$, we have

$$\begin{aligned}
\|g\|_{L^2(\mu_z)}^2 &= \mathbb{E}[K_h(X_1 - z)g^2(X_j)] \\
&= \int_{\mathcal{X}^2} K_h(x-z)g^2(y)p_{1,j}(x,y)dxdy \\
&\geq \frac{b}{B^2}\int_{\mathcal{X}^2} K_h(x-z)g^2(y)p_1(x)p_j(y)dxdy = \frac{b}{B^2}\mathbb{E}[g^2(X_j)] = \frac{b}{B^2}\|g\|_2^2.
\end{aligned}$$

Therefore, for any $z \in \mathcal{X}$,

$$\boldsymbol{\beta}_+^T \boldsymbol{\Sigma}_z \boldsymbol{\beta}_+ \geq \frac{b}{B^2}\left\|\sum_{j=1}^d h_j\right\|_2^2 \geq \frac{b\beta_{2,c\kappa}^{-2}(J)}{B^2}\sum_{j\in J}\|h_j\|_2^2 \geq \frac{b\bar{\beta}_{2,c\kappa}^{-2}}{|J|B^2}\left(\sum_{j\in J}\|h_j\|_2\right)^2. \tag{C.1}$$

Applying the cone restriction on $(h_1, \ldots, h_d)$, we further have

$$\left(\sum_{j\in J}\|h_j\|_2\right)^2 \geq \frac{1}{(c\kappa+1)^2}\left(\sum_{j=1}^d \|h_j\|_2\right)^2 \geq \frac{1}{(c\kappa+1)^2}\sum_{j=1}^d \|h_j\|_2^2 \geq \frac{Cm^{-1}}{(c\kappa+1)^2}\left(\sum_{j=2}^d \|\boldsymbol{\beta}_j\|_2^2 + m\alpha^2\right).$$

Combining the above inequality with (C.1), we obtain that for any $z \in \mathcal{X}$,

$$\boldsymbol{\beta}_+^T \boldsymbol{\Sigma}_z \boldsymbol{\beta}_+ \geq \frac{Cbm^{-1}\bar{\beta}_{2,c\kappa}^{-2}}{sB^2(c\kappa+1)^2}\left(\|\boldsymbol{\beta}\|_2^2 + m\alpha^2\right), \qquad \text{for any } \boldsymbol{\beta}_+ \in \mathbb{C}_\beta^{(\kappa)}(J).$$

This completes the proof.

# D Auxiliary Lemmas for Estimation Rate

In this section, we give detailed proofs of technical lemmas stated in Section 7. The principal technique used in the proofs of this section is the control of the suprema of empirical processes. In the entire section, we will abuse the notation $\sigma_P^2$ as the variance for certain empirical process.

## D.1 Restricted eigenvalue condition

We provide a proof of Lemma 7.1 in this section. Before stating the main part of the proof, we begin with a technical lemma.

**Lemma D.1** (Resctricted eigenvalue condition)**.** With probability larger than $1 - c/(dm)$, for any $\boldsymbol{\theta} = (\alpha, \boldsymbol{\beta}^T)^T$, with $\alpha \in \mathbb{R}$ and $\boldsymbol{\beta} \in \mathbb{R}^{(d-1)m}$, and any $z \in \mathcal{X}$ it holds that

$$\boldsymbol{\theta}^T \widehat{\boldsymbol{\Sigma}}_z \boldsymbol{\theta} \geq \boldsymbol{\theta}^T \boldsymbol{\Sigma}_z \boldsymbol{\theta} - C\left(\sqrt{\frac{m\log(dm)}{nh}} + \frac{m}{nh} + \frac{1}{\sqrt{nh}} + h^2\right)\|\boldsymbol{\theta}\|_{1,2}^2,$$

where $\|\boldsymbol{\theta}\|_{1,2} = |\alpha| + \|\boldsymbol{\beta}\|_{1,2}$. Moreover, for any $j \in [d]$ there exists a constant $C$ such that

$$\sup_{x\in\mathcal{X}} \frac{1}{n}\big\|\boldsymbol{\Psi}_{\bullet j}\mathbf{W}_z\boldsymbol{\Psi}_{\bullet j}^T\big\|_2^2 \leq Cm^{-1}.$$

*Proof.* The proof strategy is to study suprema of the entries of $\widehat{\boldsymbol{\Sigma}}_z - \boldsymbol{\Sigma}_z$. We denote the $(u, v)$ entry of $\boldsymbol{\Sigma}_z$ as $\boldsymbol{\Sigma}_z(u, v)$ and similarly for $\widehat{\boldsymbol{\Sigma}}_z$. Let $\mathbb{E}_n$ denote the empirical expectation. We first study the random variable

$$Z_{kk'jj'} = \sup_{z\in\mathcal{X}}(\mathbb{E}_n - \mathbb{E})[K_h(X_{i1} - z)\psi_{jk}(X_{ij})\psi_{j'k'}(X_{ij'})].$$

Notice that $Z_{kk'jj'} = \sup_{z\in\mathcal{X}}[\widehat{\boldsymbol{\Sigma}}_z - \boldsymbol{\Sigma}_z](1 + (j-2)m + k, 1 + (j'-2)m + k')$ for $j, j' \geq 2$ and $k \in [m]$. In order to bound $Z_{kk'jj'}$, we turn to study the covering number of the space

$$\mathcal{G}_h = \left\{g_z(x_1, x_2, x_3) = h^{-1}K(h^{-1}(x_1 - z))\psi_{jk}(x_2)\psi_{j'k'}(x_3) \,\middle|\, z \in \mathcal{X}, x_1, x_2, x_3 \in \mathcal{X}\right\}.$$

Let $\mathcal{F}_h = \left\{h^{-1}K(h^{-1}(\cdot - z)) \,|\, z \in \mathcal{X}\right\}$ and let $\|K\|_{\mathrm{TV}}$ be the total variation of $K(\cdot)$. From Lemma F.3, we have

$$\sup_Q N\left(\mathcal{F}_h, L^2(Q), \epsilon\right) \leq \left(\frac{2||K||_{\mathrm{TV}}A}{h\epsilon}\right)^4, \quad 0 < \epsilon < 1,$$

where the supremum is taken over all probability measures $Q$ on $\mathbb{R}$. Let $\widetilde{\mathcal{F}}_h$ be an $\epsilon/L$-cover of $\mathcal{F}_h$ with respect to $Q$, where $L \geq \|\psi_{jk}\|_\infty$ for any $k$. We construct an $\epsilon$-cover for $\mathcal{G}_h$ with respect to $\mathbb{P}_n = n^{-1}\sum_{i=1}^n \delta_{X_{i1},\ldots,X_{id}}$ as

$$\widetilde{\mathcal{G}}_h = \left\{f_1(x_1)\psi_{jk}(x_2)\psi_{j'k'}(x_3) \,|\, f_1 \in \widetilde{\mathcal{F}}_h\right\}.$$

For a function $g_z = h^{-1}K(h^{-1}(x_1 - z))\psi_{jk}(x_2)\psi_{j'k'}(x_3) \in \mathcal{G}_h$, let

$$\widetilde{g}_z = h^{-1}K(h^{-1}(x_1 - \widetilde{z}))\psi_{jk}(x_2)\psi_{j'k'}(x_3) \in \widetilde{\mathcal{G}}_h$$

be the corresponding element in the cover. Here $h^{-1}K(h^{-1}(x_1 - \widetilde{z})) \in \widetilde{\mathcal{F}}_h$ is the corresponding element in the cover for $h^{-1}K(h^{-1}(x_1 - z)) \in \mathcal{F}_h$. Then

$$\begin{aligned}\|g_z - \widetilde{g}_z\|^2_{L^2(Q)} &= \mathbb{E}_Q\left[\left((K_h(X_1 - z) - K_h(X_1 - \widetilde{z})\right)\psi_{jk}(X_{ji})\psi_{j'k}(X_{j'i})\right]^2 \\ &\leq \mathbb{E}_Q\left[((K_h(X_1 - z) - K_h(X_1 - \widetilde{z}))\right]^2 \leq \epsilon^2\end{aligned}$$

and the covering number can be bounded as

$$N\left(\mathcal{G}_h, L^2(\mathbb{P}_n), \epsilon\right) \leq \left(\frac{2||K||_{\mathrm{TV}}AL}{h\epsilon}\right)^4. \tag{D.1}$$

Observe that all functions in $\mathcal{G}_h$ are bounded by $U = 4h^{-1}\|K\|_\infty$ and

$$\begin{aligned}\sigma_P^2 &:= \mathbb{E}\left[\left(K_h\left(X_1 - z\right)\left(\psi_{jk}(X_j)\psi_{j'k'}(X_{j'})\right)^2\right]\right. \\ &= h^{-2}\mathbb{E}\left[K^2\left(h^{-1}(X_1 - z)\right)\mathbb{E}\left[(\psi^2_{jk}(X_j)\psi^2_{j'k'}(X_{j'})\,|\,X_1\right]\right] \\ &\leq L^2m^{-2}h^{-2}\mathbb{E}\left[K^2\left(h^{-1}(X_1 - z)\right)\right] \\ &= L^2m^{-2}h^{-1}\int K^2\left(u\right)p_{X_1}(z + uh)du \leq bL^2m^{-2}h^{-1},\end{aligned}$$

where the first and last inequalities are due to Assumption (**A1**). The bound above does not depend on the particular choice of $z$. If $m(nh)^{-1} = o(1)$, we have $n\sigma_P^2 \geq CU^2\log\left(U\sigma^{-1}\right)$, and from Lemma F.2, we have

$$\mathbb{E}[Z_{kk'jj'}] \leq C_1\sqrt{\frac{\log(C_2m)}{nm^2h}}, \tag{D.2}$$

where the constants $C_1, C_2$ are independent of $k, k', j, j'$. As $|Z_{kk'jj'}| \leq 4h^{-1}$ and $\sigma_P^2 \leq Cm^{-2}h^{-2}$, we can apply Lemma F.4 to obtain

$$\mathbb{P}\Big(Z_{kk'jj'} \geq \mathbb{E}[Z_{kk'jj'}] + t\sqrt{Cm^{-2}h^{-1} + 4h^{-1}\mathbb{E}[Z_{kk'jj'}]} + 4t^2h^{-1}/3\Big) \leq \exp(-nt^2). \tag{D.3}$$

For $t = \log d/\sqrt{n}$, there exists a constant $C$ such that

$$Z_{kk'jj'} \le C \log dm/\sqrt{nm^2h} + C/(nh)$$

with probability $1 - 1/d$. Combining (D.2) with (D.3), there exists a constant $C$ such that

$$\begin{aligned}
&\mathbb{P}\left(\max_{j,j'\ge 2,k,k'\in[m]} |Z_{kk'jj'}| > 2\mathbb{E}[Z_{kk'jj'}] + t\sqrt{Cm^{-2}h^{-1} + 4h^{-1}\mathbb{E}[Z_{kk'jj'}]} + 4t^2h^{-1}/3\right)\\
&\le \mathbb{P}\Big(\max_{k,k'\in[m],j,j'\ge 2} |Z_{kk'jj'} - \mathbb{E}[Z_{kk'jj'}]| > t\sqrt{Cm^{-2}h^{-1} + 4h^{-1}\mathbb{E}[Z_{kk'jj'}]} + 4t^2h^{-1}/3 + \mathbb{E}[Z_{kk'jj'}]\Big)\\
&\le (dm)^2 \exp\left(-nt^2\right).
\end{aligned}$$

Let $t = 3\sqrt{\log(dm)/n}$, $n_{jk} = 1 + (j-2)m + k$ and $n_{j'k'} = 1 + (j'-2)m + k'$) and we obtain that

$$\sup_{z\in\mathcal{X}} \max_{j,j'\ge 2,k,k'\in[m]} \left|\left[\widehat{\mathbf{\Sigma}}_z - \mathbf{\Sigma}_z\right](n_{jk}, n_{j'k'})\right| = O_P\left(\frac{1}{nh} + \sqrt{\frac{\log(dm)}{nm^2h}}\right). \tag{D.4}$$

Similarly, we define $\bar{Z}_{kj} = \sup_{z\in\mathcal{X}}(\mathbb{E}_n - \mathbb{E})[K_h(X_{i1} - z)\psi_{jk}(X_{ij})]$. Following the similar procedure as above, we apply Lemma F.2 to obtain that for some constant $C$,

$$\sigma_P^2 := \mathbb{E}\left[\left(K_h\left(X_1 - z\right)\psi_{jk}(X_j)\right)^2\right] \le Cm^{-1}h^{-1},$$

and $U \le h^{-1}$, which implies the following inequality

$$\mathbb{E}[\bar{Z}_{kj}] \le C_1\sqrt{\frac{\log(C_2 m)}{nmh}}. \tag{D.5}$$

We now turn to study the remaining entries of $\widehat{\mathbf{\Sigma}}_z - \mathbf{\Sigma}_z$. Using the same arguments as in (D.4) and (D.5), we can derive an upper bound on the difference

$$\sup_{z\in\mathcal{X}} \max_{j\ge 2} \left|\widehat{\mathbf{\Sigma}}_z(1 + (j-2)m + k, 1) - \mathbf{\Sigma}_z(1 + (j-2)m + k, 1)\right| = O_P\left(\frac{1}{nh} + \sqrt{\frac{\log(dm)}{nmh}}\right). \tag{D.6}$$

From Assumption **(A1)**, the density function of $X_1$, $p_1(x)$, is smooth. Recall that $\widehat{p}_1(z) = n^{-1}\sum_{i=1}^n K_h(X_{i1} - z)$. Applying the supreme norm rate for a kernel density estimator established in Theorem 2.3 of Giné and Guillou (2002), we have $\|\widehat{p}_1 - \mathbb{E}[\widehat{p}_1]\|_\infty = O_P\big(\sqrt{\log(1/h)/(nh)}\big)$ and therefore we can get the rate

$$\sup_{z\in\mathcal{X}}\left|\widehat{\boldsymbol{\Sigma}}_z(1,1) - \boldsymbol{\Sigma}_z(1,1)\right| = \sup_{z\in\mathcal{X}}\left|\widehat{p}_1(z) - \mathbb{E}[\widehat{p}_1(z)]\right| = O_P\left(\sqrt{\frac{\log(1/h)}{nh}}\right). \tag{D.7}$$

Combining (D.4), (D.6) and (D.7), according to Hölder inequality, we have for any $z \in \mathcal{X}$

$$\begin{aligned}\left|\boldsymbol{\theta}^T(\widehat{\boldsymbol{\Sigma}}_z - \boldsymbol{\Sigma}_z)\boldsymbol{\theta}\right| &\le \|\boldsymbol{\theta}\|_1^2\|\widehat{\boldsymbol{\Sigma}}_z - \boldsymbol{\Sigma}_z\|_{\max} \\ &\le \|\boldsymbol{\theta}\|_{1,2}^2\left\{\sup_{z\in\mathcal{X}}\max_{t,t'\neq 1} m\left|\widehat{\boldsymbol{\Sigma}}_z(t,t') - \boldsymbol{\Sigma}_z(t,t')\right| + \sup_{z\in\mathcal{X}}\left|\widehat{\boldsymbol{\Sigma}}_z(1,1) - \boldsymbol{\Sigma}_z(1,1)\right|\right\} \\ &\le C\left(\sqrt{\frac{m\log(dm)}{nh}} + \frac{m}{nh} + \sqrt{\frac{\log(1/h)}{nh}}\right)\|\boldsymbol{\theta}\|_{1,2}^2,\end{aligned} \tag{D.8}$$

which completes the proof of the first part of the Lemma.

An upper bound on $\sup_{x\in\mathcal{X}} n^{-1}\|\boldsymbol{\Psi}_{\bullet j}\mathbf{W}_z\boldsymbol{\Psi}_{\bullet j}^T\|_2^2$ can be obtain in a way similar to the proof of Lemma 6.2 in Zhou et al. (1998). For any $\boldsymbol{\beta}_j = (\beta_1, \ldots, \beta_m)^T$, let $u(x_j) = \sum_{k=1}^m \beta_k\psi_{jk}(x_j)$. Let the joint density function between $X_1$ and $X_j$ is $p_{1,j}(x_1, x_j)$. From Assumption (**A1**), we have for any $z \in \mathcal{X}$,

$$\begin{aligned}\frac{1}{n}\boldsymbol{\beta}_j^T\mathbb{E}[\boldsymbol{\Psi}_{\bullet j}\mathbf{W}_z\boldsymbol{\Psi}_{\bullet j}^T]\boldsymbol{\beta}_j &= \int\frac{1}{h}K\left(\frac{x_1 - z}{h}\right)u^2(x_j)p_{1,j}(x_1,x_j)dx_1dx_j \\ &\le C\int K(u)du\int u^2(x_j)dx_j \le Cm^{-1}\sum_{k=1}^m\beta_k^2.\end{aligned} \tag{D.9}$$

Furthermore, we also have

$$\begin{aligned}\sup_{x\in\mathcal{X}}\frac{1}{n}\boldsymbol{\beta}_j^T\mathbb{E}[\boldsymbol{\Psi}_{\bullet j}\boldsymbol{\Psi}_{\bullet j}^T \mid X_1 = x]\boldsymbol{\beta}_j &= \int u^2(x_j)\frac{p_{1,j}(x_1,x_j)}{p_1(x)}dx_1dx_j \\ &\le \frac{B}{b}\int K(u)du\int u^2(x_j)dx_j \le Cm^{-1}\sum_{k=1}^m\beta_k^2.\end{aligned} \tag{D.10}$$

Let $\mathbb{P}_n = n^{-1}\sum_{i=1}^n \boldsymbol{\delta}_{X_{i1},X_{ij}}$. We write the integration as

$$\sup_{z\in\mathcal{X}}\int K_h(x_1 - z)u^2(x_j)d\mathbb{P}_n = I_1 + I_2, \text{ where}$$

$$I_1 = \sup_{z\in\mathcal{X}} \int K_h(x_1-z)u^2(x_j)d\mathbb{P}_{X_1,X_j} \text{ and } I_2 = \sup_{z\in\mathcal{X}} \left| \int K_h(x_1-z)u^2(x_j)d(\mathbb{P}_n - \mathbb{P}_{X_1,X_j}) \right|.$$

Due to (D.9), we have $I_1 = O_P(m^{-1})\|\boldsymbol{\beta}_j\|_2^2$ and a similar argument to one in Lemma 6.2 of Zhou et al. (1998) will derive $I_2 = o(h)\|\boldsymbol{\beta}_j\|_2^2$. This completes the proof. □

Based on Lemma D.1, the remaining step is to prove Lemma 7.1.

*Proof of Lemma 7.1.* We can derive the restricted eigenvalue condition on the cone from Lemma D.1. We apply Lemma D.1 in the last step. If the cone condition

$$\sum_{j\in S^c} \|\boldsymbol{\beta}_j\|_2 \le 3\sum_{j\in S} \|\boldsymbol{\beta}_j\|_2 + 3\sqrt{m}|\alpha|$$

is satisfied, by Hölder inequality, we have the upper bound

$$\|\boldsymbol{\beta}\|_{1,2} \le 4\sum_{j\in S} \|\boldsymbol{\beta}_j\|_2 + 3\sqrt{m}|\alpha| \le 4\sqrt{s}\|\boldsymbol{\beta}\|_2 + 3\sqrt{m}|\alpha|.$$

With large probability, we have the following inequality

$$\begin{aligned}
\boldsymbol{\theta}^T\widehat{\boldsymbol{\Sigma}}_z\boldsymbol{\theta} &\ge \boldsymbol{\theta}^T\boldsymbol{\Sigma}_z\boldsymbol{\theta} - C\left(\sqrt{\frac{m\log(dm)}{nh}} + \frac{m}{nh} + \frac{1}{\sqrt{nh}} + h^2\right)\|\boldsymbol{\theta}\|_{1,2}^2 \\
&\ge \rho_{\min}|\alpha|^2 + \rho_{\min}\|\boldsymbol{\beta}\|_2/m - C\left(\sqrt{\frac{m\log(dm)}{nh}} + \frac{m}{nh} + \frac{1}{\sqrt{nh}} + h^2\right)(4m|\alpha|^2 + 4s\|\boldsymbol{\beta}\|_2^2) \\
&\ge \rho_{\min}|\alpha|^2/2 + \rho_{\min}m^{-1}\|\boldsymbol{\beta}\|_2^2/2
\end{aligned}$$

for any $z\in\mathcal{X}$ and sufficiently large $n$ if $s\sqrt{m^3\log(dm)/(nh)} + sm^2/(nh) = o(1)$. □

## D.2 Proof of Lemma 7.2

The proof can be separated into two cases: $j = 1$ and $j \ge 2$. For the simplicity of notation, we write $\delta_i(z)$ as $\delta_i$ in this proof. We first consider the situation when $j \ge 2$ and prove (7.1) and (7.3). From

Lemma 7.1,

$$\sup_{z\in\mathcal{X}} \|\mathbf{W}_z^{1/2}\mathbf{\Psi}_{\bullet j}\mathbf{\Psi}_{\bullet j}^T\mathbf{W}_z^{1/2}\|_2/\sqrt{n} = \sup_{z\in\mathcal{X}} \|\mathbf{\Psi}_{\bullet j}\mathbf{W}_z\mathbf{\Psi}_{\bullet j}^T\|_2/\sqrt{n} \le \rho_{\max} m^{-1/2}$$

with high probability. Therefore

$$\sup_{z\in\mathcal{X}} \frac{1}{n}\|\mathbf{\Psi}_{\bullet j}^T\mathbf{W}_z\boldsymbol{\delta}\|_2 \le \sup_{z\in\mathcal{X}} \frac{1}{n}\|\mathbf{W}_z^{1/2}\mathbf{\Psi}_{\bullet j}\mathbf{\Psi}_{\bullet j}^T\mathbf{W}_z^{1/2}\|_2\|\mathbf{W}_z^{1/2}\boldsymbol{\delta}\|_2 \le \frac{C}{\sqrt{m}} \cdot \sup_{z\in\mathcal{X}} \frac{1}{\sqrt{n}}\|\mathbf{W}_z^{1/2}\boldsymbol{\delta}\|_2. \quad \text{(D.11)}$$

To complete the proof, we need a bound on

$$\sup_{z\in\mathcal{X}} \frac{1}{n}\|\mathbf{W}_z^{1/2}\boldsymbol{\delta}\|_2^2 = \sup_{z\in\mathcal{X}} \frac{1}{n}\sum_{i=1}^n K_h(X_{i1}-z)\delta_i^2. \quad \text{(D.12)}$$

Using Equation (20) in Zhou et al. (1998) on B-spline, we have

$$\delta_i^2 = \Big(\sum_{j=2}^d f_{jz}(X_{ji}) - f_{nj;z}(X_{ji})\Big)^2 \le s^2 m^{-4}.$$

Define the following empirical process

$$U_n(z) = \frac{1}{n}\sum_{i=1}^n K_h(X_{i1}-z)\delta_i^2 - \mathbb{E}\big[K_h(X_{11}-z)\delta_1^2\big].$$

Applying Hoeffding's inequality (Hoeffding, 1963), we have

$$\mathbb{P}\left(\sup_{z\in\mathcal{X}} U_n(z) - \mathbb{E}\left[\sup_{z\in\mathcal{X}} U_n(z)\right] > t\right) \le \exp\left(-C\frac{nh^2t^2}{(sm^{-4})^2}\right). \quad \text{(D.13)}$$

Let

$$\mathcal{G}_h'' = \left\{g_z(x_1,x_2) = h^{-1}K(h^{-1}(x_1-z))\delta^2(x_2) \,|\, z\in\mathcal{X}, x_1\in\mathcal{X}, x_2\in\mathcal{X}^{d-1}\right\},$$

where $\delta(x_2) = \sum_{j=2}^d f_j(x_{2j}) - f_{nj}(x_{2j})$. Similar to the covering number of $\mathcal{G}_h$ in (D.1), since $\delta^2(x_2) \le sm^{-4}$ for any $x_2$, we have for any measure $Q$,

$$\sup_Q N\left(\mathcal{G}_h'', L^2(Q), \epsilon\right) \le \left(\frac{2\sqrt{s}||K||_{\mathrm{TV}}A}{m^2h\epsilon}\right)^4.$$

Furthermore, $\sigma_P^2 := \mathbb{E}[K_h(X_{i1}-z)\delta_i^2]^2 = O((sm^{-4})^2h^{-1})$. Since $g \le U := Ch^{-1}(sm^{-4})$ for any $g \in \mathcal{G}_h''$ and $m^4(sn)^{-1} = o(1)$, we have $n\sigma_P^2 \ge C_1U^2\log(C_2\sqrt{sm^{-4}}U/\sigma)$. By Lemma F.2, we have

$$\mathbb{E}\left[\sup_{z\in\mathcal{X}} U_n(z)\right] \le C\frac{sm^{-4}}{\sqrt{nh}}\sqrt{\log(m^2/\sqrt{sh})}. \tag{D.14}$$

We set $t = Cs(m^4h)^{-1}\sqrt{\log n/n}$ in (D.13) and combine it with (D.14) to obtain that, with probability at least $1-1/n$ ,

$$\sup_{z\in\mathcal{X}} U_n(z) \le C\frac{sm^{-4}}{\sqrt{nh}}\sqrt{\log\left(m^2/\sqrt{sh}\right)} + C\frac{s\sqrt{\log n/n}}{m^4h}. \tag{D.15}$$

Finally, we bound the maximal of the expectation by

$$\begin{aligned}\sup_{z\in\mathcal{X}}\mathbb{E}[K_h(X_{11}-z)\delta_1^2] &= \sup_{z\in\mathcal{X}}\int K_h(t-z)dP_{X_1}(t)\delta^2(u)dP_{X_{>2}|X_1=t}(u)\\ &\le Csm^{-4}\sup_{z\in\mathcal{X}}\int K_h(t-z)dP_{X_1}(t) \le Csm^{-4}.\end{aligned} \tag{D.16}$$

Combining (D.15) and (D.16), with probability at least $1-1/n$, we have

$$\begin{aligned}\sup_{z\in\mathcal{X}}\frac{1}{n}\|\mathbf{W}_z^{1/2}\boldsymbol{\delta}\|_2^2 &= \sup_{z\in\mathcal{X}}\frac{1}{n}\sum_{i=1}^n K_h(X_{i1}-z)\delta_i^2\\ &\le \sup_{z\in\mathcal{X}}U_n(z) + \sup_{z\in\mathcal{X}}\mathbb{E}[K_h(X_{11}-z)\delta_1^2]\\ &\le C\frac{sm^{-4}}{\sqrt{nh}}\sqrt{\log(m^2/\sqrt{sh})} + C\frac{s\sqrt{\log n/n}}{m^4h} + Csm^{-4}\\ &= O(sm^{-4}),\end{aligned} \tag{D.17}$$

where the last equality is due to $2/\sqrt{nh^2} = o(1)$. Therefore, we prove the upper bound in (7.3). Comibing (D.17) with (D.11), we have we can conclude that

$$\sup_{z\in\mathcal{X}}\max_{j\ge 2}\frac{1}{n}\|\boldsymbol{\Psi}_{\bullet j}^T\mathbf{W}_z\boldsymbol{\delta}\|_2 \le \frac{C}{\sqrt{m}}\cdot\sup_{z\in\mathcal{X}}\frac{1}{\sqrt{n}}\|\mathbf{W}_z^{1/2}\boldsymbol{\delta}\|_2 \le C\sqrt{\frac{s}{m^5}}.$$

This gives us the rate in (7.1). Therefore, we complete the proof by bounding all the three inequality in (7.1), (7.2) and (7.3).

The final step is to prove (7.2). Recall that $\boldsymbol{\Psi}_{\bullet 1} = (1,\dots,1)^T$. For the case $j=1$, following the proof for (7.3). According to (D.12), we have $|\delta_i| \le sm^{-2}$ for any $i \in [n]$. Let

$$U_n'(z) = \frac{1}{n}\sum_{i=1}^n K_h(X_{i1}-z)\delta_i - \mathbb{E}[K_h(X_{11}-z)\delta_1].$$

We use Hoeffding's inequality (Hoeffding, 1963) again and obtain

$$\mathbb{P}\left(\sup_{z\in\mathcal{X}} U_n'(z) - \mathbb{E}\left[\sup_{z\in\mathcal{X}} U_n'(z)\right] > t\right) \le \exp\left(-C\frac{nh^2t^2}{sm^{-4}}\right). \tag{D.18}$$

Applying symmetrization inequality again, we have

$$\mathbb{E}\left[\sup_{z\in\mathcal{X}} U_n'(z)\right] \le 2\mathbb{E}\bigg[\sup_{z\in\mathcal{X}} \frac{1}{n}\sum_{i=1}^n \xi_i K_h(X_{i1}-z)|\delta_i|\bigg],$$

where $\{\xi_i\}_{i=1}^n$ are i.i.d. Rademacher variables independent of data. Let

$$\widetilde{\mathcal{G}}_h'' = \left\{g_z(x_1,x_2) = h^{-1}K(h^{-1}(x_1-z))\delta(x_2)\,|\, z\in\mathcal{X}, x_1\in\mathcal{X}, x_2\in\mathcal{X}^{d-1}\right\},$$

where $\delta(x_2) = \big|\sum_{j=2}^d f_j(x_{2j}) - f_{nj}(x_{2j})\big|$. Just as the covering number of $\mathcal{G}_h''$, we also have $\delta(x_2) \le sm^{-2}$ for any $x_2$, we have for any measure $Q$,

$$\sup_Q N\left(\widetilde{\mathcal{G}}_h'', L^2(Q), \epsilon\right) \le \left(\frac{2s^{1/2}||K||_{\mathrm{TV}}A}{m^2h\epsilon}\right)^4.$$

The variance of the process $\sigma_P^2 := \mathbb{E}[K_h(X_{i1}-z)\delta_i]^2 = O(sm^{-4}h^{-1})$. Since $g \le U := Ch^{-1}(sm^{-4})^{1/2}$ for any $g \in \mathcal{G}_h''$ and $m^4(sn)^{-1} = o(1)$, we have $n\sigma_P^2 \ge C_1U^2\log(C_2s^{1/4}m^{-1}U/\sigma)$. Applying Lemma F.2 again, we have

$$\mathbb{E}\left[\sup_{z\in\mathcal{X}} U_n'(z)\right] \le C\frac{\sqrt{s}m^{-2}}{\sqrt{nh}}\sqrt{\log(m/\sqrt{sh})}. \tag{D.19}$$

We let $t = C\sqrt{s}(m^2h)^{-1}\sqrt{\log n/n}$ in (D.18) and use it with (D.19). Therefore, we achieve with probability at least $1 - 1/n$ ,

$$\sup_{z \in \mathcal{X}} U_n'(z) \le C \frac{\sqrt{s} m^{-2}}{\sqrt{nh}} \sqrt{\log\left(m^2/\sqrt{sh}\right)} + C \frac{\sqrt{s \log n/n}}{m^2 h}. \tag{D.20}$$

We again bound the supreme of the expectation

$$\begin{aligned} \sup_{z \in \mathcal{X}} \mathbb{E}[K_h(X_{11} - z)\delta_1] &= \sup_{z \in \mathcal{X}} \int K_h(t - z) dP_{X_1}(t) \delta(u) dP_{X_{>2}|X_1 = t}(u) \\ &\le C\sqrt{s} m^{-2} \sup_{z \in \mathcal{X}} \int K_h(t - z) dP_{X_1}(t) \le C\sqrt{s} m^{-2}. \end{aligned} \tag{D.21}$$

Combining (D.20) and (D.21), with probability at least $1 - 1/n$, we have

$$\begin{aligned} \sup_{z \in \mathcal{X}} \frac{1}{n} |\boldsymbol{\Psi}_{\bullet 1}^T \mathbf{W}_z \boldsymbol{\delta}_z| &\le \sup_{z \in \mathcal{X}} \frac{1}{n} \sum_{i=1}^n K_h(X_{i1} - z)|\delta_i| \\ &\le \sup_{z \in \mathcal{X}} U_n'(z) + \sup_{z \in \mathcal{X}} \mathbb{E}[K_h(X_{11} - z)\delta_1] = O(\sqrt{s} m^{-2}). \end{aligned}$$

Therefore, we prove the upper bound in (7.2) which completes the proof of the lemma.

## D.3 Proof of Lemma 7.3

For $j \ge 2$, we bound the two terms $\sup_{z \in \mathcal{X}} \max_{j \ge 2} \frac{1}{n} \|\boldsymbol{\Psi}_{\bullet j}^T \mathbf{W}_z \boldsymbol{\xi}_z\|_2$ and $\sup_{z \in \mathcal{X}} \max_{j \ge 2} \frac{1}{n} \|\boldsymbol{\Psi}_{\bullet j}^T \mathbf{W}_z \boldsymbol{\zeta}_z\|_2$ separately. To bound the first term, let $\Delta f_z(x) = f_1(x) - f_1(z)$ and $\boldsymbol{\Psi}_{ij}$ be the $i$th row of $\boldsymbol{\Psi}_{\bullet j}$. We can rewrite the suprema as

$$\sup_{z \in \mathcal{X}} \max_{j \ge 2} \frac{1}{n} \|\boldsymbol{\Psi}_{\bullet j}^T \mathbf{W}_z \boldsymbol{\xi}_z\|_2 = \max_{j \ge 2} \sup_{z \in \mathcal{X}} \sup_{\mathbf{v} \in \mathbb{B}^m} \frac{1}{n} \sum_{i=1}^n K_h(X_{i1} - z) \Delta f_z(X_{1i}) \mathbf{v}^T \boldsymbol{\Psi}_{ij}. \tag{D.22}$$

Let $N_v = \{\mathbf{v}_1, \ldots, \mathbf{v}_M\}$ be a 1/2-covering of the sphere $\mathbb{B}^m = \{\mathbf{v} \in \mathbb{R}^m \mid \|\mathbf{v}\|_2 \le 1\}$. Observe that for any $\mathbf{v} \in \mathbb{B}^m$, there exists $\pi(\mathbf{v}) \in N_v$ such that $\|\mathbf{v} - \pi(\mathbf{v})\|_2 \le 1/2$. Therefore we have

$$
\begin{aligned}
&\sup_{\mathbf{v}\in\mathbb{B}^m} \frac{1}{n}\sum_{i=1}^n K_h(X_{i1}-z)\Delta f_z(X_{1i})\mathbf{v}^T\mathbf{\Psi}_{ij} \\
&\le \sup_{k\in[M]} \frac{1}{n}\sum_{i=1}^n K_h(X_{i1}-z)\Delta f_z(X_{1i})\mathbf{v}_k^T\mathbf{\Psi}_{ij} + \sup_{\mathbf{v}\in\frac{1}{2}\mathbb{B}^m} \frac{1}{n}\sum_{i=1}^n K_h(X_{i1}-z)\Delta f_z(X_{1i})\mathbf{v}^T\mathbf{\Psi}_{ij} \\
&\le \sup_{k\in[M]} \frac{1}{n}\sum_{i=1}^n K_h(X_{i1}-z)\Delta f_z(X_{1i})\mathbf{v}_k^T\mathbf{\Psi}_{ij} + \frac{1}{2}\sup_{\mathbf{v}\in\mathbb{B}^m} \frac{1}{n}\sum_{i=1}^n K_h(X_{i1}-z)\Delta f_z(X_{1i})\mathbf{v}^T\mathbf{\Psi}_{ij}.
\end{aligned}
$$

If we move the second term on the right hand side of the last inequality to the left hand side, we obtain the inequality that

$$
\sup_{\mathbf{v}\in\mathbb{B}^m} \frac{1}{n}\sum_{i=1}^n K_h(X_{i1}-z)\Delta f_z(X_{1i})\mathbf{v}^T\mathbf{\Psi}_{ij} \le 2\sup_{k\in[M]} \frac{1}{n}\sum_{i=1}^n K_h(X_{i1}-z)\Delta f_z(X_{1i})\mathbf{v}_k^T\mathbf{\Psi}_{ij}.
$$

Therefore, in order to bound (D.22), we need to study the following empirical process

$$
V_n(z) = \max_{j\ge 2, k\in[M]} \sup_{z\in\mathcal{X}} \left\{ \frac{1}{n}\sum_{i=1}^n K_h(X_{1i}-z)\Delta f_z(X_{1i})\mathbf{v}_k^T\mathbf{\Psi}_{ij} - \mathbb{E}[K_h(X_{11}-z)\Delta f_z(X_{11})\mathbf{v}_k^T\mathbf{\Psi}_{ij}] \right\}.
$$

We define the following function class

$$
\mathcal{G}_h''' = \left\{ g_z(x_1, x_2) = h^{-1}K((x_1 - z)/h)\Delta f_z(x_1)\sum_{t=1}^m \mathbf{v}_{kt}\psi_t(x_j) \,\Big|\, j \ge 2, k\in[M], z\in\mathcal{X} \right\}
$$

and, similarly to argument in the covering number of $\mathcal{G}_h$ in (D.1), we have

$$
\sup_Q N\left(\mathcal{G}_h''', L^2(Q), \epsilon\right) \le dM\left(\frac{2\sqrt{m}||K||_{\mathrm{TV}}A}{h\epsilon}\right)^4.
$$

From (D.10), we bound the maximal of the expectation by

$$
\sup_{x\in\mathcal{X}} \mathbb{E}\big[(\mathbf{v}_k^T\mathbf{\Psi}_{ij})^2 \mid X_1 = x\big] \le C\|\mathbf{v}_k\|^2 m^{-1}.
$$

Furthermore, we can bound the variance by expanding the expectation as the integration and applying the Taylor expansion as follows

$$\begin{aligned}\sigma_P^2 &:= \mathbb{E}[K_h(X_1 - z)\Delta f_z(X_1)\mathbf{v}_k^T \mathbf{\Psi}_{ij}]^2 \\ &= h^{-2} \int K^2\left(\frac{x-z}{h}\right)(f_1(x) - f_1(z))^2 p_{X_1}(x)dx \cdot \mathbb{E}\left[(\mathbf{v}_k^T \mathbf{\Psi}_{ij})^2 \,|\, X_1 = x\right] \\ &\leq C(mh)^{-1} \int K^2(u)(f_1(z+hu) - f_1(z))^2 p_{X_1}(z+hu)du \\ &= C(mh)^{-1} \int K^2(u)(f_1'(z)uh + o(uh))^2(p_{X_1}(z) + p'_{X_1}(z)uh + o(uh))du \\ &= Cm^{-1}[f'(z)]^2 p_{X_1}(z) \int u^2 K^2(u)du \cdot h + o(m^{-1}h) = Chm^{-1}.\end{aligned}$$

The uniform upper bound of $K_h(x-z)\Delta f_z(x)$ can be studied under two cases: (1) $x$ is out of the support and (2) $x$ is in the support. In particular, we have

- if $x \notin [z-h, z+h]$, then $K_h(x-z)\Delta f_z^2(x) = 0$;
- if $x \in [z-h, z+h]$, then, by mean value theorem,

$$K_h(x-z)\Delta f_z(x) \leq h^{-1}K(h^{-1}(x-z))|f(x) - f(z)| \leq h^{-1}\|K\|_\infty \|f'\|_\infty^2 \cdot (2h) = 4\|f_1'\|_\infty^2 \|K\|_\infty.$$

Combining with the fact that $|\mathbf{v}_k^T \mathbf{\Psi}_{ij}| \leq \sqrt{m}$ for any $i, j, k$, we conclude that $g \leq U := 4\|f_1'\|_\infty^2 \|K\|_\infty \sqrt{m}$ for any $g \in \mathcal{G}_h'''$. Therefore by Lemma F.2 and $M = 6^m$, we have

$$\begin{aligned}\mathbb{E}V_n(z) &\leq C\sqrt{\frac{h\log(dMh^{-1})}{mn}} + C\frac{\sqrt{m}\cdot\log(dMh^{-1})}{n} \\ &= C\sqrt{\frac{h\log(dh^{-1})}{n}} + C\frac{m^{3/2}\log(dh^{-1})}{n}.\end{aligned} \tag{D.23}$$

Similar to the analysis of $\sigma_P^2$, we also expand the expectation of the process as the integration and use the Taylor expansion to bound it as follows

$$\begin{aligned}
&\mathbb{E}\big[K_h(X_1-z)\Delta f_z(X_1)\mathbf{v}_k^T\boldsymbol{\Psi}_{ij}\big] \\
&\quad= h^{-1}\int K\left(\frac{x-z}{h}\right)(f_1(x)-f_1(z))p_{X_1}(x)dx\cdot\mathbb{E}\left[\mathbf{v}_k^T\boldsymbol{\Psi}_{1j}\,|\,X_1=x\right]dx \\
&\quad= \int K(u)\big[f_1'(z)uh+f_1''(z)(uh)^2/2+o(uh)^2\big]\big[p_{X_1}(z)+p_{X_1}'(z)uh+o(uh)\big] \\
&\qquad\cdot\left(\mathbb{E}\left[\mathbf{v}_k^T\boldsymbol{\Psi}_{1j}\,|\,X_1=z\right]+uh\frac{d}{dz}\mathbb{E}\left[\mathbf{v}_k^T\boldsymbol{\Psi}_{1j}\,|\,X_1=z\right]+o(uh)\right)du\le Ch^2/\sqrt{m}.
\end{aligned}\tag{D.24}$$

The last inequality is due to the fact that $K(\cdot)$ is an even function, $\|\psi_{jk}\|_\infty\le 1$ and from (D.10). Moreover, the constant is independent to $j,k$ and $z$. Using Lemma F.4, we have

$$\mathbb{P}\left(V_n(z)-\mathbb{E}V_n(z)>t\sqrt{2(\sigma_P^2+2U\mathbb{E}V_n(z))}+\frac{2Ut^2}{3}\right)\le\exp\left(-nt^2\right).\tag{D.25}$$

Combining (D.23) and (D.24) with (D.25) for $t=\sqrt{\log n/n}$, with probability at least $1-1/n$, we have

$$\begin{aligned}
\sup_{z\in\mathcal{X}}\max_{j\ge 2}\frac{1}{n}\|\boldsymbol{\Psi}_{\bullet j}^T\mathbf{W}_z\boldsymbol{\xi}_z\|_2 &\le 2\max_{j\ge 2,k\in[M]}\sup_{z\in\mathcal{X}}\frac{1}{n}\sum_{i=1}^n K_h(X_{1i}-z)\Delta f_z(X_{1i})\mathbf{v}_k^T\boldsymbol{\Psi}_{ij} \\
&\le V_n(z)+\max_{j,k}\sup_{z\in\mathcal{X}}\mathbb{E}[K_h(X_1-z)\Delta f_z(X_1)\mathbf{v}_k^T\boldsymbol{\Psi}_{ij}] \\
&\le C\sqrt{\frac{h\log(dh^{-1})}{n}}+C\frac{m^{3/2}\log(dh^{-1})}{n}+C\frac{h^2}{\sqrt{m}},
\end{aligned}\tag{D.26}$$

where the last equality is because of $n^{-1}h=o(1)$.

Now we bound $\sup_{z\in\mathcal{X}}\max_{j\ge 2}\frac{1}{n}\|\boldsymbol{\Psi}_{\bullet j}^T\mathbf{W}_z\boldsymbol{\zeta}_z\|_2$. The procedure is similar to the first part of the proof. We again apply the 1/2-covering of $\mathbb{B}^d$ so that

$$\sup_{z\in\mathcal{X}}\max_{j\ge 2}\frac{1}{n}\|\boldsymbol{\Psi}_{\bullet j}^T\mathbf{W}_z\boldsymbol{\zeta}_z\|_2\le 2\max_{j\ge 2,k\in[M]}\sup_{z\in\mathcal{X}}\frac{1}{n}\sum_{i=1}^n K_h(X_{i1}-z)\zeta_i(z)\mathbf{v}_k^T\boldsymbol{\Psi}_{ij}.$$

Motived by the above argument, we now turn to study the following empirical process

$$V_n'(z) = \max_{j\geq 2, k\in[M]} \sup_{z\in\mathcal{X}} \left\{ \frac{1}{n}\sum_{i=1}^{n} K_h(X_{1i}-z)\zeta_i(z)\mathbf{v}_k^T\mathbf{\Psi}_{ij} - \mathbb{E}[K_h(X_{11}-z)\zeta_i(z)\mathbf{v}_k^T\mathbf{\Psi}_{ij}] \right\}$$

and the function class inspired from the above empirical process

$$\mathcal{G}_h'''' = \left\{ g_z(x_1,x_2) = h^{-1}K((x_1-z)/h)\zeta_i(z)\sum_{t=1}^{m}\mathbf{v}_{kt}\psi_t(x_j) \,\middle|\, j\geq 2, k\in[M], z, x_1, x_j\in\mathcal{X} \right\}.$$

Our method of bound the supreme of the process is same as the proof of previous lemmas. We need to study the covering number of the function space. Assembling the concentration inequality of the suprema with the upper bound of the expectation of suprema and suprema of the expectation, we will arrive at the final bound. Therefore, we first bound the covering number

$$\sup_Q N\left(\mathcal{G}_h'''', L^2(Q), \epsilon\right) \leq dM\left(\frac{2\sqrt{m}||K||_{\mathrm{TV}}A}{h\epsilon}\right)^4.$$

Using Definition 4.1, there exists $L_1(z, x_{\backslash 1})$ such that the approximation error is bounded by

$$|\zeta_i(z)| = \left| f(X_{i1},\ldots .X_{id}) - \sum_{j=1}^{d} f_{jz}(X_{ij}) \right| \leq |L_1(z,x_{\backslash 1})|\cdot|X_{i1}-z| + |U_j(z)|(X_{i1}-z)^2.$$

Therefore, the variance of the process $V_n'$ can be bounded by computing the expectation

$$\sigma_P^2 := \mathbb{E}[K_h(X_1-z)\zeta_i(z)\mathbf{v}_k^T\mathbf{\Psi}_{ij}]^2 \leq Chm^{-1} \tag{D.27}$$

and $g\leq U := 4\|L_1(z,x_{\backslash 1})\|_\infty^2\|K\|_\infty\sqrt{m}$ for all $g\in\mathcal{G}_h''''$. Lemma F.2 gives us

$$\mathbb{E}V_n'(z) \leq C\sqrt{\frac{h\log(dh^{-1})}{n}} + C\frac{m^{3/2}\log(dh^{-1})}{n}. \tag{D.28}$$

Denote $\kappa(x) = \mathbb{E}[L_1(z, X_{\setminus 1}) \,|\, X_1 = x]$. Using Definition 4.1,

$$
\begin{aligned}
&\mathbb{E}[K_h(X_1 - z)\zeta_i(z)\mathbf{v}_k^T \boldsymbol{\Psi}_{ij}] \\
&\quad = h^{-1} \int K\left(\frac{x-z}{h}\right) (\mathbb{E}[L_1(z, X_{\setminus 1}) \,|\, X_1 = x](x_1 - z) + U_j(z)(x-z)^2) \\
&\qquad\qquad \cdot \mathbb{E}\left[\mathbf{v}_k^T \boldsymbol{\Psi}_{1j} \,|\, X_1 = x\right] p_{X_1}(x) dx \\
&\quad = \int K(u)(\kappa(z)uh + (\kappa''(z) + \|U_j\|_\infty)(uh)^2/2 + o(uh)^2)(p_{X_1}(z) + p'_{X_1}(z)uh + o(uh)) \\
&\qquad \cdot \left(\mathbb{E}\left[\mathbf{v}_k^T \boldsymbol{\Psi}_{1j} \,|\, X_1 = z\right] + uh\frac{d}{dz}\mathbb{E}\left[\mathbf{v}_k^T \boldsymbol{\Psi}_{1j} \,|\, X_1 = z\right] + o(uh)\right) du \le Ch^2/\sqrt{m}. \qquad \text{(D.29)}
\end{aligned}
$$

The last inequality is due to Assumption (**A1**). Since $\mathcal{X}$ is compact, $\mathbb{E}\left[\mathbf{v}_k^T \boldsymbol{\Psi}_{1j} \,|\, X_1 = z\right]$ and $\kappa(z)$ are uniformly bounded on $z \in \mathcal{X}$. According to Lemma F.2 and Lemma F.4, similar to the first part of the proof, (D.27), (D.28) and (D.29) can yield that for some constant $C$, with probability at least $1 - 1/d$, we can bound the suprema

$$
\sup_{z\in\mathcal{X}} \max_{j\ge 2} \frac{1}{n} \|\boldsymbol{\Psi}_{\bullet j}^T \mathbf{W}_z \boldsymbol{\zeta}_z\|_2 \le C\sqrt{\frac{h\log(dh^{-1})}{n}} + C\frac{m^{3/2}\log(dh^{-1})}{n} + C\frac{h^2}{\sqrt{m}}. \qquad \text{(D.30)}
$$

Combining (D.26) and (D.30), we have the rate of $\sup_{z\in\mathcal{X}} \max_{j\ge 2} \frac{1}{n}\|\boldsymbol{\Psi}_{\bullet j}^T \mathbf{W}_z(\boldsymbol{\xi}_z + \boldsymbol{\zeta}_z)\|_2$.

For the case when $j = 1$, $\boldsymbol{\Psi}_{\bullet 1} = (1, \ldots, 1)^T \in \mathbb{R}^n$ and we can follow similar procedure to derive

$$
\sup_{z\in\mathcal{X}} \frac{1}{n} \left\|\boldsymbol{\Psi}_{\bullet 1}^T \mathbf{W}_z(\boldsymbol{\xi}_z + \boldsymbol{\zeta}_z)\right\|_2 = \sup_{z\in\mathcal{X}} \frac{1}{n} \sum_{i=1}^n K_h(z - X_{i1})\big(\xi_i(z) + \zeta_i(z)\big) = O_P\left(h^2 + \sqrt{h/n}\right).
$$

The final step is to bound $\sup_{z\in\mathcal{X}} \frac{1}{n}\|W_z^{1/2}\boldsymbol{\xi}_z^2\|_2^2$ and $\sup_{z\in\mathcal{X}} \frac{1}{n}\|W_z^{1/2}\boldsymbol{\zeta}_z^2\|_2^2$. We just repeat the procedure again and consider $V_n''''(z) = \sup_{z\in\mathcal{X}} n^{-1}\sum_{i=1}^n K_h(X_{1i} - z)\xi_i(z) - \mathbb{E}[K_h(X_{11} - z)\xi_i(z)]$. First, we find that

$$
\mathbb{E}V_n''''(z) \le C\sqrt{\frac{h^3\log(h^{-1/2})}{n}} + C\frac{h\log(h^{-1})}{n}. \qquad \text{(D.31)}
$$

Next, we have the upper bound of the supreme of the expectation

$$
\begin{aligned}
\sup_{z\in\mathcal{X}} \mathbb{E}[K_h(X_1 - z)\xi_i^2(z)] &= h^{-1}\int K\left(\frac{x-z}{h}\right)(f_1(x) - f_1(z))^2 p_{X_1}(x)dx \\
&= \int K(u)(f_1'(z)uh + o(uh))^2(p_{X_1}(z) + p_{X_1}'(z)uh + o(uh))du \\
&= [f'(z)]^2 p_{X_1}(z)\int u^2K(u)du\cdot h^2 + o(h^2) \le Ch^2. \qquad \text{(D.32)}
\end{aligned}
$$

Combining (D.31) and (D.32) with Lemma F.4 with $t = \log n/\sqrt{n}$, with probability at least $1 - 1/n$,

$$
\begin{aligned}
\sup_{z\in\mathcal{X}} \frac{1}{n}\sum_{i=1}^n K_h(X_{i1} - z)\xi_i^2(z) &\le V_n(z) + \sup_{z\in\mathcal{X}} \mathbb{E}[K_h(X_{11} - z)\xi_i^2(z)] \\
&\le C\sqrt{\frac{h^3\log(nh^{-1})}{n}} + C\frac{h^{5/4}\log(nh^{-1})}{n^{3/4}} + C\frac{h\log^2 n}{n} + Ch^2 = O\left(h^2\right).
\end{aligned}
$$

Similarly, we also have $\sup_{z\in\mathcal{X}} n^{-1}\sum_{i=1}^n K_h(X_{i1} - z)\zeta_i^2(z) = o_P(h^2)$.

## D.4 Proof of Lemma 7.4

For $j = 2, \ldots, n$, we define the process

$$
G_n(z,k,j) = \frac{1}{\sqrt{n}}\sum_{i=1}^n \frac{1}{h}K\left(\frac{X_{i1} - z}{h}\right)\psi_{jk}(X_{ji})\varepsilon_i.
$$

Since $\varepsilon_i$ are subgaussian random variables, we have $\mathbb{P}(\max_i|\varepsilon_i| > C\sqrt{\log n}) \le 1/n$. Conditioning on the event $\mathcal{A} = \{\max_i|\varepsilon_i| < C\sqrt{\log n}\}$, we can apply the Mc'Diarmid's inequality to obtain

$$
\mathbb{P}\left(\max_{j,k}\sup_{z\in\mathcal{X}} G_n(z,k,j) - \mathbb{E}\left[\max_{j,k}\sup_{z\in\mathcal{X}} G_n(z,k,j)\,|\,\mathcal{A}\right] > t\,|\,\mathcal{A}\right) \le \exp\left(-C\frac{nh^2t^2}{\log^2 n}\right). \qquad \text{(D.33)}
$$

Next, we bound $\mathbb{E}\big[\max_{j,k}\sup_{z\in\mathcal{X}} G_n(z,k,j)\,|\,\mathcal{A}\big]$. Using Dudley's entropy integral (see Corollary 2.2.5 in van der Vaart and Wellner (1996)), conditioning on $\{X_{ij}\}_{i\in[n],j\in[d]}$, we have with probability $1 - 1/n$, there exists a constant $C$ such that

$$
\mathbb{E}\left[\max_{1\le j\le d}\max_{1\le k\le m}\sup_{z\in\mathcal{X}} G_n(z,k,j)\,|\,\mathcal{A}\right] \le \mathbb{E}\left[\int_0^{\sigma_n}\sqrt{\log N(\mathcal{G}_h', L^2(\widehat{\mathbb{P}}_n), \epsilon)}d\epsilon\,|\,\mathcal{A}\right],
$$

where $\widehat{\mathbb{P}}_n = n^{-1}\sum_{i=1}^n \delta_{X_{i1},\ldots,X_{id}}$, $\sigma_n = \max_{1\le j\le d}\max_{1\le k\le m}\sup_{z\in\mathcal{X}}\widehat{\mathbb{P}}_n[K_h(\cdot - z)\psi_{jk}(\cdot)]^2$ and

$$\mathcal{G}_h' = \left\{g_z(x_1,x_2) = h^{-1}K(h^{-1}(x_1-z))\psi_{jk}(x_2)\,|\, 1\le k\le m, z\in\mathcal{X}, x_1,x_2 \ \in\mathcal{X}\right\}.$$

From Lemma F.3 and similar to the previous computation on the covering number, for any measure $Q$, we have the uniform upper bound of covering number as

$$\sup_Q N\left(\mathcal{G}_h', L^2(Q),\epsilon\right) \le dm\left(\frac{2||K||_{\mathrm{TV}}A}{h\epsilon}\right)^4.$$

Following a similar argument as in the proof of Lemma 7.1, we bound the variance of process by Cauchy-Schwarz inequality as

$$\begin{aligned}\sigma_P^2 &:= \mathbb{E}\left[\left(K_h\left(X_1-z\right)\psi_{jk}(X_j)\Big|\mathcal{A}\right)^2\right]\\ &\le h^{-2}\mathbb{E}\left[K^2\left(h^{-1}(X_1-z)\right)\mathbb{E}\left[\psi_{jk}^2(X_j)\,|\,X_1\right]\,|\,\mathcal{A}\right] \le b\|K\|_\infty^2(mh)^{-1},\end{aligned}$$

and $g\le\|K\|_\infty h^{-1}$ for any $g\in\mathcal{G}_h'$. Since $m(nh)^{-1} = o(1)$, we have $n\sigma_P^2 \ge h^{-2}\log(dm(2||K||_{\mathrm{TV}}A/(h\sigma)))$. Therefore, by Lemma F.2, we derive that

$$\mathbb{E}\left[\int_0^{\sigma_n}\sqrt{\log N(\mathcal{G}_h', L^2(\widehat{\mathbb{P}}_n),\epsilon)}d\epsilon\,\Big|\,\mathcal{A}\right] \le C\sigma_P\sqrt{\log(dm(h\sigma_P)^{-4})} \le C\sqrt{\frac{\log(dm^3h^{-2})}{mh}}$$
$$\text{and } \mathbb{E}\Big[\max_{j,k}\sup_{z\in\mathcal{X}} G_n(z,k,j)\,\Big|\,\mathcal{A}\Big] \le C\sqrt{\log(dm^3h^{-2})/(mh)}.$$

Choosing $t = C\log^2 n/(\sqrt{n}h)$ in (D.33), we have

$$\begin{aligned}&\mathbb{P}\left(\max_{j,k}\sup_{z\in\mathcal{X}} G_n(z,k,j) > C\sqrt{\frac{\log(dm^3h^{-2})}{mh}} + \frac{\log^2 n}{\sqrt{n}h}\right)\\ &\le \mathbb{P}\left(\max_{j,k}\sup_{z\in\mathcal{X}} G_n(z,k,j) > C\sqrt{\frac{\log(dm^3h^{-2})}{mh}} + \frac{\log^2 n}{\sqrt{n}h}\,\Big|\mathcal{A}\right) + \mathbb{P}(\mathcal{A}^c) \le 2/n.\end{aligned}$$

Therefore, when $m(nh)^{-1} = o(1)$, with probability $1-2/n$,

$$\sup_{z\in\mathcal{X}}\max_{j\ge 2}\frac{1}{n}\|\mathbf{\Psi}_{\bullet j}^T\mathbf{W}_z\boldsymbol{\varepsilon}\|_2 \le \sqrt{\frac{m}{n}}\max_{1\le j\le d}\max_{1\le k\le m}\sup_{z\in\mathcal{X}} G_n(z,k,j) \le 2C\sqrt{\frac{\log(dmh^{-1})}{nh}}.$$

When $j = 1$, recalling that $\mathbf{\Psi}_{\bullet 1} = (1, \dots, 1)^T \in \mathbb{R}^n$, we have

$$\sup_{z \in \mathcal{X}} \frac{1}{n} \|\mathbf{\Psi}_{\bullet 1}^T \mathbf{W}_z \boldsymbol{\varepsilon}\|_2 = \sup_{z \in \mathcal{X}} \left| \frac{1}{n} \sum_{i=1}^n K_h(z - X_{i1}) \varepsilon_i \right|,$$

and, similar to the case when $j \geq 2$, we can show that $\sup_{z \in \mathcal{X}} n^{-1} \|\mathbf{\Psi}_{\bullet 1}^T \mathbf{W}_z \boldsymbol{\varepsilon}\|_2 \leq C \sqrt{\log(h^{-1})/(nh)}$ with probability $1 - 2/n$. This completes the proof.

# E Auxiliary Lemmas for Bootstrap Confidence Bands

In this section, we describe the proof of these technical lemmas used in Section B. Section E.2 to Section E.6 provide the proofs of lemmas in Section B.1 supporting the proof of Theorem 3.4.

## E.1 Proof of Lemma B.2

Recall the rate $r_n$ of the estimated function shown in Theorem 3.2 is

$$r_n := \sqrt{\frac{s^2 \log(dmh^{-1})}{nm^{-2}h}} + \sqrt{\frac{s^3}{m^3}} + \frac{s\log(dh^{-1})}{nm^{-5/2}} + s\sqrt{m}h^2.$$

We first establish a lemma on the estimation error of $\widehat{\varepsilon}_i$.

**Lemma E.1.** Let $\widehat{\varepsilon}_i = Y_i - \widehat{f}(X_{i1}, \dots, X_{id})$ for $i = 1, \dots, n$. Under Assumption (**A4**), we have

$$\mathbb{P}\left( \max_{i \in [n]} |\widehat{\varepsilon}_i - \varepsilon_i| < 2Cr_n\sqrt{m} \right) \geq 1 - \frac{1}{n}.$$

If $h \asymp n^{-\delta}$, $m \asymp n^{\delta}$ for $\delta > 1/5$, we have $r_n\sqrt{m} = o(n^{-1/5})$.

We defer the proof of the lemma to the end of this subsection. With the rate of $\max_{i \in [n]} |\widehat{\varepsilon}_i - \varepsilon_i|$, we can first bound the rate of $\widehat{\sigma}^2 - \sigma^2$. Using the triangle inequality, we have

$$|\widehat{\sigma}^2 - \sigma^2| \leq \underbrace{\frac{1}{n} \sum_{i=1}^n (\widehat{\varepsilon}_i - \varepsilon_i)^2}_{\text{I}} + \underbrace{\frac{2}{n} \sum_{i=1}^n |(\widehat{\varepsilon}_i - \varepsilon_i)\varepsilon_i|}_{\text{II}} + \underbrace{\left| \frac{1}{n} \sum_{i=1}^n \varepsilon_i^2 - \sigma^2 \right|}_{\text{III}}. \tag{E.1}$$

From Lemma E.1, we have the convergence rate of the noise estimator

$$\mathbb{P}(\mathrm{I} > 4cr_n^2 m) \le \mathbb{P}\left(\max_{i\in[n]} |\widehat{\varepsilon}_i - \varepsilon_i|^2 > 4cr_n^2 m\right) \le 1/n. \tag{E.2}$$

Under Assumption (**A4**), $\varepsilon_i$ are subgaussian random variables with variance-proxy $\sigma_\varepsilon^2$. Using Berstein's inequality, we have

$$\mathbb{P}\left(\left|\frac{1}{n}\sum_{i=1}^n \varepsilon_i^2 - \sigma^2\right| > C_1\sqrt{\frac{\sigma_\varepsilon^2 \log n}{n}}\right) \le \frac{2}{n} \quad \text{and} \quad \mathbb{P}\left(\left|\frac{1}{n}\sum_{i=1}^n |\varepsilon_i| - \mathbb{E}|\varepsilon|\right| > c\sqrt{\frac{\sigma_\varepsilon^2 \log n}{n}}\right) \le \frac{2}{n}. \tag{E.3}$$

Suppose $n$ is large enough, so that $\mathbb{E}|\epsilon| \le \sqrt{\sigma_\varepsilon^2 \log n/n}$. We now can bound the second term by

$$\begin{aligned}
&\mathbb{P}(\mathrm{II} > 2c(c+1)\mathbb{E}|\varepsilon| r_n\sqrt{m}) \\
&\le \mathbb{P}\left(\max_{i\in[n]} |\widehat{\varepsilon}_i - \varepsilon_i| > 2cr_n\sqrt{m}\right) + \mathbb{P}\left(\frac{1}{n}\sum_{i=1}^n |\varepsilon_i| > \mathbb{E}|\varepsilon| + c\sqrt{\frac{\sigma_\varepsilon^2 \log n}{n}}\right) \le \frac{3}{n}.
\end{aligned}$$

Applying the fact that $\mathbb{E}|\epsilon_i| \le \sigma^2$, we have the upper bound of the third term as

$$\mathbb{P}\left(\mathrm{III} > 2c(c+1)\sigma^2 r_n\sqrt{m}\right) \le 3/n. \tag{E.4}$$

Combining (E.2), (E.3), (E.4) with (E.1), we have the estimation rate of the variance of noise as

$$\mathbb{P}\left(|\widehat{\sigma}^2 - \sigma^2| \ge C_1 r_n\sqrt{m}\right) \le 6/n. \tag{E.5}$$

Now we come back to prove Lemma E.1.

*Proof of Lemma E.1.* Recall that the estimator of the true function is

$$\widehat{f}(x_1,\ldots,x_d) = \widehat{f}_1(x_1) + \sum_{j=2}^d \sum_{k=1}^m \widehat{\boldsymbol{\beta}}_{jk}\psi_{jk}(x_j).$$

Similar to Lemma 7.2, let $\delta_i = \sum_{j=2}^d f_j(X_{ji}) - f_{mj}(X_{ji})$ and the B-spline theory (see Lemma 1,

Huang et al. (2010)) that $\delta_i^2 \le sm^{-2\gamma}$. Define the event

$$\mathcal{E} = \left\{ \sup_{z\in\mathcal{X}} \left\{ \sqrt{m}|\widehat{a}_z - f_1(z)| + \sum_{i=2}^{d} \|\widehat{\boldsymbol{\beta}}_j - \boldsymbol{\beta}_j\|_2 \right\} \le C r_n \right\}.$$

From Theorem 3.2, we have $\mathbb{P}(\mathcal{E}) \ge 1 - 1/n$. Conditioning the event $\mathcal{E}$, we have

$$\begin{aligned}
\max_{i\in[n]} |\widehat{\varepsilon}_i - \varepsilon_i| &= \max_{i\in[n]} \left| f(X_{i1}, \ldots, X_{id}) - \widehat{f}(X_{i1}, \ldots, X_{id}) \right| \\
&\le \max_{i\in[n]} \left| \widehat{f}_1(X_{i1}) - f_1(X_{i1}) \right| + \max_{i\in[n]} \left| \sum_{j=2}^{d} \sum_{k=1}^{m} (\widehat{\boldsymbol{\beta}}_{jk} - \boldsymbol{\beta}_{jk}) \psi_{jk}(X_{ij}) \right| + \max_{i\in[n]} |\delta_i| \\
&\le \sup_{z\in\mathcal{X}} \left| \widehat{f}_1(z) - f_1(z) \right| + \sqrt{m} \sum_{i=2}^{d} \|\widehat{\boldsymbol{\beta}}_j - \boldsymbol{\beta}_j\|_2 + \sqrt{s} m^{-\gamma} \le 2C\sqrt{m} r_n,
\end{aligned}$$

where the second inequality is because of Hölder inequality as well as the fact that $\psi_{jk} \le 1$ for all $j, k$ and the last inequality is since we are conditioning on $\mathcal{E}$. □

## E.2 Proof of Lemma 3.3

The high level idea of proving Lemma 3.3 is similar to the proof of Lemma D.1. We aim to bound the rate of $\sup_z \|\widehat{\boldsymbol{\Sigma}}_z \boldsymbol{\theta}_z - \boldsymbol{\Sigma}_z \boldsymbol{\theta}_z\|_\infty$. Therefore, we consider the random variable

$$\widetilde{Z}_{kj} = \sup_{z\in\mathcal{X}} \|\widehat{\boldsymbol{\Sigma}}_z \boldsymbol{\theta}_z - \boldsymbol{\Sigma}_z \boldsymbol{\theta}_z\|_\infty = \sup_{z\in\mathcal{X}} (\mathbb{E}_n - \mathbb{E}) \left[ K_h(X_{i1} - z) \psi_{jk}(X_{ij}) \sum_{j'=1}^{d} \sum_{k'=1}^{m} \psi_{j'k'}(X_{ij'}) (\boldsymbol{\theta}_z)_{j'k'} \right].$$

Recall that when $j$ or $k$ equals to 1, $\psi_{jk} \equiv 1$. Similar to the proof of Lemma D.1, we have three cases: (1) $j = k = 1$, (2) only one of $j$ or $k$ equals to 1 and (3) neither of $j, k$ equals to 1. We only analyze the hardest case (3) in this proof and we can deal with the first two cases through a similar procedure. For the minor differences among the analysis of these three cases, we refer to the proof of Lemma D.1.

We first study the covering number of the space

$$\overline{\mathcal{G}}_h = \left\{ h^{-1} K(h^{-1}(x_1 - z)) \psi_{jk}(x_j) \sum_{j'=1}^{d} \sum_{k'=1}^{m} \psi_{j'k'}(x_{j'}) (\boldsymbol{\theta}_z)_{j'k'} \,\Big|\, z \in \mathcal{X}, j, k \in [d] \right\}.$$

Since $\overline{\mathcal{G}}_h$ can be decomposed into a production of a few functions, we aim to apply Lemma F.1 to bound its covering number. Lemma F.3 gives us the covering number of $\{h^{-1}K(h^{-1}(\cdot - z)) \,|\, z \in \mathcal{X}\}$, it remains to bound the covering number of

$$\overline{\mathcal{G}}_h^{(1)} = \left\{ \overline{g}_z(\boldsymbol{x}) := \sum_{j'=1}^{d} \sum_{k'=1}^{m} \psi_{j'k'}(x_{j'})(\boldsymbol{\theta}_z)_{j'k'} \,\middle|\, z \in \mathcal{X} \right\}.$$

Given any $z \in \mathcal{X}$, we can find a $\widetilde{z}$ such that $|z - \widetilde{z}| \le \epsilon$. We then have given any measure $Q$,

$$\begin{aligned}
\|\overline{g}_z - \overline{g}_{\widetilde{z}}\|_{L^2(Q)}^2 &= \mathbb{E}_Q \left[ \sum_{j'=1}^{d} \sum_{k'=1}^{m} \psi_{j'k'}(X_{ij'})[(\boldsymbol{\theta}_z)_{j'k'} - (\boldsymbol{\theta}_{\widetilde{z}})_{j'k'}] \right]^2 \\
&\le L^2 \|\boldsymbol{\theta}_z - \boldsymbol{\theta}_{\widetilde{z}}\|_1^2 \le L^2 d \|\boldsymbol{\theta}_z - \boldsymbol{\theta}_{\widetilde{z}}\|_2^2 \le 2L^2 dm \rho_{\min}^{-2} (B/b) L \rho_{\max} \cdot \epsilon^2,
\end{aligned}$$

where the last inequality is due to Lemma E.4. Therefore,

$$\sup_Q \left( \overline{\mathcal{G}}_h^{(1)}, L^2(Q), \epsilon \right) \le \sqrt{2L^2 dm \rho_{\min}^{-2} (B/b) L \rho_{\max}} / \epsilon. \tag{E.6}$$

According to Corollary 8 in Chapter XI of de Boor (2001), we have

$$\sup_{z, \boldsymbol{x}} \left| \sum_{j'=1}^{d} \sum_{k'=1}^{m} \psi_{j'k'}(x_{j'})(\boldsymbol{\theta}_z)_{j'k'} \right| \le L \sum_{j=1}^{d} \sup_z \|(\boldsymbol{\theta}_z)_{j\bullet}\|_\infty = L\sqrt{d} \sup_z \|\boldsymbol{\theta}_z\|_2 \le 4L\rho_{\min}^{-1} m \sqrt{d}, \tag{E.7}$$

where the last inequality is due to Lemma E.3. By Lemma F.1, combining (E.6), (E.7) and Lemma F.3, we have

$$\sup_Q N\left(\overline{\mathcal{G}}_h, L^2(Q), \epsilon\right) \le d^2 \left( \frac{Cmd}{h\epsilon} \right)^5. \tag{E.8}$$

We then consider the envelop function of $\overline{\mathcal{G}}_h$ as

$$\overline{F}(\boldsymbol{x}) = 4h^{-1} \|K\|_\infty \sup_z \left| \sum_{j'=1}^{d} \sum_{k'=1}^{m} \psi_{j'k'}(x_{j'})(\boldsymbol{\theta}_z)_{j'k'} \right|.$$

In order to study $\overline{F}(\boldsymbol{x})$, we define $\overline{\boldsymbol{\Sigma}} = \mathbb{E}[\boldsymbol{\Psi}_{1\bullet}\boldsymbol{\Psi}_{1\bullet}^T]$ and $\overline{\boldsymbol{\theta}} = \overline{\boldsymbol{\Sigma}}^{-1}\mathbf{e}_1$. We decompose $\overline{F}(\boldsymbol{x})$ into

$$\overline{F}^{(1)}(\boldsymbol{x}) = 4h^{-1}\|K\|_\infty\Big|\sum_{j'=1}^{d}\sum_{k'=1}^{m}\psi_{j'k'}(x_{j'})(\overline{\boldsymbol{\theta}})_{j'k'}\Big| \text{ and}$$
$$\overline{F}^{(2)}(\boldsymbol{x}) = 4h^{-1}\|K\|_\infty\sup_z\Big|\sum_{j'=1}^{d}\sum_{k'=1}^{m}\psi_{j'k'}(x_{j'})(\boldsymbol{\theta}_z - \overline{\boldsymbol{\theta}})_{j'k'}\Big|.$$

According to Lemma E.2, we have

$$\|\overline{F}^{(1)}\|_{L^2(\mathbb{P})}^2 \le 4mh^{-1}\|K\|_\infty\|\overline{\boldsymbol{\theta}}\|_2^2 \le 4h^{-1}\|K\|_\infty\rho_{\min}.$$

Similarly, we also have

$$\|\overline{F}^{(2)}\|_{L^2(\mathbb{P})}^2 \le 4mh^{-1}\|K\|_\infty\sup_z\|\boldsymbol{\theta}_z - \overline{\boldsymbol{\theta}}\|_2^2 \le 8h^{-1}\|K\|_\infty\rho_{\min}.$$

Therefore, we have $\sigma_P^2 \le \|F\|_{L^2(\mathbb{P})}^2 \le 32h^{-1}\|K\|_\infty\rho_{\min}$. By Lemma F.2 we have

$$\mathbb{E}\Big[\max_{k,j}\widetilde{Z}_{kj}\Big] \le C_1\sqrt{\frac{\log^2 d}{nh}}. \tag{E.9}$$

We can also apply Lemma F.5 to obtain

$$\mathbb{P}\Big(\sqrt{n}\max_{k,j}\widetilde{Z}_{kj} \ge 2\sqrt{n}\mathbb{E}[\max_{k,j}\widetilde{Z}_{kj}] + Ch^{-1/2}\sqrt{t} + Ch^{-1/2}t\Big) \le t^{-1}. \tag{E.10}$$

Combining (E.9) with (E.10), we have

$$\max_{k,j}\widetilde{Z}_{kj} = O_P\Big(\sqrt{\log^2 d/nh}\Big).$$

Finally, we finish the proof of the lemma by

$$\big\|\widehat{\boldsymbol{\Sigma}}_z\boldsymbol{\theta}_z - \mathbf{e}_1\big\|_{2,\infty} \le \sqrt{m}\big\|\widehat{\boldsymbol{\Sigma}}_z\boldsymbol{\theta}_z - \mathbf{e}_1\big\|_\infty = O_P\Big(\sqrt{m\log^2 d/nh}\Big).$$

## E.3 Auxiliary Lemmas for Constraint Rate

In this section, we prove some auxiliary lemmas needed in the proof of Lemma 3.3.

**Lemma E.2.** Under Assumptions (**A1**), (**A2**) and (**A6**), there exists a constant $\rho_{\max} < \infty$ such that for any $\boldsymbol{\beta}_+ \in \mathbb{R}^{1+(d-1)m}$,

$$\frac{\boldsymbol{\beta}_+^T \mathbb{E}[\boldsymbol{\Psi}_{1\bullet}\boldsymbol{\Psi}_{1\bullet}^T]\boldsymbol{\beta}_+}{\|\boldsymbol{\beta}_+\|_2^2} \le \frac{3\rho_{\max}}{2m}. \tag{E.11}$$

*Proof.* We first derive some inequalities from (3.8). Denote $\Delta_{jk}(x_j, x_k) := |p_{j,k}(x_j, x_k) - p(x_j)p(x_k)|$. We have

$$\begin{aligned} \sup_k \sum_{j\neq k} \iint \Delta_{jk}(x_j, x_k) dx_j dx_k &\le \sum_{j\neq k} \int \Big| p_{1,j,k}(x_1, x_j, x_k) - p_1(x_1)p_j(x_j)p_k(x_k) \Big| dx_1 \\ &\le \sup_{k\ge 2} \sum_{j\neq k} B \iiint \Big| \frac{p_{1,j,k}(x_1, x_j, x_k)}{p_1(x_1)p_j(x_j)p_k(x_k)} - 1 \Big| dx_1 dx_j dx_k \le \frac{\rho_{\max}}{2}. \end{aligned} \tag{E.12}$$

Following a similar argument, the above inequality also holds when $k = 1$. Given any $j \neq k \ge 2$, let $u_j(x_j) = \sum_{s=1}^m \boldsymbol{\beta}_{js}\psi_{js}(x_j)$ and we have

$$\begin{aligned} \big|\boldsymbol{\beta}_+^T \mathbb{E}[\boldsymbol{\Psi}_{1j}\boldsymbol{\Psi}_{1k}^T]\boldsymbol{\beta}_+\big| &= \Big| \iint u_j(x_j)u_k(x_k)p_{jk}(x_j, x_k)dx_j dx_k \Big| \\ &\le \Big| \int u_j(x_j)p_j(x_j)dx_j \int u_k(x_k)p_k(x_k)dx_k \Big| + \iint \big|u_j(x_j)u_k(x_k)\big|\Delta_{jk}(x_j, x_k)dx_j dx_k \\ &\le m^{-1}\|\boldsymbol{\beta}_j\|_2\|\boldsymbol{\beta}_k\|_2 \cdot \iint \Delta_{jk}(x_j, x_k)dx_j dx_k, \end{aligned}$$

where the last inequality is due to $\mathbb{E}[\psi_{jk}(X_j)] = 0$ and (7.10). Therefore, we have

$$\begin{aligned} \boldsymbol{\beta}_+^T \mathbb{E}[\boldsymbol{\Psi}_{1\bullet}\boldsymbol{\Psi}_{1\bullet}^T]\boldsymbol{\beta}_+ &= \sum_{j=1}^d \boldsymbol{\beta}_j^T \mathbb{E}[\boldsymbol{\Psi}_{1j}\boldsymbol{\Psi}_{1j}^T]\boldsymbol{\beta}_j + \sum_{j\neq k} \boldsymbol{\beta}_j^T \mathbb{E}[\boldsymbol{\Psi}_{1j}\boldsymbol{\Psi}_{1k}^T]\boldsymbol{\beta}_k \\ &\le \frac{\rho_{\max}}{m} \sum_{j=1}^d \|\boldsymbol{\beta}_j\|_2^2 + \sum_{j\neq k} m^{-1}\|\boldsymbol{\beta}_j\|_2\|\boldsymbol{\beta}_k\|_2 \cdot \iint \Delta_{jk}(x_j, x_k)dx_j dx_k \le \frac{3\rho_{\max}}{2m}\|\boldsymbol{\beta}\|_2^2, \end{aligned}$$

where the last inequality is due to (E.12) and the Gershgorin circle theorem. □

**Lemma E.3.** Under Assumptions (**A1**), (**A2**), (**A3**) and (**A6**), there exists a constant $\rho_{\max} < \infty$

such that for any $z \in \mathcal{X}$ and any $\boldsymbol{\beta}_+ \in \mathbb{R}^{1+(d-1)m}$,

$$\frac{\rho_{\min}}{2m} \le \frac{\boldsymbol{\beta}_+^T \boldsymbol{\Sigma}_z \boldsymbol{\beta}_+}{\|\boldsymbol{\beta}_+\|_2^2} \le \frac{3\rho_{\max}}{2m} \text{ and } \sup_z \|\boldsymbol{\theta}_z\|_2 \le \frac{2m}{\rho_{\min}}. \tag{E.13}$$

*Proof.* We first derive some inequalities from (3.8). Denote $\Delta_{1jk}(x_1, x_j, x_k) := |p_{1,j,k}(x_1, x_j, x_k) - p_1(x_1)p_j(x_j)p_k(x_k)|$. Given any $j \neq k \geq 2$, let $u_j(x_j) = \sum_{s=1}^m \boldsymbol{\beta}_{js}\psi_{js}(x_j)$ and we have

$$\begin{aligned}
&\left|\boldsymbol{\beta}_j^T \mathbb{E}[K_h(X_1 - z)\boldsymbol{\Psi}_{1j}\boldsymbol{\Psi}_{1k}^T]\boldsymbol{\beta}_k\right| = \left|\iiint K_h(x_1 - z)u_j(x_j)u_k(x_k)p_{1,j,k}(x_1, x_j, x_k)dx_1 dx_j dx_k\right| \\
&\le \left|\int K_h(x_1 - z)dx_1 \int u_j(x_j)p_j(x_j)dx_j \int u_k(x_k)p_k(x_k)dx_k\right| + \int K(x_1)dx_1 \int \big|u_j(x_j)u_k(x_k)\big|\Delta_{1jk}dx_j dx_k \\
&\le m^{-1}\|\boldsymbol{\beta}_j\|_2\|\boldsymbol{\beta}_k\|_2 \iiint \Delta_{1jk}(x_1, x_j, x_k)dx_1 dx_j dx_k, \qquad\qquad \text{(E.14)}
\end{aligned}$$

where the last inequality is due to $\mathbb{E}[\psi_{jk}(X_j)] = 0$ and Assumption (**A6**). Therefore, we have

$$\begin{aligned}
\boldsymbol{\beta}_+^T \boldsymbol{\Sigma}_z \boldsymbol{\beta}_+ &= \sum_{j=1}^d \boldsymbol{\beta}_j^T \mathbb{E}[K_h(X_1 - z)\boldsymbol{\Psi}_{1j}\boldsymbol{\Psi}_{1j}^T]\boldsymbol{\beta}_j + \sum_{j \neq k} \boldsymbol{\beta}_j^T \mathbb{E}[K_h(X_1 - z)\boldsymbol{\Psi}_{1j}\boldsymbol{\Psi}_{1k}^T]\boldsymbol{\beta}_k \\
&\le \frac{\rho_{\max}}{m}\sum_{j=1}^d \|\boldsymbol{\beta}_j\|_2^2 + \sum_{j \neq k} m^{-1}\|\boldsymbol{\beta}_j\|_2\|\boldsymbol{\beta}_k\|_2 \iiint \Delta_{1jk}(x_1, x_j, x_k)dx_1 dx_j dx_k \le \frac{3\rho_{\max}}{2m}\|\boldsymbol{\beta}\|_2^2,
\end{aligned}$$

where the first inequality is due to (D.9) and the last inequality is due to (E.12) and the Gershgorin circle theorem. Similarly, we apply the Gershgorin circle theorem to the lower side and have

$$\begin{aligned}
\boldsymbol{\beta}_+^T \boldsymbol{\Sigma}_z \boldsymbol{\beta}_+ &= \sum_{j=1}^d \boldsymbol{\beta}_j^T \mathbb{E}[K_h(X_1 - z)\boldsymbol{\Psi}_{1j}\boldsymbol{\Psi}_{1j}^T]\boldsymbol{\beta}_j + \sum_{j \neq k} \boldsymbol{\beta}_j^T \mathbb{E}[K_h(X_1 - z)\boldsymbol{\Psi}_{1j}\boldsymbol{\Psi}_{1k}^T]\boldsymbol{\beta}_k \\
&\ge \frac{\rho_{\min}}{m}\sum_{j=1}^d \|\boldsymbol{\beta}_j\|_2^2 - \sum_{j \neq k} \Delta_{1jk} m^{-1}\|\boldsymbol{\beta}_j\|_2\|\boldsymbol{\beta}_k\|_2 \ge \frac{\rho_{\min}}{2m}\|\boldsymbol{\beta}\|_2^2,
\end{aligned}$$

where the first inequality is due to Assumption (**A3**). □

The following lemma shows the Lipschitz properties of $\boldsymbol{\theta}_z$.

**Lemma E.4.** Under Assumption (**A6**), we have

$$\|\boldsymbol{\theta}_z - \boldsymbol{\theta}_{z'}\|_2 \le 2m\rho_{\min}^{-2}(B/b)L\rho_{\max} \cdot |z - z'|,$$

where $C_K$ is a constant only depending on the kernel $K$.

*Proof.* The idea of proving this lemma is similar to Lemma E.3. Given any $j \neq k \geq 2$, again let $u_j(x_j) = \sum_{s=1}^m \boldsymbol{\beta}_{js}\psi_{js}(x_j)$ and we have

$$\begin{aligned}
&\big|\boldsymbol{\beta}_j^T \mathbb{E}[(K_h(X_1-z) - K_h(X_1-z'))\boldsymbol{\Psi}_{1j}\boldsymbol{\Psi}_{1k}^T]\boldsymbol{\beta}_k\big| \\
&= \Big|\int (K_h(x_1-z) - K_h(x_1-z'))u_j(x_j)u_k(x_k)p_{1,j,k}(x_1,x_j,x_k)dx_1dx_jdx_k\Big| \\
&= \Big|\int K(x_1)\sup_u \mathbb{E}[u_j(X_j)u_k(X_k)|X_1=u](p_1(z+x_1h) - p_1(z'+x_1h))dx_1\Big| \\
&\leq (B/b)L|z-z'|\int \big|u_j(x_j)u_k(x_k)\big|\Delta_{1jk}dx_jdx_k \\
&\leq (B/b)L|z-z'|m^{-1}\|\boldsymbol{\beta}_j\|_2\|\boldsymbol{\beta}_k\|_2 \iiint \Delta_{1jk}(x_1,x_j,x_k)dx_1dx_jdx_k.
\end{aligned}$$

Similarly, we also have

$$\begin{aligned}
&\big|\boldsymbol{\beta}_j^T \mathbb{E}[(K_h(X_1-z) - K_h(X_1-z'))\boldsymbol{\Psi}_{1j}\boldsymbol{\Psi}_{1j}^T]\boldsymbol{\beta}_j\big| \\
&= \Big|\int (K_h(x_1-z) - K_h(x_1-z'))u_j(x_j)^2 p_{1,j}(x_1,x_j)dx_1dx_j\Big| \\
&= \Big|\int K(x_1)\sup_u \mathbb{E}[u_j(X_j)^2|X_1=u](p_1(z+x_1h) - p_1(z'+x_1h))dx_1\Big| \leq (B/b)L|z-z'|m^{-1}\|\boldsymbol{\beta}_j\|_2^2.
\end{aligned}$$

Therefore, for any $\boldsymbol{\beta}_+ \in \mathbb{R}^{1+(d-1)m}$ and $z, z'$,

$$\begin{aligned}
\boldsymbol{\beta}_+^T(\boldsymbol{\Sigma}_z - \boldsymbol{\Sigma}_{z'})\boldsymbol{\beta}_+ &\leq (B/b)L|z-z'|\cdot m^{-1}\sum_{j=1}^d \|\boldsymbol{\beta}_j\|_2^2 \\
&\quad + (B/b)L|z-z'|\cdot \sum_{j\neq k} m^{-1}\|\boldsymbol{\beta}_j\|_2\|\boldsymbol{\beta}_k\|_2 \iiint \Delta_{1jk}(x_1,x_j,x_k)dx_1dx_jdx_k \\
&\leq \frac{2(B/b)L\rho_{\max}}{m}\|\boldsymbol{\beta}_+\|_2^2\cdot|z-z'|.
\end{aligned}$$

Therefore, combining with Lemma E.3, we can apply the matrix inverse perturbation inequality (see e.g., Demmel (1992)) and have

$$\|\boldsymbol{\theta}_z - \boldsymbol{\theta}_{z'}\|_2 \leq \|\boldsymbol{\Sigma}_z^{-1}\|_2^2\|\boldsymbol{\Sigma}_z - \boldsymbol{\Sigma}_{z'}\|_2 \leq 2m\rho_{\min}^{-2}(B/b)L\rho_{\max}\cdot|z-z'|.$$

□

## E.4 Proof of Lemma B.4

Applying the fact that $\widehat{\boldsymbol{\theta}}_z^T \widehat{\boldsymbol{\Sigma}}_z \widehat{\boldsymbol{\theta}}_z \le \boldsymbol{\theta}_z^T \widehat{\boldsymbol{\Sigma}}_z \boldsymbol{\theta}_z$, we have the following inequality

$$\begin{aligned}
\left(\widehat{\boldsymbol{\theta}}_z - \boldsymbol{\theta}_z\right)^T \widehat{\boldsymbol{\Sigma}}_z \left(\widehat{\boldsymbol{\theta}}_z - \boldsymbol{\theta}_z\right) &= \widehat{\boldsymbol{\theta}}_z^T \widehat{\boldsymbol{\Sigma}}_z \widehat{\boldsymbol{\theta}}_z - 2\widehat{\boldsymbol{\theta}}_z^T \widehat{\boldsymbol{\Sigma}}_z \boldsymbol{\theta}_z + \boldsymbol{\theta}_z^T \widehat{\boldsymbol{\Sigma}}_z \boldsymbol{\theta}_z \\
&\le 2\boldsymbol{\theta}_z^T \widehat{\boldsymbol{\Sigma}}_z \boldsymbol{\theta}_z - 2\left(\widehat{\boldsymbol{\theta}}_z^T \widehat{\boldsymbol{\Sigma}}_z - e_i\right)\boldsymbol{\theta}_z - 2e_i^T \boldsymbol{\theta}_z \\
&= 2\boldsymbol{\theta}_z^T \left(\widehat{\boldsymbol{\Sigma}}_z - \boldsymbol{\Sigma}_z\right)\boldsymbol{\theta}_z - 2\left(\widehat{\boldsymbol{\theta}}_z^T \widehat{\boldsymbol{\Sigma}}_z - e_i\right)\boldsymbol{\theta}_z \\
&\le 2\|\boldsymbol{\theta}_z\|_1^2 \|\widehat{\boldsymbol{\Sigma}}_z - \boldsymbol{\Sigma}_z\|_{\max} + 2\|\widehat{\boldsymbol{\Sigma}}_z \widehat{\boldsymbol{\theta}}_z - \mathbf{e}_1\|_{2,\infty} \|\boldsymbol{\theta}_z\|_1. \qquad \text{(E.15)}
\end{aligned}$$

We now study the rate of $\boldsymbol{\theta}_z$ in this subsection. We separate $\boldsymbol{\Sigma}_z$ into four blocks such that

$$\boldsymbol{\Sigma}_z = \begin{pmatrix} \boldsymbol{\Sigma}_z^{(1,1)} & \boldsymbol{\Sigma}_z^{(2,1)T} \\ \boldsymbol{\Sigma}_z^{(2,1)} & \boldsymbol{\Sigma}_z^{(2,2)} \end{pmatrix},$$

where $\boldsymbol{\Sigma}_z^{(1,1)} \in \mathbb{R}$, $\boldsymbol{\Sigma}_z^{(2,1)} \in \mathbb{R}^{(d-1)m}$ and $\boldsymbol{\Sigma}_z^{(2,2)} \in \mathbb{R}^{(d-1)m \times (d-1)m}$. By Lemma E.3, both $[\boldsymbol{\Sigma}_z^{(1,1)}]^{-1}$ and $[\boldsymbol{\Sigma}_z^{(2,2)}]^{-1}$ exist for any $z \in \mathcal{X}$. By the inversion formula of a block matrix, we have

$$\boldsymbol{\Sigma}_z^{-1} = \begin{pmatrix} \boldsymbol{\Theta}_z^{(1,1)} & \boldsymbol{\Theta}_z^{(2,1)T} \\ \boldsymbol{\Theta}_z^{(2,1)} & \boldsymbol{\Theta}_z^{(2,2)} \end{pmatrix},$$

where the concrete formulations of these four submatrices are

$$\begin{aligned}
\boldsymbol{\Theta}_z^{(1,1)} &= \left(\boldsymbol{\Sigma}_z^{(1,1)} - [\boldsymbol{\Sigma}_z^{(2,1)}]^T [\boldsymbol{\Sigma}_z^{(2,2)}]^{-1} \boldsymbol{\Sigma}_z^{(2,1)}\right)^{-1}, \\
\boldsymbol{\Theta}_z^{(2,1)} &= -\boldsymbol{\Theta}_z^{(1,1)} [\boldsymbol{\Sigma}_z^{(2,2)}]^{-1} \boldsymbol{\Sigma}_z^{(2,1)}, \\
\boldsymbol{\Theta}_z^{(2,2)} &= [\boldsymbol{\Sigma}_z^{(2,2)}]^{-1} - \boldsymbol{\Theta}_z^{(2,1)} [\boldsymbol{\Sigma}_z^{(2,1)}]^T [\boldsymbol{\Sigma}_z^{(2,2)}]^{-1}.
\end{aligned}$$

Denote $\Delta_{1,j}(x_1, x_j) = \left|p_{1,j}(x_1, x_j) - p_1(x_1)p_j(x_j)\right|$ for any $j \geq 2$. By (E.12), we have

$$\sum_{j=2}^{d}\sum_{k=1}^{m}\iint \Delta_{1,j}(x_1, x_j)dx_1dx_j \leq \frac{\rho_{\max}}{2}. \tag{E.16}$$

In order to bound $\boldsymbol{\theta}_z = (\boldsymbol{\Theta}_z^{(1,1)}, \boldsymbol{\Theta}_z^{(2,1)T})^T$, we first bound the $\ell_1$ norm of the second part

$$\|\boldsymbol{\Sigma}_z^{(2,1)}\|_1 = \sum_{j=2}^{d}\sum_{k=1}^{m}|\mathbb{E}[K_h(X_1 - z)\psi_{jk}(X_j)]| \,.$$

By the triangle inequality, we can bound the norm by

$$\begin{aligned}
\left\|\boldsymbol{\Sigma}_z^{(2,1)}\right\|_1 \leq & \sum_{j=2}^{d}\sum_{k=1}^{m}\left|\int K_h(x_1 - z)\psi_{jk}(x_j)p_1(x_1)p_j(x_j)dx_1dx_j\right| \\
& + \sum_{j=2}^{d}\sum_{k=1}^{m}\left|\int K_h(x_1 - z)|\psi_{jk}(x_j)|\Delta(x_1, x_j)dx_1dx_j\right| \\
\leq & \sum_{j=2}^{d}\sum_{k=1}^{m}|\mathbb{E}[K_h(X_1 - z)]\mathbb{E}[\psi_{jk}(X_j)]| + \sum_{j=2}^{d}\sum_{k=1}^{m}\iint \Delta_{1,j}(x_1, x_j)dx_1dx_j \leq \frac{\rho_{\max}}{2},
\end{aligned}$$

where the last inequality is due to $\mathbb{E}[\psi_{jk}(X_j)] = 0$ and (E.16). Hence, we have

$$\left\|[\boldsymbol{\Sigma}_z^{(2,2)}]^{-1}\boldsymbol{\Sigma}_z^{(2,1)}\right\|_1 \leq \left\|\boldsymbol{\Sigma}_z^{(2,1)}\right\|_1 \cdot \rho_{\min}^{-1}m \leq \rho_{\max}\rho_{\min}^{-1}m/2.$$

and we can also bound

$$\boldsymbol{\Sigma}_z^{(2,1)T}[\boldsymbol{\Sigma}_z^{(2,2)}]^{-1}\boldsymbol{\Sigma}_z^{(2,1)} \leq \left\|\boldsymbol{\Sigma}_z^{(2,1)}\right\|_1^2\rho_{\min}^{-1}m \leq \rho_{\max}^2\rho_{\min}^{-1}m/4$$

In fact, by Lemma E.3, $\boldsymbol{\Theta}_z^{(1,1)} > 0$. Combining with $\boldsymbol{\Sigma}_z^{(1,1)} = \mathbb{E}[K_h(X_1 - z)] = p_1(z) + o(1)$, we can have $\boldsymbol{\Sigma}_z^{(2,1)T}[\boldsymbol{\Sigma}_z^{(2,2)}]^{-1}\boldsymbol{\Sigma}_z^{(2,1)} \leq p_1(z) + o(1)$ for any $z \in \mathcal{X}$.

Summarizing the inequalities above, we have

$$\begin{aligned}\sup_{z\in\mathcal{X}}\|\boldsymbol{\theta}_z\|_1 &\le \sup_{z\in\mathcal{X}}|\boldsymbol{\Theta}_z^{(1,1)}| + \sup_{z\in\mathcal{X}}\|\boldsymbol{\Theta}_z^{(2,1)}\|_1 \\ &= \sup_{z\in\mathcal{X}}\left|\boldsymbol{\Sigma}_z^{(1,1)} - \boldsymbol{\Sigma}_z^{(2,1)T}[\boldsymbol{\Sigma}_z^{(2,2)}]^{-1}\boldsymbol{\Sigma}_z^{(2,1)}\right|^{-1} + \sup_{z\in\mathcal{X}}\left\|\boldsymbol{\Theta}_z^{(1,1)}[\boldsymbol{\Sigma}_z^{(2,2)}]^{-1}\boldsymbol{\Sigma}_z^{(2,1)}\right\|_1 \\ &= \sup_{z\in\mathcal{X}}\left\{(p_1(z)+O(1))^{-1} + (p_1(z)+O(1))^{-1}\cdot O(m)\right\} \le Cm. \end{aligned} \tag{E.17}$$

Plugging (3.9), (E.17) and (D.8) into (E.15), we have the rate in (E.17).

## E.5 Proof of Lemma B.1

We can expand the difference between two processes as

$$\widetilde{\mathbb{H}}_n'(z) - \widetilde{Z}_n'(z) = \underbrace{\sqrt{n^{-1}h}\sum_{i=1}^n K_h(X_{i1}-z)\eta_i'\boldsymbol{\Psi}_{i\bullet}^T\widehat{\boldsymbol{\theta}}_z}_{T_1(z)} + \underbrace{\sqrt{nh}(\mathbf{e}_1^T - \widehat{\boldsymbol{\theta}}_z^T\widehat{\boldsymbol{\Sigma}}_z)(\widehat{\boldsymbol{\beta}}_+ - \boldsymbol{\beta}_+)}_{T_2(z)},$$

where $\eta_i'$ is defined as

$$\eta_i' = f(X_{1i},\ldots,X_{di}) - \sum_{j=1}^d f_{mj}(X_{ji}) \qquad \text{for any } i\in[n]. \tag{E.18}$$

Using Theorem 3.2 and Lemma 3.3, with probability $1-c/n$, we have

$$\begin{aligned}&\sup_{z\in\mathcal{X}}|T_2(z)| \le \sqrt{nh}\|\widehat{\boldsymbol{\Sigma}}_z\boldsymbol{\theta}_z - \mathbf{e}_1\|_{2,\infty}\|\widehat{\boldsymbol{\beta}}_+ - \boldsymbol{\beta}_+\|_{2,1} \\ &\le C\sqrt{nh}\left(\sqrt{\frac{m\log^2 d}{nh}}\right)\cdot sm\left(\sqrt{\frac{\log(dmh^{-1})}{nh}} + \frac{\sqrt{s}}{m^{5/2}} + \frac{m^{3/2}\log(dh^{-1})}{n} + \frac{h^2}{\sqrt{m}}\right).\end{aligned} \tag{E.19}$$

Since $mh = o(1)$ and $h \asymp n^{-\delta}$ for $\delta > 1/5$, we have $\sup_{z\in\mathcal{X}}|T_2(z)| = o_P(n^{-1/10})$.

To bound $T_1(z)$, we first apply the triangle inequality and Cauchy-Schwartz inequality to decompose $T_1(z)$ into three smaller fragments

$$\begin{aligned}T_1(z) &= \sqrt{h/n}\sum_{i=1}^{n} K_h(X_{i1}-z)\eta_i' \boldsymbol{\Psi}_{i\bullet}^T(\widehat{\boldsymbol{\theta}}_z - \boldsymbol{\theta}_z) + \sqrt{h/n}\sum_{i=1}^{n} K_h(X_{i1}-z)\eta_i' \boldsymbol{\Psi}_{i\bullet}^T\boldsymbol{\theta}_z \\ &\le \sqrt{nh}\cdot T_{11}^{1/2}(z)\cdot T_{12}^{1/2}(z) + \sqrt{nh}\cdot T_{13}(z),\end{aligned}$$

where the three processes $T_{11}, T_{12}$ and $T_{13}$ are defined as follows

$$\begin{aligned}&T_{11}(z) = \frac{1}{n}\sum_{i=1}^{n} K_h(X_{i1}-z)(\boldsymbol{\Psi}_{i\bullet}^T(\widehat{\boldsymbol{\theta}}_z - \boldsymbol{\theta}_z))^2, \quad T_{12}(z) = \frac{1}{n}\sum_{i=1}^{n} K_h(X_{i1}-z)(\eta_i')^2 \\ &\text{and} \quad T_{13}(z) = \frac{1}{n}\sum_{i=1}^{n} K_h(X_{i1}-z)\eta_i' \boldsymbol{\Psi}_{i\bullet}^T\boldsymbol{\theta}_z.\end{aligned}$$

From Lemma B.4, we can bound the supreme of $T_{11}(z)$ by

$$\sup_{z\in\mathcal{X}} |T_{11}(z)| = \sup_{z\in\mathcal{X}} \left(\widehat{\boldsymbol{\theta}}_z - \boldsymbol{\theta}_z\right)^T \widehat{\boldsymbol{\Sigma}}_z \left(\widehat{\boldsymbol{\theta}}_z - \boldsymbol{\theta}_z\right) \le Cm\left(\sqrt{\frac{m\log(dm)}{nh}} + \frac{m}{nh} + \sqrt{\frac{\log(1/h)}{nh}}\right). \quad \text{(E.20)}$$

Let $\boldsymbol{\delta}, \boldsymbol{\xi}_z$ and $\boldsymbol{\zeta}_z$ be as defined in Lemma 7.2 and Lemma 7.3. From those two lemmas, with probability $1 - c/n$, we have

$$\sup_{z\in\mathcal{X}} |T_{12}(z)| \le \sup_{z\in\mathcal{X}} \frac{2}{n}\|\mathbf{W}_z^{1/2}\boldsymbol{\delta}\|_2^2 + \sup_{z\in\mathcal{X}} \frac{2}{n}\|\mathbf{W}_z^{1/2}(\boldsymbol{\xi}_z + \boldsymbol{\zeta}_z)\|_2^2 \le C(sm^{-4} + h^2). \quad \text{(E.21)}$$

Lemma 7.2 and Lemma 7.3 also give us

$$\begin{aligned}\sup_{z\in\mathcal{X}} |T_{13}(z)| &\le \sup_{z\in\mathcal{X}} \frac{1}{n}\|\boldsymbol{\Psi}^T\mathbf{W}_z(\boldsymbol{\delta} + \boldsymbol{\xi}_z + \boldsymbol{\zeta}_z)\|_{2,\infty}\|\boldsymbol{\theta}_z\|_1 \\ &\le Cm\left(\sqrt{s}\cdot m^{-5/2} + \sqrt{\frac{h\log(dh^{-1})}{n}} + \frac{m^{3/2}\log(dh^{-1})}{n} + \frac{h^2}{\sqrt{m}}\right).\end{aligned} \quad \text{(E.22)}$$

Combining (E.20), (E.21) with (E.22), if $h \asymp n^{-\delta}$ for $\delta > 1/5$ and $m \asymp n^p$ for $0 < p \le 10(\delta - 1/5)/3$, we have

$$\sup_{z\in\mathcal{X}} |T_1(z)| \le C\sqrt{nh}m^{3/4}h^2 = o(n^{-c}).$$

Combining this inequality with the rate of $\sup_{z\in\mathcal{X}} |T_1(z)|$ in (E.19), we have our lemma proved.

## E.6 Proof of Lemma B.3

We first bound the difference between $\widehat{\boldsymbol{\theta}}_z^T \boldsymbol{\Sigma}_z' \widehat{\boldsymbol{\theta}}_z$ and $\boldsymbol{\theta}_z^T \boldsymbol{\Sigma}_z' \boldsymbol{\theta}_z$ by applying triangle inequality and Cauchy-Schwartz inequality

$$
\begin{aligned}
&\left|\widehat{\boldsymbol{\theta}}_z^T \boldsymbol{\Sigma}_z' \widehat{\boldsymbol{\theta}}_z - \boldsymbol{\theta}_z^T \boldsymbol{\Sigma}_z' \boldsymbol{\theta}_z\right| \\
&\le (\widehat{\boldsymbol{\theta}}_z - \boldsymbol{\theta}_z)^T \boldsymbol{\Sigma}_z' (\widehat{\boldsymbol{\theta}}_z - \boldsymbol{\theta}_z) + 2(\widehat{\boldsymbol{\theta}}_z - \boldsymbol{\theta}_z)^T \boldsymbol{\Sigma}_z' \boldsymbol{\theta}_z \\
&\le h^{-1}(\widehat{\boldsymbol{\theta}}_z - \boldsymbol{\theta}_z)^T \widehat{\boldsymbol{\Sigma}}_z (\widehat{\boldsymbol{\theta}}_z - \boldsymbol{\theta}_z) + 2h^{-1/2}\sqrt{(\widehat{\boldsymbol{\theta}}_z - \boldsymbol{\theta}_z)^T \widehat{\boldsymbol{\Sigma}}_z (\widehat{\boldsymbol{\theta}}_z - \boldsymbol{\theta}_z)}\sqrt{\boldsymbol{\theta}_z^T \boldsymbol{\Sigma}_z' \boldsymbol{\theta}_z}.
\end{aligned} \tag{E.23}
$$

From Lemma B.4, we have the desired upper bound in the lemma that

$$
\sup_{z\in\mathcal{X}} (\widehat{\boldsymbol{\theta}}_z - \boldsymbol{\theta}_z)^T \widehat{\boldsymbol{\Sigma}}_z (\widehat{\boldsymbol{\theta}}_z - \boldsymbol{\theta}_z) \le m\left(\sqrt{\frac{m\log(dm)}{nh}} + \frac{m}{nh} + \sqrt{\frac{\log(1/h)}{nh}}\right). \tag{E.24}
$$

The following lemma gives us a bound on the term $\boldsymbol{\theta}_z^T \boldsymbol{\Sigma}_z' \boldsymbol{\theta}_z$.

**Lemma E.5.** Under Assumption (**A1**), for any $z \in \mathcal{X}$,

$$
\mathbb{E}[\boldsymbol{\Sigma}_z'] = h^{-1}\lambda(K)\big[\boldsymbol{\Sigma}_z + o(h)\big],
$$

where $\lambda(K) = \int K^2(u)du$. Furthermore, with probability at least $1 - c/n$,

$$
\sup_{z\in\mathcal{X}} ||\boldsymbol{\Sigma}_z' - \mathbb{E}[\boldsymbol{\Sigma}_z']||_{\max} \le C\left(\frac{1}{nh^2} + \frac{1}{\sqrt{nh^3}} + \sqrt{\frac{\log(dm)}{nmh^3}}\right).
$$

We defer the proof of the lemma to the end of the section. Using Lemma E.5, we have

$$
\begin{aligned}
\boldsymbol{\theta}_z^T \boldsymbol{\Sigma}_z' \boldsymbol{\theta}_z &\ge \boldsymbol{\theta}_z^T \mathbb{E}[\boldsymbol{\Sigma}_z'] \boldsymbol{\theta}_z - \|\boldsymbol{\Sigma}_z' - \mathbb{E}[\boldsymbol{\Sigma}_z']\|_{\max} \|\boldsymbol{\theta}_z\|_1^2 \\
&\ge h^{-1}\lambda(K)\boldsymbol{\theta}_z^T \boldsymbol{\Sigma}_z \boldsymbol{\theta}_z - \|\boldsymbol{\Sigma}_z' - \mathbb{E}[\boldsymbol{\Sigma}_z']\|_{\max} \|\boldsymbol{\theta}_z\|_1^2 - o(1) \\
&\ge h^{-1}\lambda(K)\mathbf{e}_1^T \boldsymbol{\theta}_z - Cm^2\left(\frac{1}{nh^2} + \frac{1}{\sqrt{nh^3}} + \sqrt{\frac{\log(dm)}{nmh^3}}\right) - o(1).
\end{aligned} \tag{E.25}
$$

We can also bound from the other direction as

$$\boldsymbol{\theta}_z^T \boldsymbol{\Sigma}_z' \boldsymbol{\theta}_z \le h^{-1}\lambda(K)\mathbf{e}_1^T\boldsymbol{\theta}_z + Cm^2\left(\frac{1}{nh^2} + \frac{1}{\sqrt{nh^3}} + \sqrt{\frac{\log(dm)}{nmh^3}}\right) + o(1). \tag{E.26}$$

Combining (E.23), (E.24), (E.25) and (E.26), if $mh = o(1), h = n^{-\delta}$ for $\delta > 1/5$, there exists a constant $c$ such that for any $z \in \mathcal{X}$,

$$\begin{aligned}\widehat{\boldsymbol{\theta}}_z^T \boldsymbol{\Sigma}_z' \widehat{\boldsymbol{\theta}}_z &\ge \boldsymbol{\theta}_z^T \boldsymbol{\Sigma}_z' \boldsymbol{\theta}_z - \left|\widehat{\boldsymbol{\theta}}_z^T \boldsymbol{\Sigma}_z' \widehat{\boldsymbol{\theta}}_z - \boldsymbol{\theta}_z^T \boldsymbol{\Sigma}_z' \boldsymbol{\theta}_z\right| \\ &= h^{-1}\left(\lambda(K)\mathbf{e}_1^T\boldsymbol{\theta}_z + o(1)\right) \ge ch^{-1}\mathbf{e}_1^T\boldsymbol{\theta}_z\end{aligned}$$

Similarly, we also have $\widehat{\boldsymbol{\theta}}_z^T \boldsymbol{\Sigma}_z' \widehat{\boldsymbol{\theta}}_z \le Ch^{-1}\mathbf{e}_1^T\boldsymbol{\theta}_z$. The proof will be done once we prove Lemma E.5.

*Proof of Lemma E.5.* For any $j, j' \in [d]$ and $k, k' \in [m]$, we have

$$\begin{aligned}\mathbb{E}[\boldsymbol{\Sigma}_z']_{jj'kk'} &= \int K_h^2(x-z)(\psi_{jk}(x_j)\psi_{j'k'}(x_{j'}))p_{1,j,j'}(x_1, x_j, x_{j'})dx_1 dx_j dx_{j'} \\ &= h^{-1}\int K^2(u)(\psi_{jk}(x_j)\psi_{j'k'}(x_{j'}))p_{1,j,j'}(z+uh, x_j, x_{j'})du dx_j dx_{j'} \\ &= h^{-1}\lambda(K)\int K(u)(\psi_{jk}(x_j)\psi_{j'k'}(x_{j'}))(p_{1,j,j'}(z, x_j, x_{j'}) + o(h))du dx_j dx_{j'} \\ &= h^{-1}\lambda(K)\left[\boldsymbol{\Sigma}_z + o(h)\right]_{jj'kk'}.\end{aligned}$$

The second part of the proof is similar to the proof of Lemma D.1. Consider the random variable

$$Z_{kk'jj'} = \sup_{z\in\mathcal{X}}(\mathbb{E}_n - \mathbb{E})[K_h^2(X_{i1} - z)\psi_k(X_{ij})\psi_{k'}(X_{ij'})].$$

Define the following two function classes

$$\begin{aligned}&\mathcal{G}_h = \left\{g_z(x_1, x_2, x_3) = h^{-2}K^2(h^{-1}(x_1 - z))\psi_k(x_2)\psi_{k'}(x_3)\,\Big|\, z \in \mathcal{X}, x_1, x_2, x_3 \in \mathcal{X}\right\} \text{ and} \\ &\mathcal{F}_h^2 = \left\{h^{-2}K^2(h^{-1}(\cdot - z))\,\middle|\, z \in \mathcal{X}\right\}.\end{aligned}$$

Using Lemma F.3, we bound the covering number by

$$\sup_Q N\left(\mathcal{F}_h, L^2(Q), \epsilon\right) \leq \left(\frac{8||K||_{\mathrm{TV}}^2 A^2}{h^2\epsilon}\right)^8,$$

where $Q$ is any measure on $\mathbb{R}$. Therefore the covering number for $\mathcal{G}_h$ satisfies

$$N\left(\mathcal{G}_h, L^2(\mathbb{P}), \epsilon\right) \leq \left(\frac{8||K||_{\mathrm{TV}}^2 A^2}{h^2\epsilon}\right)^8.$$

The envelope of $\mathcal{G}_h$ is $U = 4h^{-2}\|K\|_\infty$ and we bound the variance of the process by

$$\begin{aligned}\sigma_P^2 &:= \mathbb{E}\left[\left(K_h^2\left(X_1 - z\right)\left(\psi_k(X_j)\psi_{k'}(X_{j'})\right)\right)^2\right] \\ &= h^{-3}\mathbb{E}\left[K^2\left(h^{-1}(X_1 - z)\right)\mathbb{E}\left[(\psi_k^2(X_j)\psi_{k'}^2(X_{j'})\,|\,X_1\right]\right] \leq Cm^{-2}h^{-3}.\end{aligned}$$

Using Lemma F.2, we obtain the upper bound of the expectation

$$\mathbb{E}[Z_{kk'jj'}] \leq C_1\sqrt{\frac{\log(C_2 m)}{nm^2h^3}}.$$

As $|Z_{kk'jj'}| \leq 4h^{-2}$ and $\sigma^2 \leq Cm^{-2}h^{-3}$, Lemma F.4 gives us

$$\mathbb{P}\left(Z_{kk'jj'} \geq \mathbb{E}[Z_{kk'jj'}] + t\sqrt{Cm^{-2}h^{-3} + 4h^{-2}\mathbb{E}[Z_{kk'jj'}]} + 4t^2h^{-2}/3\right) \leq \exp(-nt^2).$$

By letting $t = 3\sqrt{\log(dm)/n}$, we obtain

$$\sup_{z\in\mathcal{X}}\max_{j,j'\geq 2}\left|\boldsymbol{\Sigma}_z'(j,j') - \mathbb{E}\boldsymbol{\Sigma}_z'(j,j')\right| = O_P\left(\frac{1}{nh^2} + \sqrt{\frac{\log(dm)}{nm^2h^3}}\right). \tag{E.27}$$

We also define the empirical process

$$\bar{Z}_{kj} = \sup_{z\in\mathcal{X}}\frac{1}{n}\sum_{i=1}^n K_h(X_{i1} - z)\psi_k(X_{ij}) - \mathbb{E}\left[K_h(X_1 - z)\psi_k(X_j)\right].$$

As above, we can show that the suprema of the empirical process has the convergence rate as

$$\sup_{z\in\mathcal{X}}\max_{j\geq 2}\left|\mathbf{\Sigma}'_z(j,1)-\mathbb{E}\mathbf{\Sigma}'_z(j,1)\right| = O_P\left(\frac{1}{nh^2}+\sqrt{\frac{\log(dm)}{nmh^3}}\right). \tag{E.28}$$

Finally, we have the following upper bound

$$\begin{aligned}\sup_{z\in\mathcal{X}}\left|\mathbf{\Sigma}'_z(1,1)-\mathbb{E}\mathbf{\Sigma}'_z(1,1)\right| &\leq \sup_{z\in\mathcal{X}}\left|\frac{1}{n}\sum_{i=1}^{n}K_h^2(X_{i1}-z)-\mathbb{E}[K_h^2(X_1-z)]\right| \\ &\leq C\left(\frac{1}{\sqrt{nh^3}}+\frac{1}{nh^2}\right).\end{aligned} \tag{E.29}$$

Combining (E.27), (E.28) and (E.29), with probability at least $1-c/n$, we have

$$\sup_{z\in\mathcal{X}}\|\mathbf{\Sigma}'_z-\mathbb{E}[\mathbf{\Sigma}'_z]\|_{\max}\leq C\left(\frac{1}{nh^2}+\frac{1}{\sqrt{nh^3}}+\sqrt{\frac{\log(dm)}{nmh^3}}\right),$$

which completes the proof of the Lemma. □

# F Results on Empirical Processes

**Lemma F.1** (Lemma H.2, Lu et al. (2015))**.** Let $\mathcal{F}_1$ and $\mathcal{F}_2$ be two function classes satisfying

$$N(\mathcal{F}_1,\|\cdot\|_{L_2(Q)},a_1\epsilon)\leq C_1\epsilon^{-v_1} \qquad \text{and} \qquad N(\mathcal{F}_2,\|\cdot\|_{L_2(Q)},a_2\epsilon)\leq C_2\epsilon^{-v_2}$$

for some $C_1, C_2, a_1, a_2, v_1, v_2 > 0$ and any $0 < \epsilon < 1$. Define $\|\mathcal{F}_\ell\|_\infty = \sup\{\|f\|_\infty, f\in\mathcal{F}_\ell\}$ for $\ell = 1,2$ and $U = \|\mathcal{F}_1\|_\infty \vee \|\mathcal{F}_2\|_\infty$. For the function classes $\mathcal{F}_\times = \{f_1f_2 \,|\, f_1\in\mathcal{F}_1, f_2\in\mathcal{F}_2\}$ and $\mathcal{F}_+ = \{f_1+f_2 \,|\, f_1\in\mathcal{F}_1, f_2\in\mathcal{F}_2\}$, we have for any $\epsilon\in(0,1)$,

$$\begin{aligned}N(\mathcal{F}_\times,\|\cdot\|_{L_2(Q)},\epsilon)&\leq C_1C_2\left(\frac{2a_1U}{\epsilon}\right)^{v_1}\left(\frac{2a_2U}{\epsilon}\right)^{v_2};\\ N(\mathcal{F}_+,\|\cdot\|_{L_2(Q)},\epsilon)&\leq C_1C_2\left(\frac{2a_1}{\epsilon}\right)^{v_1}\left(\frac{2a_2}{\epsilon}\right)^{v_2}.\end{aligned}$$

**Lemma F.2** (Corollary 5.1, Chernozhukov et al. (2014b))**.** Assume that the functions in $\mathcal{F}$ defined on $\mathcal{X}$ are uniformly bounded by an envelope function $F(\cdot)$ such that $|f(x)|\leq F(x)$ for all $x\in\mathcal{X}$ and $f \in \mathcal{F}$. Define $\sigma_P^2 = \sup_{f\in\mathcal{F}} \mathbb{E}[f^2]$. Let $Q$ be any measure over $\mathcal{X}$. If for some $A \geq e, V \geq 0$ and for all $\varepsilon > 0$, the covering entropy satisfies

$$\sup_Q N(\mathcal{F}, L^2(Q); \epsilon) \leq \left(\frac{A\|F\|_{L^2(Q)}}{\varepsilon}\right)^V,$$

then for any i.i.d. subgaussian mean zero random variables $\varepsilon_1, \dots, \varepsilon_n$ there exits a universal constant $C$ such that

$$\mathbb{E}\left[\sup_{f\in\mathcal{F}} \frac{1}{n}\sum_{i=1}^n (f(X_{i1}) - \mathbb{E}f(X))\right] \leq C\left[\sqrt{\frac{V}{n}}\sigma_P\sqrt{\log\frac{A\|F\|_{L^2(\mathbb{P})}}{\sigma_P}} + \frac{V\|F\|_{L^2(\mathbb{P})}}{\sqrt{n}}\log\frac{A\|F\|_{L^2(\mathbb{P})}}{\sigma_P}\right].$$

**Lemma F.3** (Lemma 3, Giné and Nickl (2009))**.** Let $K : \mathbb{R} \mapsto \mathbb{R}$ be a bounded variation function. Define the function class $\mathcal{F}_h = \{K((t - \cdot)/h) \,|\, t \in \mathbb{R}\}$. There exists $A < \infty$ such that for all probability measures $Q$ on $\mathbb{R}$, we have

$$\sup_Q N(\mathcal{F}_h, L^2(Q), \epsilon) \leq \left(\frac{2\|K\|_{\mathrm{TV}}A}{\epsilon}\right)^4, \text{ for any } \epsilon \in (0,1).$$

**Lemma F.4** (Bousquet (2002))**.** Let $X_1, \dots, X_n$ be independent random variables and $\mathcal{F}$ is a function class such that there exist $\eta_n$ and $\tau_n^2$ satisfying

$$\sup_{f\in\mathcal{F}} \|f\|_\infty \leq \eta_n \quad \text{and} \quad \sup_{f\in\mathcal{F}} \frac{1}{n}\sum_{i=1}^n \mathrm{Var}(f(X_{i1})) \leq \tau_n^2.$$

Define the random variable $Z$ being the suprema of an empirical process

$$Z = \sup_{f\in\mathcal{F}} \left|\frac{1}{n}\sum_{i=1}^n (f(X_{i1}) - \mathbb{E}f(X_{i1}))\right|. \tag{F.1}$$

Then for any $z > 0$, we have the following concentration inequality on the suprema

$$\mathbb{P}\left(Z \geq \mathbb{E}Z + z\sqrt{2(\tau_n^2 + 2\eta_n\mathbb{E}Z)} + 2z^2\eta_n/3\right) \leq \exp(-nz^2).$$

The following lemma gives the deviation inequality when $\mathcal{F}$ is not universally bounded.

**Lemma F.5** (Theorem 5.1, Chernozhukov et al. (2014b))**.** Let $F(\cdot)$ be the envelope function of $\mathcal{F}$ such that $F \in L^2(\mathbb{P})$ and $Z$ is defined in (F.1), For every $t \geq 1$, there exists a universal constant $C$ such that

$$\mathbb{P}\left(Z \geq 2\mathbb{E}Z + C(\sigma_P + \|F\|_{L^2(\mathbb{P})})z + \|F\|_{L^2(\mathbb{P})}z^2\right) \leq 1/z^2.$$